\documentclass{article}

    \PassOptionsToPackage{numbers, compress}{natbib}

\usepackage[final]{neurips_2024}

\usepackage[utf8]{inputenc} 
\usepackage[T1]{fontenc}    
\usepackage{hyperref}       
\usepackage{url}            
\usepackage{booktabs}       
\usepackage{amsfonts}       
\usepackage{nicefrac}       
\usepackage{microtype}      
\usepackage{xcolor}         

\usepackage{amsmath}
\usepackage{amssymb}
\usepackage{amsbsy}
\usepackage{graphicx}
\usepackage{wrapfig}
\usepackage{subcaption}
\usepackage{caption}
\usepackage{url}
\usepackage{color}
\usepackage[linesnumbered]{algorithm2e}
\usepackage{algpseudocode}
\usepackage{multicol}
\usepackage{multirow}
\usepackage{amsthm}
\usepackage{lipsum}
\usepackage{bbm}
\usepackage{xspace}
\usepackage{diagbox}
\usepackage{tikz}
\usepackage{makecell}

\newcommand{\swatch}[3]{\tikz[baseline=-0.6ex] \node[fill={rgb,255:red,#1; green,#2; blue,#3},shape=rectangle,draw=black,thick,minimum width=2mm,rounded corners=2pt](){};}
\newcommand{\bE}[0]{\mathbb{E}}
\newcommand{\bR}[0]{\mathbb{R}}

\newcommand{\cD}[0]{\mathcal{D}}
\newcommand{\cN}[0]{\mathcal{N}}
\newcommand{\cM}[0]{\mathcal{M}}
\newcommand{\cO}[0]{\mathcal{O}}
\newcommand{\cL}[0]{\mathcal{L}}
\newcommand{\w}[1]{\pmb{w}^{(#1)}}
\newcommand{\wprime}[1]{\pmb{w}'^{(#1)}}

\newcommand{\wm}[2]{\pmb{w}_{#2}^{(#1)}}

\newcommand{\wmtrain}[2]{\overline{\pmb{w}}_{#2}^{(#1)}}
\newcommand{\wmtrainparam}[2]{\overline{w}_{#2}^{(#1)}}
\newcommand{\wround}[1]{{w}^{(#1)}}
\newcommand{\deltar}[1]{{\Delta}^{(#1)}}
\newcommand{\vr}[1]{{\Lambda}^{(#1)}}

\newcommand{\subcM}[0]{\mathcal{\widetilde{M}}}
\newcommand{\subM}[0]{\widetilde{M}}
\newcommand{\partialw}[0]{w_{\overline{m}}}
\newcommand{\partialv}[0]{v_{\overline{m}}}
\newcommand{\mrange}[0]{\overline{m} \in \left[\frac{(m-1)d}{M} + 1 , \frac{md}{M}\right]}

\SetKwInput{KwData}{Input}
\SetKwInput{KwResult}{Output}
\SetKwProg{Function}{Function}{}{end}
\SetKwProg{Main}{Main}{}{end}

\theoremstyle{plain}
\newtheorem{theorem}{Theorem}[section]

\newtheorem{lemma}[theorem]{Lemma}

\theoremstyle{definition}

\newtheorem{assumption}[theorem]{Assumption}
\theoremstyle{remark}

\title{Thinking Forward: Memory-Efficient\\ 
Federated Finetuning of Language Models}

\author{%
  Kunjal Panchal\\
  University of Massachusetts\\
  Amherst, MA 01003-9264 \\
  \texttt{kpanchal@umass.edu} \\
\And
  Nisarg Parikh\\
  University of Massachusetts\\
  Amherst, MA 01003-9264 \\
  \texttt{nkparikh@umass.edu} \\
\And
  Sunav Choudhary \\
  Adobe Research \\
  Bangalore, India 560103 \\ 
  \texttt{schoudha@adobe.com} \\
\And  
  Lijun Zhang\\
  University of Massachusetts\\
  Amherst, MA 01003-9264 \\
  \texttt{lijunzhang@cs.umass.edu} \\
\And
  Yuriy Brun\\
  University of Massachusetts\\
  Amherst, MA 01003-9264 \\
  \texttt{brun@cs.umass.edu} \\
\And
  Hui Guan\\
  University of Massachusetts\\
  Amherst, MA 01003-9264 \\
  \texttt{huiguan@cs.umass.edu} \\
}
\newcommand{\projectname}{\textsc{Spry}\xspace}

\begin{document}

\maketitle

\begin{abstract}
  Finetuning large language models (LLMs) in federated learning (FL) settings has become increasingly important as it allows resource-constrained devices to finetune a model using private data. 
However, finetuning LLMs using backpropagation requires excessive memory (especially from intermediate activations) for resource-constrained devices. 
While Forward-mode Auto-Differentiation (AD) can significantly reduce memory footprint from activations, we observe that directly applying it to LLM finetuning results in slow convergence and poor accuracy.
In this paper, we introduce \projectname, an FL algorithm that splits trainable weights of an LLM among participating clients, such that each client computes gradients using Forward-mode AD that are closer estimations of the true gradients. 
\projectname achieves a low memory footprint, high accuracy, and fast convergence.
We formally prove that the global gradients in \projectname are unbiased estimators of true global gradients for homogeneous data distributions across clients, while heterogeneity increases bias of the estimates.
We also derive \projectname's convergence rate, showing that the gradients decrease inversely proportional to the number of FL rounds, indicating the convergence up to the limits of heterogeneity.
Empirically, \projectname reduces the memory footprint during training by 1.4--7.1$\times$ in contrast to backpropagation, while reaching comparable accuracy, across a wide range of language tasks, models, and FL settings. 
\projectname reduces the convergence time by 1.2--20.3$\times$ and achieves 5.2--13.5\% higher accuracy against state-of-the-art zero-order methods.
When finetuning Llama2-7B with LoRA, compared to the peak memory consumption of 33.9GB of backpropagation, \projectname only consumes 6.2GB of peak memory.
For OPT13B, the reduction is from 76.5GB to 10.8GB.
\projectname makes feasible previously impossible FL deployments on commodity mobile and edge devices.
Our source code is available for replication at \url{https://github.com/Astuary/Spry}.
\end{abstract}

\section{Introduction}
\label{sec:introduction}
In cross-device federated learning (FL), thousands of edge devices (called \emph{clients}) collaborate through an orchestrator (called \emph{server}) to jointly train a machine learning (ML) model~\cite{macmahan2017aFedLearning, kairouz2021open}. 
In each round of FL, the server sends an ML model to participating clients, who then update the model weights for several epochs on their individual data and send the new weights back to the server.  
The server aggregates the weights to update the model and initiates the next round of FL training. 
Due to the inherent privacy-preserving nature of FL, it has been adopted in many privacy-sensitive domains, such as healthcare~\cite{loftus2022FLHealthcare}, IoT~\cite{li2022healthiot, dara2022FLiot}, and e-commerce~\cite{pinelli2023flirt}.


In parallel, large language models (LLMs) have demonstrated impressive performance on natural language processing tasks~\cite{openai2024gpt4, anil2023palm}, creating a surge of interest in finetuning LLMs in FL settings~\cite{lin2022fednlp, tian2022fedbert}. 
However, the problem is challenging in practice because of the memory requirements in finetuning LLMs. 
LLMs can have billions of weights, and finetuning them with backpropagation requires dozens of GBs of memory. 
These requirements easily overwhelm edge devices, particularly mobile phones with limited memory. 
The memory footprint of finetuning LLMs mainly comes from model weights and their gradients, optimizer states, and intermediate activations.

There are three categories of existing algorithms that reduce the memory footprint of finetuning LLMs: 
parameter-efficient finetuning (PEFT)~\cite{borzunov2024distributed, hu2022lora, cai2023fedadapter, qiu2023oft, liu2022ia3,li2021prefixtuning}, quantization~\cite{jacob2018quant}, and 
zero-order gradient estimation methods~\cite{malladi2023mezo, xu2024fwdllm, feng2023baffle}.
Although PEFT and quantization can reduce the memory consumption from parameters and optimizer states, 
the memory consumed by intermediate activations remains a significant bottleneck because these methods still use backpropagation to finetune LLMs. 
Backpropagation requires storing all intermediate activations during the forward pass to estimate gradients in the backward pass.  
For example, finetuning a 4-bit quantized Llama2-7B model~\cite{touvron2023llama} with LoRA techniques~\cite{hu2022lora} requires $\sim$33.9GB of RAM, with 83.8\% used for intermediate activations. 
Zero-order methods leverage finite difference~\cite{richardson1955finitedifferences} to estimate gradients and thus reduce the memory consumption from intermediate activations~\cite{xu2024fwdllm, feng2023baffle, baydin2022gradients}.
However, these methods suffer from slow convergence and poor model quality because the accumulation of truncation and round-off errors~\cite{baydin2017autodiff}, a fundamental issue of finite difference, leads to noisy estimation of weight gradients. 

\emph{Forward-mode Auto-Differentiation (AD)}~\cite{baydin2017autodiff,baydin2022gradients} has the potential to address the memory consumption problems of backpropagation without introducing the round-off errors of finite difference. 
It estimates gradients by computing a Jacobian-vector product (\texttt{jvp}) based on random perturbations of weights during the forward pass, alleviating the need to store all intermediate activations, similar to zero-order methods. 
\texttt{jvp} represents how much changing the weights in the direction of a random perturbation affects the outputs. 
However, simply replacing backpropagation with Forward-mode AD in FL settings does not produce convergence speed and accuracy performance comparable to established federated optimizers based on backpropagation, such as \textsc{FedAvg}~\cite{macmahan2017aFedLearning}, \textsc{FedYogi}~\cite{reddi2021adaptive}, \textsc{FedSgd}~\cite{macmahan2017aFedLearning}.
Forward gradients are \emph{computationally inefficient} and \emph{inaccurate} in estimating true gradients when the number of trainable weights is large. 
We empirically observed that, for LLMs finetuned with the LoRA technique, Forward-mode AD suffers from 8.3--57.2\% accuracy loss and is 1.4--3.9$\times$ slower to converge for models whose number of trainable weights exceed approximately 1.15M (see Appendix~\ref{adx:ablation-studies}). 


In this paper, we propose \projectname\footnote{Named after its light memory consumption and speed.}, 
an FL algorithm for finetuning LLMs using Forward-mode AD while achieving low memory footprint, high accuracy, and fast convergence. 
\projectname tackles the shortcomings of Forward-mode AD by splitting the trainable weights among participating clients per FL round.
\projectname improves computation efficiency as each client only needs to perturb a small fraction of trainable weights to derive their gradients, reducing the number of computations in each forward pass. 
\projectname achieves higher accuracy and faster convergence because the smaller number of trainable weights for each participating client allows computing gradients that are closer estimations of the true gradients.  
In contrast to zero-order methods where one training iteration requires 20--100 forward passes, each with a different perturbation~\cite{feng2023baffle, xu2024fwdllm} to estimate weight gradients well, \projectname allows computing weight gradients from only one forward pass per iteration for each client. 
Unlike split learning~\cite{thapa2022splitfed}, \projectname does not need to transfer intermediate activations among clients.  
The union of the partial weights trained from each participating client in an FL round updates all the trainable weights of the language model.
Since only a subset of weights is finetuned per client, \projectname also saves client-to-server communication bandwidth. 

We formally prove that the global gradients aggregated on the server side in \projectname are unbiased estimators of the true global gradients in case of homogeneous data distributions across clients, while the heterogeneity increases the bias of the estimations.
We also derive the convergence rate of \projectname, showing that the norm of global gradients decreases linearly with the inverse of the number of FL rounds. 
We further discuss how configurations in \projectname affect the convergence behavior of the algorithm and empirically validate the theoretical analysis. 

We empirically evaluate \projectname's memory efficiency, accuracy, computation efficiency, and communication efficiency through experiments on a wide range of language tasks, models, and FL settings.
\projectname achieves within 0.6--6.2\% of the accuracy of the best-performing FL backpropagation, with 1.4--7.1$\times$ less memory consumption for each client and comparable time to convergence.
\projectname also outperforms zero-order-based baselines with 5.2--13.5\% higher accuracy, an average of 1.5--28.6$\times$ faster per-round computation time, and 1.2--20.3$\times$ faster convergence. 
We also compare \projectname's communication efficiency to that of \textsc{FedAvg} (per-epoch communication) and \textsc{FedSgd} (per-iteration communication). 
For communication frequency of per-epoch, \projectname reduces the number of model weights sent from a client to the server by $M$ times, where $M$ is the number of participating clients per round. 
For per-iteration communication frequency, each client of \projectname only needs to send back a scalar to the server, fixing the client-to-server total communication cost to $M$ scalar values. 


We make the following contributions:
\begin{enumerate}

    \item \projectname, the first work that demonstrate the potential of Forward-mode AD for finetuning language models (with 18M to 13B parameters) in FL settings with low memory footprint, high accuracy, and fast convergence. 
    
    
    \item  
    A federated optimization strategy that only requires a single forward pass per batch on each client to finetune a language model.

    \item A theoretical analysis of how \projectname's global gradients estimate true gradients based on the heterogeneity of FL clients, and a proof that \projectname's convergence is linearly dependent on the number of FL rounds when a client's learning rate is inversely proportional to the size of perturbations and client data heterogeneity.
    
    \item An empirical evaluation shows that \projectname consumes 1.4--7.1$\times$ less memory than its backpropagation-based counterparts, and converges 1.2--20.3$\times$ faster with 5.2--13.5\% higher accuracy compared to its zero-order counterparts.
\end{enumerate}

\section{Forward-mode Automatic Differentiation}
\label{sec:preliminaries}

This section presents the background on Forward-mode Auto-Differentiation (AD) necessary to follow the work. Related works on zero-order optimization methods, Forward-mode AD, and FL for LLMs are discussed in detail in Appendix~\ref{sec:related-work}. 

Forward-mode AD computes gradients by measuring how changes in model weights, in the direction of a random perturbation, affect the loss.
Since these gradients are derived from a forward pass, they are referred to as \emph{forward gradients}~\cite{baydin2022gradients}. 
In contrast, backpropagation (also Reverse-mode AD) calculates a direction to adjust weights in, to decrease the loss.
Formally, for each training iteration, given the trainable weights $\pmb{w}$, Forward-mode AD generates a random perturbation $\pmb{v}$ whose size is the same as $\pmb{w}$.
In one forward pass, given training data $\cD$, Forward-mode AD computes the value of the objective function $f(\pmb{w}; \cD)$ 
and the 
Jacobian-vector product (\texttt{jvp}) as follows: 
\begin{align}
        \mathbf{J}_f \pmb{v} = \nabla f_{\pmb{v}}(\pmb{w}; \cD) = \begin{bmatrix}
        \frac{\partial f(\pmb{w}; \cD)}{\partial w_1} & \dots & \frac{\partial f(\pmb{w}; \cD)}{\partial w_d}
    \end{bmatrix}
    \begin{bmatrix}
        v_1 & \hdots & v_d
    \end{bmatrix}^T.
    \label{eq:jvp}
\end{align}
This \texttt{jvp} is a scalar for neural networks since the output of the objective function $f$ is a scalar.
Multiplying $\texttt{jvp}$ with the perturbation $\pmb{v}$ gives us the unbiased estimate of true gradients~\cite{baydin2022gradients}, 
\begin{align}
    \nabla F (\pmb{w})  
    & = \bE_{\pmb{v}} \left[ \mathbf{J}_f \pmb{v} \cdot \pmb{v} \right] = \bE_{\pmb{v}, \cD} \left[ \left(\begin{bmatrix}
        \frac{\partial f(\pmb{w}; \cD)}{\partial w_1} & \dots & \frac{\partial f(\pmb{w}; \cD)}{\partial w_d}
    \end{bmatrix} \pmb{v}^T \right) \pmb{v} \right] \\
    & = \bE_{\cD} \begin{bmatrix}
        \frac{\partial f(\pmb{w}; \cD)}{\partial w_1} & \dots & \frac{\partial f(\pmb{w}; \cD)}{\partial w_d}
    \end{bmatrix} = \bE_{\cD} \left[\nabla f(\pmb{w}; \cD)\right].
    \label{eq:gradient-approximation}
\end{align}
The partial derivative ${\partial f(\pmb{w}; \cD)}/{\partial \pmb{w}}$ is computed by chain rule on intermediate activations in the forward pass~\cite{baydin2017autodiff}.
Unlike backpropagation where all the intermediate activations need to be stored during the forward pass, Forward-mode AD only stores the previous layer's activations in the forward pass for the chain rule derivatives.
Hence, the memory overhead of deriving gradients would be the size of the largest activation in the forward pass.

\section{\projectname: Memory-Efficient Federated Finetuning of Language Models}
\label{sec:methodology}
\begin{figure}
    \centering
    \includegraphics[width=\textwidth]{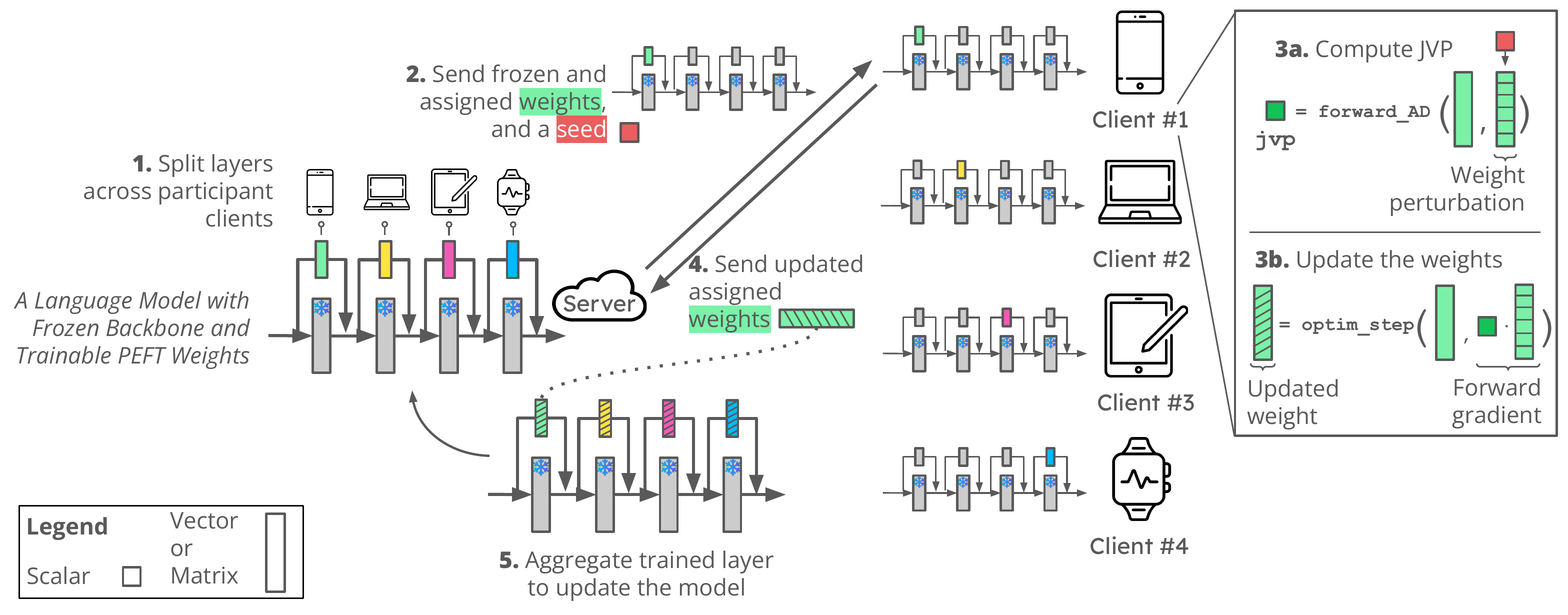}
    \caption{Overview of \projectname{}, a federated learning framework to finetune language models with low memory footprint. The term ``PEFT'' stands for parameter-efficient fine-tuning.}
    \vspace{-0.7cm}
  \label{fig:workflow-spry}
\end{figure}

While Forward-mode AD can decrease memory usage during model finetuning, merely substituting backpropagation with Forward-mode AD in FL scenarios often results in poor accuracy and computational inefficiency.
To address this challenge, \projectname{} recognizes that Forward-mode AD operates more efficiently and yields better gradient estimations when the trainable weight count is minimized.
Therefore, \projectname{} optimizes performance by distributing trainable weights among participating clients, assigning each client a responsibility to compute gradients for only a subset of trainable weights.
\projectname{} is compatible with PEFT methods such as \textsc{IA3}~\cite{liu2022ia3}, \textsc{Adapter}-based methods~\cite{zhang2024llamaadapter}, \textsc{BitFit}~\cite{zaken2022bitfit}, and \textsc{LoRA}~\cite{hu2022lora}, which mitigate the memory consumption from gradients and optimizer states. 
In this work, we focus on \textsc{LoRA} due to its demonstrated superiority, as highlighted in Appendix~\ref{adx:ablation-studies}.

Figure~\ref{fig:workflow-spry} gives an overview of \projectname.
It includes 5 main steps:
\textbf{(1)}~At the beginning of each FL round, the server assigns a few trainable layers to each of the participating clients of that round. 
\textbf{(2)}~Each client is sent 
(i)~the trainable weights of the layers assigned to it, 
(ii)~frozen weights of the remaining layers if not previously received, and 
(iii)~a scalar seed value. 
\textbf{(3)}~On the client side, weight perturbations for the allocated layers are generated based on the received seed. 
These perturbations are utilized to update the assigned weights through computation of the forward gradients. 
\textbf{(4)}~Clients only transmit the updated weights back to the server.
\textbf{(5)}~The server aggregates all the trained layer weights and updates the language model for the subsequent round. 

Next, we discuss the two key steps of \projectname in detail: Step (1), where the server assigns trainable weights to the participating clients and Step (3), where each client finetunes the assigned trainable weights using Forward-mode AD.



\subsection{Assigning Trainable Layers to Clients at the Server-side}
\label{subsubsec:splitting-layers}
To enable closer gradient estimations through forward gradients, the server reduces the number of trainable weights per client by selecting a layer and assigning it to a client in a cyclic manner. 
With \textsc{LoRA}, the server selects a LoRA layer, which consists of a pair of weights ($w_A$ and $w_B$ matrices) for each client. 
When \# trainable layers $>$ \# participating clients, each client will be assigned more than one layer.
Otherwise, each layer will be assigned to more than one client. 
The server aggregates the trained weights from each client and updates the model using adaptive optimizers such as \textsc{FedYogi}. 
Adaptive optimizers are shown to be less prone to noisy updates compared to \textsc{FedAvg} in the literature~\cite{reddi2021adaptive, panchal2023flash}. 
The server keeps a mapping of layer names to client IDs,  hence it can gather updated layer weights from all clients and update the model.

FL often faces the data heterogeneity issue, where the data distribution of one client differs from another, leading to poor model accuracy. 
While the primary aim of \projectname{} does not directly tackle this issue, it seamlessly integrates with existing finetuning-based personalization techniques~\cite{wu2022motley} to mitigate it. 
\projectname{} distributes trainable classifier layers to all participating clients, enabling each client to finetune these layers to personalize the jointly trained model. 

\subsection{Finetuning Weights with Forward Gradients on the Client-side}

Clients update the assigned trainable weights with gradients estimated through Forward-mode AD. 
Specifically, each participating client will get a copy of the trainable weights assigned to it and a scalar seed value from the server.
Using the seed value, the client generates a random perturbation for each trainable weight, following a normal distribution with a mean of 0 and a standard deviation of 1. 
Forward gradients are obtained during forward pass, as shown in 
Eq.~\ref{eq:gradient-approximation}.
The subsequent steps depend on the communication frequency. 

\textbf{Per-Epoch Communication.} 
Per-epoch communication means that each client transmits the updated trainable weights to the server after every one or more epochs. Locally, the trainable weights are updated using optimizers such as \textsc{SGD} and \textsc{Adam}. Only the updated trainable weights are sent back to the server, which reduces communication costs. Each client transmits the weights of \(\max \left\{ \frac{\text{\#Trainable Layers}}{\text{\#participating clients}}, 1\right\}\) layers. This means that if there are more trainable layers than participating clients, each client sends back weights for multiple layers. 


\textbf{Per-Iteration Communication.}
Unlike \textsc{FedSgd}, where gradients are sent back to the server and aggregated after each iteration, \projectname offers additional communication cost savings. In this communication mode, \projectname only requires sending the \texttt{jvp} scalar value back to the server after each iteration of fine-tuning. Since the server has the seed value, it can generate the same random perturbation used by each client. Using the received \texttt{jvp} values and the generated random perturbations, the server can then compute the gradients and update the model weights.

A detailed breakdown of communication and computation costs is given in Appendix~\ref{adx:communication-and-computation-costs}.

\vspace{-0.2cm}
\section{Theoretical Analysis}
\label{sec:convergence-analysis}
This section theoretically analyzes convergence behaviors of \projectname.
\projectname has a unique aggregation rule for the trainable weights as each client trains a subset of weights. 
It is also the first work to utilize forward gradients to train LLMs in FL settings where clients could have heterogeneous data distributions.
Therefore, we theoretically analyze the effects of
(a)~data heterogeneity on gradient estimations of \projectname, and
(b)~configurations in \projectname, including the number of FL rounds, dimension of perturbations, the number of perturbations per iteration, data heterogeneity, and the number of participating clients, on the convergence of \projectname. 
The proofs are detailed in Appendix~\ref{adx:proofs}.

\begin{theorem} [Estimation of the Global Gradient]
\label{thm:estimation-global-gradient-main}
    In \projectname, global forward gradients $\nabla \hat{f}$ of the trainable weights $w \in \bR^d$, with the corresponding weight perturbations $v\in \bR^d$, computed by $M$ participating clients is estimated in terms of true global gradients $\nabla f$ as,
    {
    \footnotesize
    \begin{align}
    \!\!\!\!\bE &[ \nabla \hat{f}(w, v; \cD)]\!=\!\nabla f(w)\!+\!\frac{1}{\subM}\!\!
        \begin{bmatrix}
            \!\sum_{m \in \subcM_1} \sum_{c=1}^{C} \alpha_{m,c} \bE_{(x,y_c) \in \cD} \left[ \nabla \hat{f}_m (w_{\left[1, \frac{d}{M} \right]}, v_{\left[1, \frac{d}{M} \right]}; (x,y_c)) \right]\\
            \sum_{m \in \subcM_2}\!\!\sum_{c=1}^{C}\!\!\alpha_{m,c} \bE_{(x,y_c) \in \cD}\!\!\left[\!\nabla \hat{f}_m\!(w_{\left[\frac{d}{M}+1,\frac{2d}{M}\right]}, v_{\left[\frac{d}{M}+1,\frac{2d}{M}\right]};\!(x,y_c)) \right]\!\!\!\\
        \vdots 
        \end{bmatrix}^T\!\! \label{eq:estimation-bias-main}
    \end{align}
    }%
    where the expectation is under the randomness of sampled data and random perturbation $v$. $C$ is total number of classes and $\alpha_{m,c} = \left(\frac{n_c}{|\cD|} - \frac{n_{m, c} \alpha_c}{|\cD_m|} \right)$. 
    For a class $c$, $n_c$ is its sample count, $\alpha_c$ is its Dirichlet concentration parameter. For a client $m$, $n_{m,c}$ is the sample count of the $c^{th}$ class and $\cD_m$ is the size of the data of client $m$. The global data is $\cD = \sum_{m \in \cM} \cD_m$. $\subcM$ is the set of clients training an arbitrary subset of weights, $\subM = |\subcM_i|, \forall i \in [M/d]$.
\end{theorem}
\textbf{Discussion.}
We focus on analyzing how data heterogeneity affects the estimation error between the global forward gradient and the global true gradient. 
Specifically, the estimation error of \projectname 
depends on the coefficient $\alpha_{m,c} = \left(\frac{n_c}{|\cD|} - \frac{n_{m, c} \alpha_c}{|\cD_m|} \right)$.  
%
In \textit{data homogeneous} settings, where all clients share the same data distribution, the Dirichlet concentration parameter $\alpha_c$, is 1 for any class $c$.
As the ratio of $n_c / |\cD|$ (total samples in a class to total samples globally) matches the ratio of $n_{m,c} / |\cD_m|$ (total samples in a class to total samples for a client $m$), the bias term becomes 0 since $\alpha_{m,c} = 0, \forall m,c$.
Hence, 
the \textit{global forward gradients are unbiased estimators} of the true global gradients. 
In \textit{data heterogeneous} settings, 
when data across clients becomes more heterogeneous, $\alpha_c \rightarrow 0$ and thus $\alpha_{m,c} \rightarrow n_c / |\cD|$, increasing the estimation error.
Hence the global forward gradients are \textit{biased estimators} that \textit{depend on the distances of the data distributions} across participating clients. 

\begin{theorem}[Convergence Analysis]
Under the assumptions on $L$-smoothness (Asmp~\ref{asm:smoothness}), bounded global variance $\sigma_g^2$ between global true gradients and aggregated expected gradients of each client (Asmp~\ref{asm:bounded-global-variance}), and bound on gradient magnitude $G$ (Asmp~\ref{asm:bounded-gradients}); and the following conditions on the local learning rate $\eta_\ell$,
    {
    \footnotesize
    \begin{align}
        \eta_\ell &= \min \Bigg \{ \cO \left( \frac{\tau^2}{\sqrt{\beta_2} \eta G L} \right)^{\frac{1}{2}}, \cO \left( \frac{1}{\sqrt{\beta_2} G} \right), \cO \left( \frac{\tau^3}{\sqrt{\beta_2(1 - \beta_2)} G^2}\right)^{\frac{1}{2}}, \nonumber\\
        &\qquad\qquad\qquad \cO \left( \frac{\subM K}{\beta_2 G (3d + K - 1) \sum_{m \in [M]} \sum_{c \in [C]} \alpha_{m, c}^2} \right) \Bigg \}; \label{eq:conditions-on-local-learning-rate-main}
    \end{align}
    }%
    The global true gradients of \projectname satisfies the following bound,
    {
    \footnotesize
    \begin{align}
        & \min_{0 \leq r \leq R} \bE_r || \nabla f(\wround{r})||^2 \leq \frac{f(\wround{0}) - \bE_R [f(\wround{R})]}{\eta R} \nonumber \\
        &\qquad + \left(2 + \frac{\eta \eta_\ell L}{2 \tau^2} + \frac{ \sqrt{1 - \beta_2}G \eta_\ell}{\tau^3} \right) \left(\frac{\sigma_g^2 (1 - s) (3d + K - 1)}{\subM K}\right)  \sum_{m \in \cM}  \sum_{c \in [C]} \alpha_{m, c}^2, \label{eq:gradient-norm-upper-bound-main}
    \end{align}
    }%
    where $R$ is the total number of FL rounds, $w \in \bR^d$ are trainable weights, $v \in \bR^d$ are random perturbation, $K$ is count of random perturbations per iteration, $\eta$ is global learning rate, $\tau$ is adaptability constant, and $s$ is client sampling rate.
    Rest of the symbols are defined in Theorem~\ref{thm:estimation-global-gradient-main}. 
    \label{thm:convergence-main}
\end{theorem}
\textbf{Discussion.} We focus on analyzing how different configurations in \projectname affect its convergence. 
\textbf{(a)}~\textit{The number of FL rounds ($R$):} The upper bounds of the norm of the global forward gradient in Eq.~\ref{eq:gradient-norm-upper-bound-main} decrease in proportion to the inverse of $R$, indicating the convergence of the algorithm up to the limits of data heterogeneity.   
\textbf{(b)}~\textit{Dimension of perturbations ($d$):}
~$\eta_\ell \propto \frac{1}{d}$ shows that as the number of weights to be perturbed increases, the learning rate must decrease.
A lower learning rate can make convergence slower, or worse, as our empirical experiments in Appendix~\ref{subsubsec:ablation-splitting-layers} will show, not converge at all. 
\textbf{(c)}~\textit{The number of perturbations per iteration ($K$):}
We observe $K$ both in nominator and denominator, which indicates that increasing $K$ brings little advantage in convergence speed.
Results in Appendix~\ref{subsubsec:perturbations-per-batch} confirm the above statement. 
\textbf{(d)}~\textit{Data heterogeneity:} $\eta_\ell \propto \frac{1}{\alpha_{m,c}^2}$ shows that more homogeneous data distributions across clients allow higher learning rate, and hence faster error reduction. This observation is corroborated by   the comparison of convergence speeds between homogeneous and heterogeneous clients in Appendix~\ref{adx:additional-results}. 
\textbf{(e)}~\textit{The number of clients training same subset of weights:} $\eta_\ell \propto \subM$ shows that more clients training the same subset of weights is beneficial for faster convergence, similar observation is shown in Appendix~\ref{subsubsec:ablation-client-count}.




The theorem also sheds light on the accuracy performance gap between Forward-mode AD and backpropagation in FL settings.
The upper bounds on the global forward gradient norm includes a second term that increases with $\alpha^2_{m,c}$ and variance $\sigma_g^2$, but does not decrease with $R$.  
This results in a gap between estimation errors of backpropagation-based methods like \textsc{FedAvg} and \projectname.


\vspace{-0.2cm}
\section{Empirical Evaluation}
\vspace{-0.2cm}
\label{sec:experiments-and-results}
\vspace{-0.1cm}
We empirically evaluate \projectname on 8 language tasks, 5 medium, and 3 large language models, under various FL settings.
Our evaluation measures \projectname's prediction performance, peak memory consumption, and time-to-convergence.
We also ablate \projectname's components to study the impact of communication frequency, the number of trainable parameters, the number of perturbations per iteration, the number of participating clients, and the importance of splitting layers on performance.  

\textbf{Datasets, Tasks, and Models.}
Our evaluation uses 8 datasets: \textbf{AG News}~\cite{zhang2015agnewsyelpyahoo} (4-class classification), \textbf{SST2}~\cite{socher2013sst2} (2-class classification), \textbf{Yelp}~\cite{zhang2015agnewsyelpyahoo} (2-class classification), \textbf{Yahoo}~\cite{zhang2015agnewsyelpyahoo} (10-class classification), \textbf{SNLI}~\cite{bowman2015snli} (3-class classification), \textbf{MNLI}~\cite{williams2018mnli} (3-class classification), \textbf{SQuADv2}~\cite{rajpurkar2018squad} (Closed-book question answering), and \textbf{MultiRC}~\cite{khashabi2018multirc} (2-class classification).
We chose these datasets because they allow us to generate heterogeneous splits in FL settings using Dirichlet distribution~\cite{panchal2023flow}.
The default dataset split is across 1,000 clients, except the smallest datasets \textbf{SST2} and \textbf{MultiRC}, where there are 100 clients.
\textbf{SQuADv2} has 500 total clients.
Each dataset has two versions: 
\textbf{(i)}~Dirichlet $\alpha=1.0$ (Homogeneous split), and 
\textbf{(ii)}~Dirichlet $\alpha=0.1$ (Heterogeneous split).

Our evaluation uses the following language models: \textbf{OPT13B}~\cite{zhang2022opt}, \textbf{Llama2-7B}~\cite{touvron2023llama}, \textbf{OPT6.7B}~\cite{zhang2022opt}, \textbf{RoBERTa~Large} (355M)~\cite{liu2019roberta}, \textbf{BERT~Large} (336M)~\cite{devlin2018bert}, \textbf{BERT~Base} (110M)~\cite{devlin2018bert}, \textbf{DistilBERT~Base} (67M)~\cite{sanh2019distilbert}, and \textbf{Albert~Large~V2} (17.9M)~\cite{lan2019albert}.
For the billion-sized models, we use 4-bit quantization. 
For all the models, we use \textsc{LoRA} as the \textsc{PEFT} method. Appendix~\ref{adx:datasets-hyperparameters} describes the datasets and hyperparameters in more detail.

\textbf{Comparison Counterparts and Metrics.}
We compare \projectname to 
(a)~Backpropagation-based federated optimizers \textsc{FedAvg}~\cite{macmahan2017aFedLearning}, \textsc{FedYogi}~\cite{reddi2021adaptive}, \textsc{FedSgd} (Variant of \textsc{FedAvg} with per-iteration communication)~\cite{macmahan2017aFedLearning},
(b) Zero-order federated methods \textsc{FedMeZO} (federated version of \textsc{MeZO}~\cite{malladi2023mezo}), \textsc{Baffle}~\cite{feng2023baffle}, \textsc{FwdLLM}~\cite{xu2024fwdllm}, all based on finite difference.
\textsc{MeZO} uses prompt-based finetuning to improve the performance of finite differences.
\textsc{FwdLLM} generates a random perturbation that has a high cosine similarity to the global gradients of the previous rounds.
\textsc{Baffle} generates $\sim$100--500 perturbations per iteration.
More details of these methods are in Appendix~\ref{sec:related-work}.
The original implementations of \textsc{FwdLLM} and \textsc{Baffle} had excessive memory usage in their implementations.
We improve their codebase to be memory-efficient by perturbing only the trainable weights, similar to \projectname. 
We refer to our implementation as \textsc{FwdLLM+} and \textsc{Baffle+}.


Evaluation of \projectname and its counterparts for classification tasks is on \textbf{generalized accuracy}  $Acc_g$ and \textbf{personalized accuracy} $Acc_p$, which are metrics measured on server-side aggregated model and client-side locally updated model, respectively.  
Similarly, for question-answering tasks, we measure \textbf{Exact Matches} and \textbf{F1 Score}.
We also measure \textbf{time to convergence} and \textbf{peak memory consumption} during training. 
Our convergence criterion is the absence of change in the variance of a performance metric, assessed at intervals of 50 rounds.  

\projectname is implemented in Flower~\cite{beutel2020flower} library.
Quantization is done using AutoGPTQ~\cite{autogptq}.
For the zero-order methods, we used their respective client-side implementations with the server simulation structure of Flower. 
We utilized two Nvidia 1080ti to conduct all experiments of sub-billion sized models and billion-sized models for \projectname and its zero-order methods.
We used two RTX8000s and two A100s for Llama2-7B and OPT models on backpropagation-based methods respectively.
Each experiment was run thrice with 0, 1, and 2 as seeds.

\vspace{-0.18cm} 
\subsection{Accuracy Performance Comparison}
\vspace{-0.18cm} 
\label{subsec:performance-comparison}
\begin{table}
\captionsetup{justification=centering}
\scriptsize
\tabcolsep=0.06cm
\caption{Generalized accuracy 
for \projectname and its backpropagation- and zero-order-based counterparts on RoBERTa Large and LLMs. 
SQuADv2 uses F1 score.
$\uparrow$ shows that higher values are better.\\
The datasets are split with Dir $\alpha=0.1$.
$\diamond$ = Llama2-7B. 
$\star$ = OPT6.7B. 
$\square$ = OPT13B. \\
\projectname outperforms the best-performing zero-order-based methods by 5.15--13.50\% and approaches the performance of backpropagation-based methods, with a difference of 0.60--6.16\%. 
}
\label{tbl:acc-gen-main}

\resizebox{\textwidth}{!}{
\begin{tabular}{lcccccccc}
\toprule 
 \multirow{3}{*}{} 
 & \multicolumn{2}{c}{\begin{tabular}[c]{@{}c@{}}Backpropagation-based \\ Methods $\uparrow$\end{tabular}} & \multicolumn{3}{c}{\begin{tabular}[c]{@{}c@{}}Zero-order-based \\ Methods $\uparrow$\end{tabular}} & \begin{tabular}[c]{@{}c@{}}First-order \\ Forward Mode AD $\uparrow$\end{tabular} & \multicolumn{2}{c}{\begin{tabular}[c]{@{}c@{}}Difference between \\ performances of \projectname  and  \\ \end{tabular}} \\ \cmidrule{2-7} 
 & \textsc{FedAvg}  & \multicolumn{1}{c}{\textsc{FedYogi}} & \textsc{FwdLLM+} & \textsc{FedMeZO} & \multicolumn{1}{c}{\textsc{Baffle+}} & \projectname  & \begin{tabular}[c]{@{}c@{}}best-performing \\ backpropagation \\ method $\uparrow$\end{tabular} & \begin{tabular}[c]{@{}c@{}}best-performing\\ zero-order \\ method $\uparrow$\end{tabular} \\ \midrule 
AG News & 93.07\% & \multicolumn{1}{c}{92.77\%} & 76.94\% & 70.56\% & \multicolumn{1}{c}{57.69\%} & 87.89\% & $-$5.18\% & 10.95\% \\
SST2 & 88.00\% & \multicolumn{1}{c}{92.14\%} & 84.41\% & 72.17\% & \multicolumn{1}{c}{61.57\%} & 91.54\% & $-$0.60\% & \phantom{0}7.13\% \\
SNLI & 86.45\% & \multicolumn{1}{c}{79.31\%} & 74.30\% & 69.57\% & \multicolumn{1}{c}{62.10\%} & 82.66\% & $-$3.79\% & \phantom{0}8.36\% \\
MNLI & 84.29\% & \multicolumn{1}{c}{84.98\%} & 72.66\% & 66.66\% & \multicolumn{1}{c}{62.85\%} & 80.32\% & $-$4.66\% & \phantom{0}7.66\% \\
Yahoo & 67.37\% & \multicolumn{1}{c}{63.08\%} & 56.06\% & 44.69\% & \multicolumn{1}{c}{37.81\%} & 61.21\% & $-$6.16\% & \phantom{0}5.15\% \\
Yelp & 90.48\% & \multicolumn{1}{c}{79.10\%} & 71.83\% & 65.10\% & \multicolumn{1}{c}{55.99\%} & 85.33\% & $-$5.15\% & 13.50\%  \\
MultiRC $\diamond$ & 47.56\% & \multicolumn{1}{c}{72.53\%} & 64.58\% & N/A & \multicolumn{1}{c}{58.12\%} & 68.65\% & $-$3.88\% & \phantom{0}4.07\%  \\ 
SQuADv2 $\star$ & 19.06\phantom{\%} & \multicolumn{1}{c}{19.91\phantom{\%}} & 13.46\phantom{\%} & 13.09\phantom{\%} & \multicolumn{1}{c}{11.09\phantom{\%}} & 16.75\phantom{\%} & $-$3.16\phantom{\%} & \phantom{0}3.29\phantom{\%}  \\
SQuADv2 $\square$ & 11.88\phantom{\%} & \multicolumn{1}{c}{11.30\phantom{\%}} & \phantom{0}7.85\phantom{\%} & \phantom{0}8.07\phantom{\%} & \multicolumn{1}{c}{\phantom{0}6.92\phantom{\%}} & \phantom{0}8.84\phantom{\%} & $-$3.04\phantom{\%} & \phantom{0}0.77\phantom{\%} \\
\bottomrule
\end{tabular}
}
\vspace{-0.65cm}
\end{table}

Table~\ref{tbl:acc-gen-main} reports the accuracy performance of \projectname and its backpropagation- and zero-order-based counterparts 
on heterogeneous datasets for million-sized RoBERTa Large, and 
 billion-sized Llama2-7B, OPT6.7B, and OPT13B.
Similar results on personalized performance is shown in Appendix~\ref{adx:additional-results}, Figure~\ref{tbl:acc-pers-adx}.
Results on MultiRC for \textsc{FedMeZO} are absent since the prompt-based finetuning variant of Llama2-7B was unavailable. 
Results on more model architectures and dataset combinations are available in Appendix~\ref{adx:ablation-studies}.
Details on the learning curves, homogeneous dataset splits, and variance across 3 runs are in Appendix \ref{adx:additional-results}.

Overall, \projectname achieves 5.15--13.50\% higher generalized accuracy and 4.87--12.79\% higher personalized accuracy over the best-performing zero-order-based methods across all datasets. 
\textsc{FwdLLM+}, the best-performing zero-order counterpart, attempts to reduce the effect of numeric instability of finite differences by 
(a)~Sampling $K$ perturbations (default $K=10$) per batch for each client and picking 1 perturbation per batch that has the highest cosine similarity with the previous round's aggregated gradients and 
(b)~Only picking trained weights from clients whose computed gradients have variance lower than a set threshold. 
However, we posit that this strategy leads to some clients getting excluded due to a low variance threshold or outlying clients getting included due to a high variance threshold. 
Besides, picking new perturbations based on the previous round's aggregated gradients in the initial rounds can damage the learning trajectory. 
While \textsc{Baffle+} samples more perturbation for each batch to make zero-order finite differences more tractable, the scale of language models demands perturbations on the scale of 500-1000 per batch, which becomes computationally infeasible. 
\textsc{FedMeZO} manages to outperform \textsc{Baffle+} due to its prompt-based finetuning trick but still falls short due to only using 1 perturbation per batch for finite differences on each client.
In contrast, Forward-mode AD used in \projectname avoids the numerical instability from finite differences and improves accuracy by reducing trainable weights assigned to each client. 


Compared to backpropagation-based methods, \textsc{FedAvg} and \textsc{FedYogi}, \projectname manages to come as close as 0.60-6.16\% of generalized accuracy and 2.50--14.12\% of personalized accuracy.
The performance gap between backpropagation and Forward-mode AD arises because in backpropagation, weight updates are more accurate as all gradients are computed directly using the error signal of the objective function. 
In contrast, Forward-mode AD relies on random perturbations, which is relatively less accurate for gradient estimation.
Nonetheless, the advantages of \projectname become evident when we see the peak memory consumption of Forward-more AD compared to backpropagation, which we will discuss next. 

\vspace{-0.2cm} 
\subsection{Peak Memory Consumption Comparison}
\vspace{-0.2cm} 
\label{subsec:peak-memory-consumption}
\begin{wrapfigure}{r}{0.5\textwidth}
  \begin{center}
  \vspace{-1.4cm}
    \includegraphics[width=0.48\textwidth]{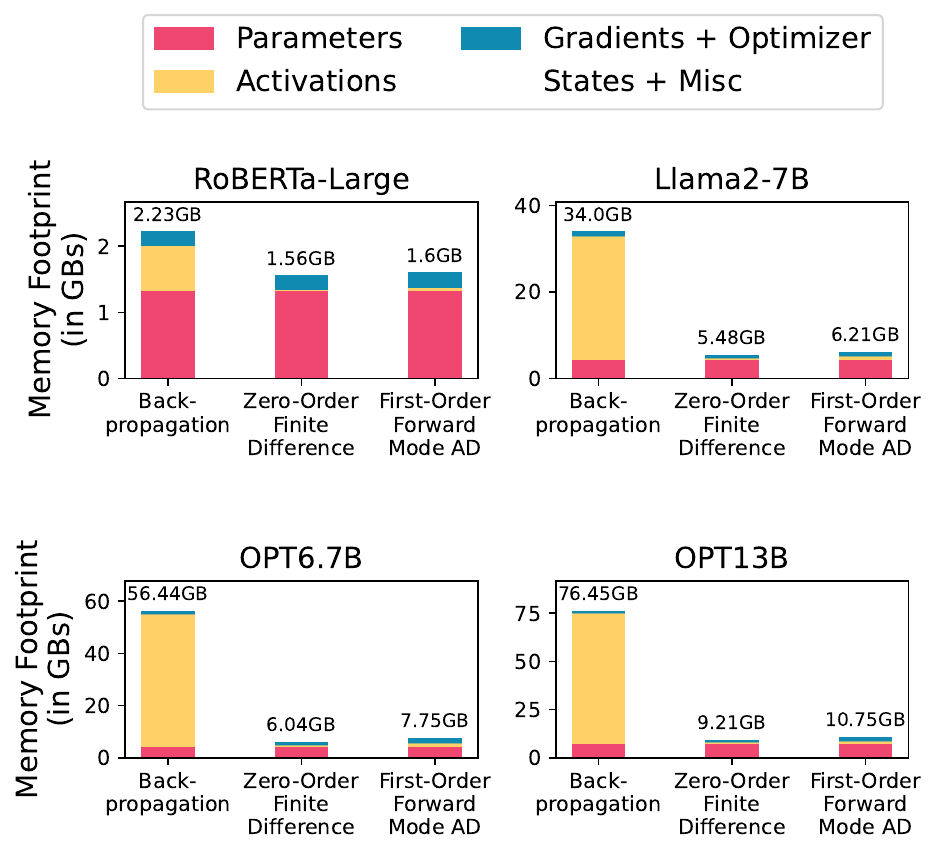}
  \end{center}
  \vspace{-.3cm}
  \caption{Peak memory consumption of \projectname's Forward-mode AD versus backpropgation- and zero-order-based methods. RoBERTa Large, Llama2-7B, and OPT6.7B are profiled with a batch size of 8, and OPT13B with a batch size of 4. \projectname reduces total memory usage by 27.90--86.26\% compared to backpropagation- based methods.
  The 1.54--1.96$\times$ additional memory \projectname uses, compared to zero-order-based methods, is offset by the accuracy gains (\S~\ref{subsec:performance-comparison}).}
  \label{fig:peak-memory-consumption}
  \vspace{-0.6cm}
\end{wrapfigure}
Figure~\ref{fig:peak-memory-consumption} shows peak memory consumption of backpropagation (used in \textsc{FedAvg}, \textsc{FedYogi}), zero-order finite differences (used in \textsc{FwdLLM+}, \textsc{Baffle+}, \textsc{FedMeZO}), and first-order forward mode AD (used in \projectname). 
The methods are profiled for a single client.

Compared to backpropagation, Forward-mode AD reduces peak memory usage of RoBERTa Large by 27.90\%, Llama2-7B by 81.73\%, OPT6.7B by 86.26\% and OPT13B by 85.93\%. 
The sizes of trainable parameter (colored \swatch{239}{71}{111}) and gradient + optimizer state (colored \swatch{17}{138}{178}) are consistent across the 3 modes of computing gradients for all 4 models.
Hence the savings come from the reduced memory footprint related to activations (colored \swatch{255}{209}{102}) in Forward-mode AD.
Compared to backpropagation-based methods, the memory cost related to activations is decreased by 12.12--49.25$\times$ in \projectname.
Unlike storing all the intermediate activations in backpropagation, Forward-mode AD only has to store the previous layer's activation in a forward pass.
The activation footprint of Forward-mode AD is equal to the size of the largest activation.

Against zero-order methods, Forward-mode AD activations cost 1.96$\times$, 1.95$\times$, 1.83$\times$, and 1.54$\times$ more for RoBERTa Large, Llama2-7B, OPT6.7B, and OPT13B respectively.
The increasing cost comes from parallel evaluations of 
(a)~the objective function on the original weights and 
(b)~\texttt{jvp} computation on the perturbations in a single forward pass.
However, as discussed in \S~\ref{subsec:performance-comparison}, the increased memory cost is offset by a boost of up to 13.50\% in accuracy performance.
And as \S~\ref{subsec:wall-clock-time} will discuss, Forward-mode AD reaches convergence faster than zero-order methods since it takes fewer steps to compute a closer gradient estimation.

\vspace{-0.2cm} 
\subsection{Time to Convergence Comparison}
\begin{wrapfigure}{r}{0.40\textwidth}
    \centering
    \vspace{-1.1cm}
    \begin{subfigure}{0.40\textwidth}
    \includegraphics[width=\linewidth]{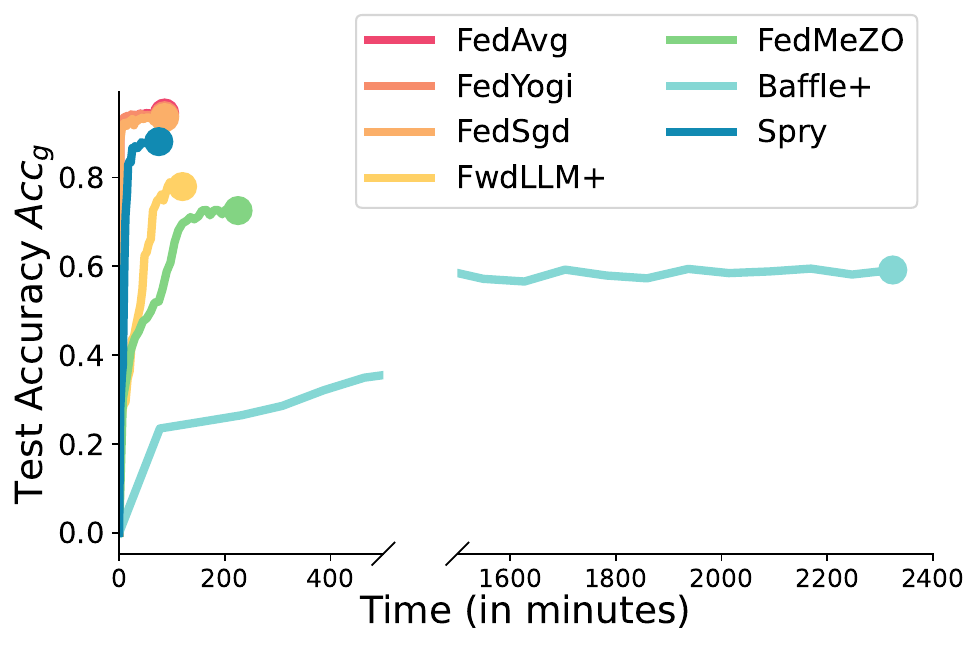}
        \vspace{-4ex}
        \caption{AG News with RoBERTa Large}
    \end{subfigure}
    \rule{0.4\textwidth}{0.5pt}
    \begin{subfigure}{0.40\textwidth}
    \includegraphics[width=\linewidth]{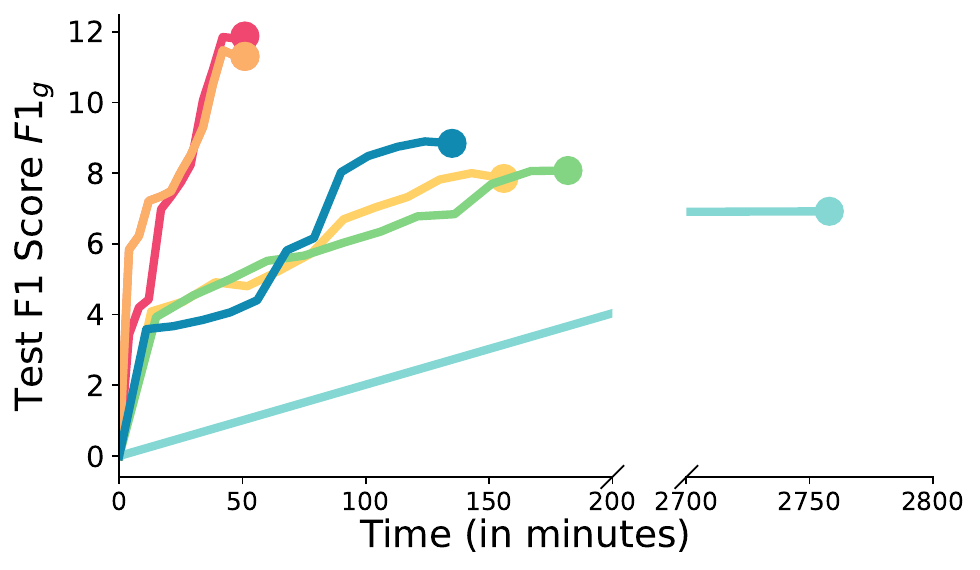}
        \vspace{-4ex}
        \caption{SQuADv2 with OPT13B}
    \end{subfigure}
    \vspace{-1ex}
    \caption{Time to convergence for \projectname and its counterparts. \projectname achieves faster convergence than zero-order methods due to more accurate gradient estimations in a single perturbation.
    }
    \label{fig:wall-clock-time}
    \vspace{-1.3cm}
\end{wrapfigure}

\vspace{-0.18cm} 
\label{subsec:wall-clock-time}
Figure~\ref{fig:wall-clock-time} shows wallclock time-to-convergence for \projectname and its counterparts.
We observe that \projectname is 1.15-1.59$\times$, 6.16-20.28$\times$, and 1.34-2.98$\times$ faster than the zero-order methods \textsc{FwdLLM+}, \textsc{Baffle+}, and \textsc{FedMeZO} respectively. 
For a client in each round, \projectname achieves a faster per-round computation time of 1.46$\times$, 28.57$\times$, and 1.80$\times$ on average, against \textsc{FwdLLM+}, \textsc{Baffle+}, and \textsc{FedMeZO}.
Forward-mode AD achieves faster convergence and faster per-round computation by providing a more accurate gradient estimation through a single perturbation per batch, leading to fewer steps needed to reach convergence.
%
Since each client in \projectname only trains partial weights, it gains a speedup of 1.14$\times$ over backpropagation-based \textsc{FedAvg}, \textsc{FedYogi}, and \textsc{FedSgd} for RoBERTa Large.
However, compared to the backpropagation-based methods, \projectname slows down for billion-sized LMs.
We attribute this loss of speedup to the way \texttt{jvp} is computed in Forward-mode AD. 
\texttt{jvp} is computed column-wise, while its counterpart \texttt{vjp} in backpropagation are computed row-wise.
The column-wise computation incurs time overhead.


\vspace{-0.3cm}
\subsection{Ablation Studies}
\vspace{-0.2cm}
\label{subsec:ablation-studies}
We summarize the ablation experiments on various components of \projectname.
Further discussions are in Appendix~\ref{adx:ablation-studies}. 

\textbf{\projectname can generalize to other language model architectures.} 
Similar to the observations from Table~\ref{tbl:acc-gen-main}, we see a trend of \projectname outperforming the best performing zero-order method \textsc{FwdLLM+} by 3.15-10.25\% for generalized accuracy, demonstrating that \projectname can generalize to other language model architectures.

\textbf{\projectname is compatible with other PEFT methods.}
We integrate \projectname with different \textsc{PEFT} methods like \textsc{IA3}, \textsc{BitFit}, and \textsc{Classifier-Only Finetuning}. 
Results shows that \textsc{LoRA} with \projectname performs the best, with accuracy improvements of 10.60-16.53\%.

\textbf{Effects of the number of trainable weights.} 
We change the number of trainable weights by controlling the rank $r$ and scale $\alpha$ hyperparameters in \textsc{LoRA}. 
Results show that \projectname achieves the highest accuracy with ($r$=1, $\alpha$=1) setting, which has the smallest trainable weight count. The result is consistent with our theoretical analysis in \S\ref{sec:convergence-analysis}. 

\textbf{Effects of communication frequency.} 
Per-iteration communication in \projectname has been shown to boost accuracy by 4.47\% compared to per-epoch communication. This improvement brings the accuracy within 0.92\% and 0.96\% of \textsc{FedAvg} and \textsc{FedSgd}, respectively.

\textbf{Effects of perturbation count per batch.}
We observe that increasing the number of perturbations per batch ($K$) for Forward-mode AD has little to no impact on the end prediction performance of \projectname, with $K=100$ improving the generalized test accuracy by 1.1\% over $K=1$. 
The benefits of increasing $K$ are however seen in the convergence speed. 
Setting $K=10$ achieves a steady state (of accuracy $\sim$86\%) around the 200\textsuperscript{th} round, while the setting with $K=1$ takes 500 rounds.

\textbf{Effects of participating client count.} 
Increasing the client count increases the prediction performance of \projectname.
For the SST2 dataset, with the total client count fixed to 100, the three settings $C=10$, $C=50$, and $C=100$ produce accuracies of $85.14\%$, $86.56\%$, and $88.08\%$, respectively.
We also see an improvement in the convergence speed as the participating client count increases.
To achieve an accuracy of $\sim$85\%; $C=10$, $C=50$, $C=100$ require 500, 450, and 150 rounds, respectively.

\textbf{Importance of splitting weights.}
To understand the effects of splitting, we conduct the following two experiments:
(a)~With \textsc{FedAvgSplit}, we apply the strategy of splitting trainable layers across clients (see \S~\ref{subsubsec:splitting-layers}) to backpropagation-based \textsc{FedAvg}, and 
(b)~With \textsc{FedFgd}, we omit the splitting strategy of \projectname for FL with Forward-mode AD.
We observe that \textsc{FedAvgSplit} fails to achieve similar accuracy to \textsc{FedAvg} with a drop of 2.60-10.00\%.
\textsc{FedFgd} fails to converge as the size of trainable weights increases, e.g., with RoBERTa Large with \textsc{LoRA}, that has 1.15M trainable weights.
This proves the necessity of splitting trainable weights for Forward-mode AD in \projectname. 

\vspace{-0.2cm} 
\subsection{Communication and Computation Costs}
\label{subsec:comm-and-comp-costs}
Tables~\ref{tbl:communication-cost} and~\ref{tbl:computation-cost} in Appendix~\ref{adx:communication-and-computation-costs} shows the communication and computation overhead of \projectname against all its baselines. 
We summarize the main observations here:\\\\
\textbf{\projectname has lower communication cost due to the splitting strategy and scalar \texttt{jvp}.  }
Suppose $w_g$ is the total trainable parameter count of a model to be trained in federated setting.
The backpropagation-based baselines; \textsc{FedAvg} (per-epoch communication), \textsc{FedSgd} (per-iteration communication), and \textsc{FedYogi} (per-epoch communication) transmit the entire set of trainable parameters to each participating client, and receives the same from each participating client.
That results in ``client to server'' communication cost of $w_g$ and ``server to client'' communication cost of $w_g \times M$.
The per-epoch versions of the zero-order baselines (\textsc{FedMeZO}, \textsc{Baffle}, \textsc{FwdLLM}) follow a similar logic due to all the parameters needing to be transmitted to each client, with ``client to server'' communication costing $w_g$, and ``server to client'' communication taking  the cost of $w_g \times M$. 
However, the per-iteration versions of the zero-order baselines fare better, with ``client to server'' communication only requiring each client sending a scalar finite difference (cost of 1), and ``server to client'' communication accruing $(w_g + 1)\times M$ cost, due to the server also needing to send a scalar seed to each client now.

Meanwhile, \projectname only needs to send $\max\left(\frac{L}{M}, 1\right)$ layers (where $L$ is the total layer count of a model, and $M$ is the participating count of clients for a round), each layer of size $w_\ell$.
Hence, for per-epoch ``client to server'' communication accrues $w_\ell \max\left(\frac{L}{M}, 1\right)$, which is smaller than $w_g$.
Similarly, per-epoch ``server to client'' communication costs $w_\ell M \max\left(\frac{L}{M}, 1\right)$, which is also smaller than the cost of $w_g M$ related to the baselines. 
For per-iteration communication, clients only need to send a scalar \texttt{jvp} (cost of 1) to the server; this matches the communication cost of per-iteration zero-order methods.
Server needs to send a total of $w_\ell M \max(L, M)$ (derivation given in Table~\ref{tbl:communication-cost}), which is a smaller cost than the costs of backpropagation and zero-order methods.

\textbf{\projectname accrues lower computation cost due to lower trainable parameter count.  }
\projectname{}’s client-side computation cost is traded off by a faster convergence to higher accuracy through better gradient approximations compared to finite difference-based methods.
And \projectname is the least computationally expensive on the server side due to needing to aggregate fewer parameters from the clients.

Let's assume that matrix multiplication costs $c$ for each layer, resulting in a forward pass cost of $c$. 
The cost of backpropagation is $2c$ because the computation of the current layer’s weight gradient is $c$, and the cost of computing the previous layer’s activation gradient is another $c$. 
\texttt{jvp} computation in \projectname takes additional cost of $c$ for each layer. 
Moreover, since \texttt{jvp} calculation happens through column-by-column vector multiplications, the related overhead is quantified by $v$.

Hence backpropagation-based methods \textsc{FedAvg}, \textsc{FedSgd}, and \textsc{FedYogi} computationally costs $3Lc$ at client, and costs $w_g(M-1)$ at server (due to additions).
Note that $L$ amounts to all layers in the model here.
\textsc{FedMeZO} costs $L(2c + 3w_\ell)$ at client through two forward passes, and generating perturbations three times. 
\textsc{FwdLLM} and \textsc{Baffle} costs $KL (2c + w_\ell)$ due to $K$ perturbations for all $L$ layers, with two forward passes and generation perturbations once.
Against that, \projectname costs $2 \max(\frac{L}{M}, 1) (c + v) + w_\ell L$ for a smaller count of $L$, traded-off by the \texttt{jvp} computation cost of $v$.

On the server side, \projectname is the least computationally demanding. 
\projectname needs to aggregate a subset of layer weights from only the clients that were assigned to those layers.
Computation cost on the server-side changes based on the communication frequency per-iteration communication incurs an additional overhead of $w_\ell L (\frac{M}{L} + 1)$ and $w_\ell L (M + 1)$ (generation of perturbations at the server-side, and multiplying those perturbations with aggregate of the \texttt{jvp} values received from the clients) for \projectname and its zero-order counterparts respectively.

\vspace{-0.3cm}
\section{Conclusion}
\vspace{-0.3cm}
\label{sec:conclusion}
\projectname enables finetuning medium and large language models in cross-device FL. 
It introduces a training strategy where trainable weights are split across federated clients, so each client only applies Forward-mode AD to a fraction of the weights. 
This approach significantly reduces the memory footprint compared to backpropagation and achieves better gradient estimation, resulting in higher accuracy and faster convergence than zero-order methods. 
Experiments on various language tasks and models demonstrate \projectname's effectiveness in reducing memory usage while maintaining accuracy comparable to backpropagation.
We formally prove that the estimation bias of the global forward gradients in \projectname depends on data heterogeneity across clients.
We also analyzed how the convergence rate of \projectname relates to the configurations of \projectname and FL settings including properties of weight perturbations, data heterogeneity, and the number of clients and FL rounds.

\vspace{-0.2cm}
\begin{ack}
\vspace{-0.2cm}
This material is based upon work supported by the National Science Foundation under grant no.\ CCF-2210243, DMS-2220211, CNS-2224054, CNS-2312396, and CNS-2338512, 
and by Adobe.
\end{ack}

{
\small
\bibliographystyle{unsrtnat}
\bibliography{bibliography}

\begin{thebibliography}{58}
\providecommand{\natexlab}[1]{#1}
\providecommand{\url}[1]{\texttt{#1}}
\expandafter\ifx\csname urlstyle\endcsname\relax
  \providecommand{\doi}[1]{doi: #1}\else
  \providecommand{\doi}{doi: \begingroup \urlstyle{rm}\Url}\fi

\bibitem[McMahan et~al.(2017)McMahan, Moore, Ramage, Hampson, and Arcas]{macmahan2017aFedLearning}
Brendan McMahan, Eider Moore, Daniel Ramage, Seth Hampson, and Blaise Aguera~y Arcas.
\newblock {Communication-Efficient Learning of Deep Networks from Decentralized Data}.
\newblock In \emph{Proceedings of the 20th International Conference on Artificial Intelligence and Statistics}, Proceedings of Machine Learning Research. PMLR, 2017.

\bibitem[Kairouz et~al.(2021)Kairouz, McMahan, Avent, Bellet, Bennis, Bhagoji, Bonawitz, Charles, Cormode, Cummings, D'Oliveira, Eichner, Rouayheb, Evans, Gardner, Garrett, Gascón, Ghazi, Gibbons, Gruteser, Harchaoui, He, He, Huo, Hutchinson, Hsu, Jaggi, Javidi, Joshi, Khodak, Konecný, Korolova, Koushanfar, Koyejo, Lepoint, Liu, Mittal, Mohri, Nock, Özgür, Pagh, Qi, Ramage, Raskar, Raykova, Song, Song, Stich, Sun, Suresh, Tramèr, Vepakomma, Wang, Xiong, Xu, Yang, Yu, Yu, and Zhao]{kairouz2021open}
Peter Kairouz, H.~Brendan McMahan, Brendan Avent, Aurélien Bellet, Mehdi Bennis, Arjun~Nitin Bhagoji, Kallista~A. Bonawitz, Zachary Charles, Graham Cormode, Rachel Cummings, Rafael G.~L. D'Oliveira, Hubert Eichner, Salim~El Rouayheb, David Evans, Josh Gardner, Zachary Garrett, Adrià Gascón, Badih Ghazi, Phillip~B. Gibbons, Marco Gruteser, Zaïd Harchaoui, Chaoyang He, Lie He, Zhouyuan Huo, Ben Hutchinson, Justin Hsu, Martin Jaggi, Tara Javidi, Gauri Joshi, Mikhail Khodak, Jakub Konecný, Aleksandra Korolova, Farinaz Koushanfar, Sanmi Koyejo, Tancrède Lepoint, Yang Liu, Prateek Mittal, Mehryar Mohri, Richard Nock, Ayfer Özgür, Rasmus Pagh, Hang Qi, Daniel Ramage, Ramesh Raskar, Mariana Raykova, Dawn Song, Weikang Song, Sebastian~U. Stich, Ziteng Sun, Ananda~Theertha Suresh, Florian Tramèr, Praneeth Vepakomma, Jianyu Wang, Li~Xiong, Zheng Xu, Qiang Yang, Felix~X. Yu, Han Yu, and Sen Zhao.
\newblock Advances and open problems in federated learning.
\newblock \emph{Foundations and Trends in Machine Learning}, 2021.

\bibitem[Loftus et~al.(2022)Loftus, Ruppert, Shickel, Ozrazgat-Baslanti, Balch, Efron, Gilbert R~Upchurch, Rashidi, Tignanelli, Bian, and Bihorac]{loftus2022FLHealthcare}
Tyler~J Loftus, Matthew~M Ruppert, Benjamin Shickel, Tezcan Ozrazgat-Baslanti, Jeremy~A Balch, Philip~A Efron, Jr. Gilbert R~Upchurch, Parisa Rashidi, Christopher Tignanelli, Jiang Bian, and Azra Bihorac.
\newblock Federated learning for preserving data privacy in collaborative healthcare research.
\newblock \emph{Digital Health}, 2022.

\bibitem[Li et~al.(2022)Li, Yu, Cai, He, and Liu]{li2022healthiot}
Li~Li, Xi~Yu, Xuliang Cai, Xin He, and Yanhong Liu.
\newblock Contract theory based incentive mechanism for federated learning in health crowdsensing.
\newblock \emph{IEEE Internet of Things Journal}, 2022.

\bibitem[Dara et~al.(2022)Dara, Kanapala, Babu, Dhamercherala, Vidyarthi, and Agarwal]{dara2022FLiot}
Suresh Dara, Ambedkar Kanapala, A.~Ramesh Babu, Swetha Dhamercherala, Ankit Vidyarthi, and Ruchi Agarwal.
\newblock Scalable federated-learning and internet-of-things enabled architecture for chest computer tomography image classification.
\newblock \emph{Computers and Electrical Engineering}, 2022.

\bibitem[Pinelli et~al.(2023)Pinelli, Tolomei, and Trappolini]{pinelli2023flirt}
Fabio Pinelli, Gabriele Tolomei, and Giovanni Trappolini.
\newblock Flirt: Federated learning for information retrieval.
\newblock In \emph{Proceedings of the 46th International ACM SIGIR Conference on Research and Development in Information Retrieval}, 2023.

\bibitem[OpenAI et~al.(2024)OpenAI, Achiam, Adler, Agarwal, Ahmad, Akkaya, Aleman, Almeida, Altenschmidt, Altman, Anadkat, Avila, Babuschkin, Balaji, Balcom, Baltescu, Bao, Bavarian, Belgum, Bello, Berdine, Bernadett-Shapiro, Berner, Bogdonoff, Boiko, Boyd, Brakman, Brockman, Brooks, Brundage, Button, Cai, Campbell, Cann, Carey, Carlson, Carmichael, Chan, Chang, Chantzis, Chen, Chen, Chen, Chen, Chen, Chess, Cho, Chu, Chung, Cummings, Currier, Dai, Decareaux, Degry, Deutsch, Deville, Dhar, Dohan, Dowling, Dunning, Ecoffet, Eleti, Eloundou, Farhi, Fedus, Felix, Fishman, Forte, Fulford, Gao, Georges, Gibson, Goel, Gogineni, Goh, Gontijo-Lopes, Gordon, Grafstein, Gray, Greene, Gross, Gu, Guo, Hallacy, Han, Harris, He, Heaton, Heidecke, Hesse, Hickey, Hickey, Hoeschele, Houghton, Hsu, Hu, Hu, Huizinga, Jain, Jain, Jang, Jiang, Jiang, Jin, Jin, Jomoto, Jonn, Jun, Kaftan, Łukasz Kaiser, Kamali, Kanitscheider, Keskar, Khan, Kilpatrick, Kim, Kim, Kim, Kirchner, Kiros, Knight, Kokotajlo, Łukasz Kondraciuk, Kondrich,
  Konstantinidis, Kosic, Krueger, Kuo, Lampe, Lan, Lee, Leike, Leung, Levy, Li, Lim, Lin, Lin, Litwin, Lopez, Lowe, Lue, Makanju, Malfacini, Manning, Markov, Markovski, Martin, Mayer, Mayne, McGrew, McKinney, McLeavey, McMillan, McNeil, Medina, Mehta, Menick, Metz, Mishchenko, Mishkin, Monaco, Morikawa, Mossing, Mu, Murati, Murk, Mély, Nair, Nakano, Nayak, Neelakantan, Ngo, Noh, Ouyang, O'Keefe, Pachocki, Paino, Palermo, Pantuliano, Parascandolo, Parish, Parparita, Passos, Pavlov, Peng, Perelman, de~Avila Belbute~Peres, Petrov, de~Oliveira~Pinto, Michael, Pokorny, Pokrass, Pong, Powell, Power, Power, Proehl, Puri, Radford, Rae, Ramesh, Raymond, Real, Rimbach, Ross, Rotsted, Roussez, Ryder, Saltarelli, Sanders, Santurkar, Sastry, Schmidt, Schnurr, Schulman, Selsam, Sheppard, Sherbakov, Shieh, Shoker, Shyam, Sidor, Sigler, Simens, Sitkin, Slama, Sohl, Sokolowsky, Song, Staudacher, Such, Summers, Sutskever, Tang, Tezak, Thompson, Tillet, Tootoonchian, Tseng, Tuggle, Turley, Tworek, Uribe, Vallone, Vijayvergiya,
  Voss, Wainwright, Wang, Wang, Wang, Ward, Wei, Weinmann, Welihinda, Welinder, Weng, Weng, Wiethoff, Willner, Winter, Wolrich, Wong, Workman, Wu, Wu, Wu, Xiao, Xu, Yoo, Yu, Yuan, Zaremba, Zellers, Zhang, Zhang, Zhao, Zheng, Zhuang, Zhuk, and Zoph]{openai2024gpt4}
OpenAI, Josh Achiam, Steven Adler, Sandhini Agarwal, Lama Ahmad, Ilge Akkaya, Florencia~Leoni Aleman, Diogo Almeida, Janko Altenschmidt, Sam Altman, Shyamal Anadkat, Red Avila, Igor Babuschkin, Suchir Balaji, Valerie Balcom, Paul Baltescu, Haiming Bao, Mohammad Bavarian, Jeff Belgum, Irwan Bello, Jake Berdine, Gabriel Bernadett-Shapiro, Christopher Berner, Lenny Bogdonoff, Oleg Boiko, Madelaine Boyd, Anna-Luisa Brakman, Greg Brockman, Tim Brooks, Miles Brundage, Kevin Button, Trevor Cai, Rosie Campbell, Andrew Cann, Brittany Carey, Chelsea Carlson, Rory Carmichael, Brooke Chan, Che Chang, Fotis Chantzis, Derek Chen, Sully Chen, Ruby Chen, Jason Chen, Mark Chen, Ben Chess, Chester Cho, Casey Chu, Hyung~Won Chung, Dave Cummings, Jeremiah Currier, Yunxing Dai, Cory Decareaux, Thomas Degry, Noah Deutsch, Damien Deville, Arka Dhar, David Dohan, Steve Dowling, Sheila Dunning, Adrien Ecoffet, Atty Eleti, Tyna Eloundou, David Farhi, Liam Fedus, Niko Felix, Simón~Posada Fishman, Juston Forte, Isabella Fulford, Leo
  Gao, Elie Georges, Christian Gibson, Vik Goel, Tarun Gogineni, Gabriel Goh, Rapha Gontijo-Lopes, Jonathan Gordon, Morgan Grafstein, Scott Gray, Ryan Greene, Joshua Gross, Shixiang~Shane Gu, Yufei Guo, Chris Hallacy, Jesse Han, Jeff Harris, Yuchen He, Mike Heaton, Johannes Heidecke, Chris Hesse, Alan Hickey, Wade Hickey, Peter Hoeschele, Brandon Houghton, Kenny Hsu, Shengli Hu, Xin Hu, Joost Huizinga, Shantanu Jain, Shawn Jain, Joanne Jang, Angela Jiang, Roger Jiang, Haozhun Jin, Denny Jin, Shino Jomoto, Billie Jonn, Heewoo Jun, Tomer Kaftan, Łukasz Kaiser, Ali Kamali, Ingmar Kanitscheider, Nitish~Shirish Keskar, Tabarak Khan, Logan Kilpatrick, Jong~Wook Kim, Christina Kim, Yongjik Kim, Jan~Hendrik Kirchner, Jamie Kiros, Matt Knight, Daniel Kokotajlo, Łukasz Kondraciuk, Andrew Kondrich, Aris Konstantinidis, Kyle Kosic, Gretchen Krueger, Vishal Kuo, Michael Lampe, Ikai Lan, Teddy Lee, Jan Leike, Jade Leung, Daniel Levy, Chak~Ming Li, Rachel Lim, Molly Lin, Stephanie Lin, Mateusz Litwin, Theresa Lopez, Ryan
  Lowe, Patricia Lue, Anna Makanju, Kim Malfacini, Sam Manning, Todor Markov, Yaniv Markovski, Bianca Martin, Katie Mayer, Andrew Mayne, Bob McGrew, Scott~Mayer McKinney, Christine McLeavey, Paul McMillan, Jake McNeil, David Medina, Aalok Mehta, Jacob Menick, Luke Metz, Andrey Mishchenko, Pamela Mishkin, Vinnie Monaco, Evan Morikawa, Daniel Mossing, Tong Mu, Mira Murati, Oleg Murk, David Mély, Ashvin Nair, Reiichiro Nakano, Rajeev Nayak, Arvind Neelakantan, Richard Ngo, Hyeonwoo Noh, Long Ouyang, Cullen O'Keefe, Jakub Pachocki, Alex Paino, Joe Palermo, Ashley Pantuliano, Giambattista Parascandolo, Joel Parish, Emy Parparita, Alex Passos, Mikhail Pavlov, Andrew Peng, Adam Perelman, Filipe de~Avila Belbute~Peres, Michael Petrov, Henrique~Ponde de~Oliveira~Pinto, Michael, Pokorny, Michelle Pokrass, Vitchyr~H. Pong, Tolly Powell, Alethea Power, Boris Power, Elizabeth Proehl, Raul Puri, Alec Radford, Jack Rae, Aditya Ramesh, Cameron Raymond, Francis Real, Kendra Rimbach, Carl Ross, Bob Rotsted, Henri Roussez,
  Nick Ryder, Mario Saltarelli, Ted Sanders, Shibani Santurkar, Girish Sastry, Heather Schmidt, David Schnurr, John Schulman, Daniel Selsam, Kyla Sheppard, Toki Sherbakov, Jessica Shieh, Sarah Shoker, Pranav Shyam, Szymon Sidor, Eric Sigler, Maddie Simens, Jordan Sitkin, Katarina Slama, Ian Sohl, Benjamin Sokolowsky, Yang Song, Natalie Staudacher, Felipe~Petroski Such, Natalie Summers, Ilya Sutskever, Jie Tang, Nikolas Tezak, Madeleine~B. Thompson, Phil Tillet, Amin Tootoonchian, Elizabeth Tseng, Preston Tuggle, Nick Turley, Jerry Tworek, Juan Felipe~Cerón Uribe, Andrea Vallone, Arun Vijayvergiya, Chelsea Voss, Carroll Wainwright, Justin~Jay Wang, Alvin Wang, Ben Wang, Jonathan Ward, Jason Wei, CJ~Weinmann, Akila Welihinda, Peter Welinder, Jiayi Weng, Lilian Weng, Matt Wiethoff, Dave Willner, Clemens Winter, Samuel Wolrich, Hannah Wong, Lauren Workman, Sherwin Wu, Jeff Wu, Michael Wu, Kai Xiao, Tao Xu, Sarah Yoo, Kevin Yu, Qiming Yuan, Wojciech Zaremba, Rowan Zellers, Chong Zhang, Marvin Zhang, Shengjia
  Zhao, Tianhao Zheng, Juntang Zhuang, William Zhuk, and Barret Zoph.
\newblock Gpt-4 technical report.
\newblock \emph{arXiv 2303.08774}, 2024.

\bibitem[Anil et~al.(2023)Anil, Dai, Firat, Johnson, Lepikhin, Passos, Shakeri, Taropa, Bailey, Chen, Chu, Clark, Shafey, Huang, Meier-Hellstern, Mishra, Moreira, Omernick, Robinson, Ruder, Tay, Xiao, Xu, Zhang, Abrego, Ahn, Austin, Barham, Botha, Bradbury, Brahma, Brooks, Catasta, Cheng, Cherry, Choquette-Choo, Chowdhery, Crepy, Dave, Dehghani, Dev, Devlin, Díaz, Du, Dyer, Feinberg, Feng, Fienber, Freitag, Garcia, Gehrmann, Gonzalez, Gur-Ari, Hand, Hashemi, Hou, Howland, Hu, Hui, Hurwitz, Isard, Ittycheriah, Jagielski, Jia, Kenealy, Krikun, Kudugunta, Lan, Lee, Lee, Li, Li, Li, Li, Li, Lim, Lin, Liu, Liu, Maggioni, Mahendru, Maynez, Misra, Moussalem, Nado, Nham, Ni, Nystrom, Parrish, Pellat, Polacek, Polozov, Pope, Qiao, Reif, Richter, Riley, Ros, Roy, Saeta, Samuel, Shelby, Slone, Smilkov, So, Sohn, Tokumine, Valter, Vasudevan, Vodrahalli, Wang, Wang, Wang, Wang, Wieting, Wu, Xu, Xu, Xue, Yin, Yu, Zhang, Zheng, Zheng, Zhou, Zhou, Petrov, and Wu]{anil2023palm}
Rohan Anil, Andrew~M. Dai, Orhan Firat, Melvin Johnson, Dmitry Lepikhin, Alexandre Passos, Siamak Shakeri, Emanuel Taropa, Paige Bailey, Zhifeng Chen, Eric Chu, Jonathan~H. Clark, Laurent~El Shafey, Yanping Huang, Kathy Meier-Hellstern, Gaurav Mishra, Erica Moreira, Mark Omernick, Kevin Robinson, Sebastian Ruder, Yi~Tay, Kefan Xiao, Yuanzhong Xu, Yujing Zhang, Gustavo~Hernandez Abrego, Junwhan Ahn, Jacob Austin, Paul Barham, Jan Botha, James Bradbury, Siddhartha Brahma, Kevin Brooks, Michele Catasta, Yong Cheng, Colin Cherry, Christopher~A. Choquette-Choo, Aakanksha Chowdhery, Clément Crepy, Shachi Dave, Mostafa Dehghani, Sunipa Dev, Jacob Devlin, Mark Díaz, Nan Du, Ethan Dyer, Vlad Feinberg, Fangxiaoyu Feng, Vlad Fienber, Markus Freitag, Xavier Garcia, Sebastian Gehrmann, Lucas Gonzalez, Guy Gur-Ari, Steven Hand, Hadi Hashemi, Le~Hou, Joshua Howland, Andrea Hu, Jeffrey Hui, Jeremy Hurwitz, Michael Isard, Abe Ittycheriah, Matthew Jagielski, Wenhao Jia, Kathleen Kenealy, Maxim Krikun, Sneha Kudugunta, Chang
  Lan, Katherine Lee, Benjamin Lee, Eric Li, Music Li, Wei Li, YaGuang Li, Jian Li, Hyeontaek Lim, Hanzhao Lin, Zhongtao Liu, Frederick Liu, Marcello Maggioni, Aroma Mahendru, Joshua Maynez, Vedant Misra, Maysam Moussalem, Zachary Nado, John Nham, Eric Ni, Andrew Nystrom, Alicia Parrish, Marie Pellat, Martin Polacek, Alex Polozov, Reiner Pope, Siyuan Qiao, Emily Reif, Bryan Richter, Parker Riley, Alex~Castro Ros, Aurko Roy, Brennan Saeta, Rajkumar Samuel, Renee Shelby, Ambrose Slone, Daniel Smilkov, David~R. So, Daniel Sohn, Simon Tokumine, Dasha Valter, Vijay Vasudevan, Kiran Vodrahalli, Xuezhi Wang, Pidong Wang, Zirui Wang, Tao Wang, John Wieting, Yuhuai Wu, Kelvin Xu, Yunhan Xu, Linting Xue, Pengcheng Yin, Jiahui Yu, Qiao Zhang, Steven Zheng, Ce~Zheng, Weikang Zhou, Denny Zhou, Slav Petrov, and Yonghui Wu.
\newblock Palm 2 technical report.
\newblock \emph{arXiv 2305.10403}, 2023.

\bibitem[Lin et~al.(2022)Lin, He, Ze, Wang, Hua, Dupuy, Gupta, Soltanolkotabi, Ren, and Avestimehr]{lin2022fednlp}
Bill~Yuchen Lin, Chaoyang He, Zihang Ze, Hulin Wang, Yufen Hua, Christophe Dupuy, Rahul Gupta, Mahdi Soltanolkotabi, Xiang Ren, and Salman Avestimehr.
\newblock {F}ed{NLP}: Benchmarking federated learning methods for natural language processing tasks.
\newblock In \emph{Findings of the Association for Computational Linguistics: NAACL 2022}. Association for Computational Linguistics, 2022.

\bibitem[Tian et~al.(2022)Tian, Wan, Lyu, Yao, Jin, and Sun]{tian2022fedbert}
Yuanyishu Tian, Yao Wan, Lingjuan Lyu, Dezhong Yao, Hai Jin, and Lichao Sun.
\newblock Fedbert: When federated learning meets pre-training.
\newblock \emph{ACM Transactions on Intelligent Systems and Technology (TIST)}, 2022.

\bibitem[Borzunov et~al.(2024)Borzunov, Ryabinin, Chumachenko, Baranchuk, Dettmers, Belkada, Samygin, and Raffel]{borzunov2024distributed}
Alexander Borzunov, Max Ryabinin, Artem Chumachenko, Dmitry Baranchuk, Tim Dettmers, Younes Belkada, Pavel Samygin, and Colin~A Raffel.
\newblock Distributed inference and fine-tuning of large language models over the internet.
\newblock \emph{Advances in Neural Information Processing Systems}, 2024.

\bibitem[Hu et~al.(2022)Hu, yelong shen, Wallis, Allen-Zhu, Li, Wang, Wang, and Chen]{hu2022lora}
Edward~J Hu, yelong shen, Phillip Wallis, Zeyuan Allen-Zhu, Yuanzhi Li, Shean Wang, Lu~Wang, and Weizhu Chen.
\newblock Lo{RA}: Low-rank adaptation of large language models.
\newblock In \emph{International Conference on Learning Representations}, 2022.

\bibitem[Cai et~al.(2023)Cai, Wu, Wang, and Xu]{cai2023fedadapter}
Dongqi Cai, Yaozong Wu, Shangguang Wang, and Mengwei Xu.
\newblock Fedadapter: Efficient federated learning for mobile nlp.
\newblock In \emph{Proceedings of the ACM Turing Award Celebration Conference - China 2023}. Association for Computing Machinery, 2023.

\bibitem[Qiu et~al.(2023)Qiu, Liu, Feng, Xue, Feng, Liu, Zhang, Weller, and Sch{\"o}lkopf]{qiu2023oft}
Zeju Qiu, Weiyang Liu, Haiwen Feng, Yuxuan Xue, Yao Feng, Zhen Liu, Dan Zhang, Adrian Weller, and Bernhard Sch{\"o}lkopf.
\newblock Controlling text-to-image diffusion by orthogonal finetuning.
\newblock In \emph{Thirty-seventh Conference on Neural Information Processing Systems}, 2023.

\bibitem[Liu et~al.(2022)Liu, Tam, Muqeeth, Mohta, Huang, Bansal, and Raffel]{liu2022ia3}
Haokun Liu, Derek Tam, Mohammed Muqeeth, Jay Mohta, Tenghao Huang, Mohit Bansal, and Colin~A Raffel.
\newblock Few-shot parameter-efficient fine-tuning is better and cheaper than in-context learning.
\newblock In \emph{Advances in Neural Information Processing Systems}, 2022.

\bibitem[Li and Liang(2021)]{li2021prefixtuning}
Xiang~Lisa Li and Percy Liang.
\newblock Prefix-tuning: Optimizing continuous prompts for generation.
\newblock In \emph{Proceedings of the 59th Annual Meeting of the Association for Computational Linguistics and the 11th International Joint Conference on Natural Language Processing (Volume 1: Long Papers)}. Association for Computational Linguistics, 2021.

\bibitem[Jacob et~al.(2018)Jacob, Kligys, Chen, Zhu, Tang, Howard, Adam, and Kalenichenko]{jacob2018quant}
Benoit Jacob, Skirmantas Kligys, Bo~Chen, Menglong Zhu, Matthew Tang, Andrew Howard, Hartwig Adam, and Dmitry Kalenichenko.
\newblock Quantization and training of neural networks for efficient integer-arithmetic-only inference.
\newblock In \emph{Proceedings of the IEEE Conference on Computer Vision and Pattern Recognition (CVPR)}, 2018.

\bibitem[Malladi et~al.(2023)Malladi, Gao, Nichani, Damian, Lee, Chen, and Arora]{malladi2023mezo}
Sadhika Malladi, Tianyu Gao, Eshaan Nichani, Alex Damian, Jason~D. Lee, Danqi Chen, and Sanjeev Arora.
\newblock Fine-tuning language models with just forward passes.
\newblock In \emph{Thirty-seventh Conference on Neural Information Processing Systems}, 2023.

\bibitem[Xu et~al.(2024)Xu, Cai, Wu, Li, and Wang]{xu2024fwdllm}
Mengwei Xu, Dongqi Cai, Yaozong Wu, Xiang Li, and Shangguang Wang.
\newblock Fwdllm: Efficient fedllm using forward gradient.
\newblock \emph{arXiv 2308.13894}, 2024.

\bibitem[Feng et~al.(2023)Feng, Pang, Du, Chen, Yan, and Lin]{feng2023baffle}
Haozhe Feng, Tianyu Pang, Chao Du, Wei Chen, Shuicheng Yan, and Min Lin.
\newblock Does federated learning really need backpropagation?
\newblock \emph{arXiv 2301.12195}, 2023.

\bibitem[Touvron et~al.(2023)Touvron, Martin, Stone, Albert, Almahairi, Babaei, Bashlykov, Batra, Bhargava, Bhosale, Bikel, Blecher, Ferrer, Chen, Cucurull, Esiobu, Fernandes, Fu, Fu, Fuller, Gao, Goswami, Goyal, Hartshorn, Hosseini, Hou, Inan, Kardas, Kerkez, Khabsa, Kloumann, Korenev, Koura, Lachaux, Lavril, Lee, Liskovich, Lu, Mao, Martinet, Mihaylov, Mishra, Molybog, Nie, Poulton, Reizenstein, Rungta, Saladi, Schelten, Silva, Smith, Subramanian, Tan, Tang, Taylor, Williams, Kuan, Xu, Yan, Zarov, Zhang, Fan, Kambadur, Narang, Rodriguez, Stojnic, Edunov, and Scialom]{touvron2023llama}
Hugo Touvron, Louis Martin, Kevin Stone, Peter Albert, Amjad Almahairi, Yasmine Babaei, Nikolay Bashlykov, Soumya Batra, Prajjwal Bhargava, Shruti Bhosale, Dan Bikel, Lukas Blecher, Cristian~Canton Ferrer, Moya Chen, Guillem Cucurull, David Esiobu, Jude Fernandes, Jeremy Fu, Wenyin Fu, Brian Fuller, Cynthia Gao, Vedanuj Goswami, Naman Goyal, Anthony Hartshorn, Saghar Hosseini, Rui Hou, Hakan Inan, Marcin Kardas, Viktor Kerkez, Madian Khabsa, Isabel Kloumann, Artem Korenev, Punit~Singh Koura, Marie-Anne Lachaux, Thibaut Lavril, Jenya Lee, Diana Liskovich, Yinghai Lu, Yuning Mao, Xavier Martinet, Todor Mihaylov, Pushkar Mishra, Igor Molybog, Yixin Nie, Andrew Poulton, Jeremy Reizenstein, Rashi Rungta, Kalyan Saladi, Alan Schelten, Ruan Silva, Eric~Michael Smith, Ranjan Subramanian, Xiaoqing~Ellen Tan, Binh Tang, Ross Taylor, Adina Williams, Jian~Xiang Kuan, Puxin Xu, Zheng Yan, Iliyan Zarov, Yuchen Zhang, Angela Fan, Melanie Kambadur, Sharan Narang, Aurelien Rodriguez, Robert Stojnic, Sergey Edunov, and Thomas
  Scialom.
\newblock Llama 2: Open foundation and fine-tuned chat models.
\newblock \emph{arXiv 2307.09288}, 2023.

\bibitem[Richardson(1955)]{richardson1955finitedifferences}
C.~H. Richardson.
\newblock An introduction to the calculus of finite differences. by c.h. richardson pp. vi, 142. 28s. 1954. (van nostrand, new york; macmillan, london).
\newblock \emph{The Mathematical Gazette}, 39\penalty0 (330), 1955.
\newblock \doi{10.2307/3608616}.

\bibitem[Baydin et~al.(2022)Baydin, Pearlmutter, Syme, Wood, and Torr]{baydin2022gradients}
Atılım~Güneş Baydin, Barak~A. Pearlmutter, Don Syme, Frank Wood, and Philip Torr.
\newblock Gradients without backpropagation.
\newblock \emph{arXiv 2202.08587}, 2022.

\bibitem[Baydin et~al.(2017)Baydin, Pearlmutter, Radul, and Siskind]{baydin2017autodiff}
At\i{}l\i{}m~G\"{u}nes Baydin, Barak~A. Pearlmutter, Alexey~Andreyevich Radul, and Jeffrey~Mark Siskind.
\newblock Automatic differentiation in machine learning: a survey.
\newblock \emph{The Journal of Machine Learning Research}, 2017.

\bibitem[Reddi et~al.(2021)Reddi, Charles, Zaheer, Garrett, Rush, Kone{\v{c}}n{\'y}, Kumar, and McMahan]{reddi2021adaptive}
Sashank~J. Reddi, Zachary Charles, Manzil Zaheer, Zachary Garrett, Keith Rush, Jakub Kone{\v{c}}n{\'y}, Sanjiv Kumar, and Hugh~Brendan McMahan.
\newblock Adaptive federated optimization.
\newblock In \emph{International Conference on Learning Representations}, 2021.

\bibitem[Thapa et~al.(2022)Thapa, Arachchige, Camtepe, and Sun]{thapa2022splitfed}
Chandra Thapa, Pathum Chamikara~Mahawaga Arachchige, Seyit Camtepe, and Lichao Sun.
\newblock Splitfed: When federated learning meets split learning.
\newblock In \emph{Proceedings of the AAAI Conference on Artificial Intelligence}, 2022.

\bibitem[Zhang et~al.(2024)Zhang, Han, Liu, Zhou, Lu, Li, Gao, and Qiao]{zhang2024llamaadapter}
Renrui Zhang, Jiaming Han, Chris Liu, Aojun Zhou, Pan Lu, Hongsheng Li, Peng Gao, and Yu~Qiao.
\newblock {LL}a{MA}-adapter: Efficient fine-tuning of large language models with zero-initialized attention.
\newblock In \emph{The Twelfth International Conference on Learning Representations}, 2024.

\bibitem[Ben~Zaken et~al.(2022)Ben~Zaken, Goldberg, and Ravfogel]{zaken2022bitfit}
Elad Ben~Zaken, Yoav Goldberg, and Shauli Ravfogel.
\newblock {B}it{F}it: Simple parameter-efficient fine-tuning for transformer-based masked language-models.
\newblock In \emph{Proceedings of the 60th Annual Meeting of the Association for Computational Linguistics (Volume 2: Short Papers)}. Association for Computational Linguistics, 2022.

\bibitem[Panchal et~al.(2023{\natexlab{a}})Panchal, Choudhary, Mitra, Mukherjee, Sarkhel, Mitra, and Guan]{panchal2023flash}
Kunjal Panchal, Sunav Choudhary, Subrata Mitra, Koyel Mukherjee, Somdeb Sarkhel, Saayan Mitra, and Hui Guan.
\newblock Flash: concept drift adaptation in federated learning.
\newblock In \emph{International Conference on Machine Learning}, pages 26931--26962. PMLR, 2023{\natexlab{a}}.

\bibitem[Wu et~al.(2022)Wu, Li, Charles, Xiao, Liu, Xu, and Smith]{wu2022motley}
Shanshan Wu, Tian Li, Zachary Charles, Yu~Xiao, Ziyu Liu, Zheng Xu, and Virginia Smith.
\newblock Motley: Benchmarking heterogeneity and personalization in federated learning.
\newblock \emph{arXiv 2206.09262}, 2022.

\bibitem[Zhang et~al.(2015)Zhang, Zhao, and LeCun]{zhang2015agnewsyelpyahoo}
Xiang Zhang, Junbo~Jake Zhao, and Yann LeCun.
\newblock Character-level convolutional networks for text classification.
\newblock In \emph{Neural Information Processing Systems}, 2015.
\newblock Available at \url{https://huggingface.co/datasets/ag_news}, \url{https://huggingface.co/datasets/yelp_polarity}, \url{https://huggingface.co/datasets/yahoo_answers_topics}, Accessed on 15 May, 2024.

\bibitem[Socher et~al.(2013)Socher, Perelygin, Wu, Chuang, Manning, Ng, and Potts]{socher2013sst2}
Richard Socher, Alex Perelygin, Jean Wu, Jason Chuang, Christopher~D. Manning, Andrew Ng, and Christopher Potts.
\newblock Recursive deep models for semantic compositionality over a sentiment treebank.
\newblock In \emph{Proceedings of the 2013 Conference on Empirical Methods in Natural Language Processing}. Association for Computational Linguistics, 2013.
\newblock Available at \url{https://huggingface.co/datasets/stanfordnlp/sst2}, Accessed on 15 May, 2024.

\bibitem[Bowman et~al.(2015)Bowman, Angeli, Potts, and Manning]{bowman2015snli}
Samuel~R. Bowman, Gabor Angeli, Christopher Potts, and Christopher~D. Manning.
\newblock A large annotated corpus for learning natural language inference.
\newblock In \emph{Proceedings of the 2015 Conference on Empirical Methods in Natural Language Processing}. Association for Computational Linguistics, 2015.
\newblock Available at \url{https://huggingface.co/datasets/stanfordnlp/snli}, Accessed on 15 May, 2024.

\bibitem[Williams et~al.(2018)Williams, Nangia, and Bowman]{williams2018mnli}
Adina Williams, Nikita Nangia, and Samuel Bowman.
\newblock A broad-coverage challenge corpus for sentence understanding through inference.
\newblock In \emph{Proceedings of the 2018 Conference of the North American Chapter of the Association for Computational Linguistics: Human Language Technologies, Volume 1 (Long Papers)}. Association for Computational Linguistics, 2018.
\newblock Available at \url{https://huggingface.co/datasets/SetFit/mnli}, Accessed on 15 May, 2024.

\bibitem[Rajpurkar et~al.(2018)Rajpurkar, Jia, and Liang]{rajpurkar2018squad}
Pranav Rajpurkar, Robin Jia, and Percy Liang.
\newblock Know what you don{'}t know: Unanswerable questions for {SQ}u{AD}.
\newblock In \emph{Proceedings of the 56th Annual Meeting of the Association for Computational Linguistics (Volume 2: Short Papers)}. Association for Computational Linguistics, 2018.
\newblock Available at \url{https://huggingface.co/datasets/rajpurkar/squad_v2}, Accessed on 15 May, 2024.

\bibitem[Khashabi et~al.(2018)Khashabi, Chaturvedi, Roth, Upadhyay, and Roth]{khashabi2018multirc}
Daniel Khashabi, Snigdha Chaturvedi, Michael Roth, Shyam Upadhyay, and Dan Roth.
\newblock Looking beyond the surface: A challenge set for reading comprehension over multiple sentences.
\newblock In \emph{Proceedings of the 2018 Conference of the North {A}merican Chapter of the Association for Computational Linguistics: Human Language Technologies, Volume 1 (Long Papers)}. Association for Computational Linguistics, 2018.
\newblock Available at \url{https://huggingface.co/datasets/mtc/multirc}, Accessed on 15 May, 2024.

\bibitem[Panchal et~al.(2023{\natexlab{b}})Panchal, Choudhary, Parikh, Zhang, and Guan]{panchal2023flow}
Kunjal Panchal, Sunav Choudhary, Nisarg Parikh, Lijun Zhang, and Hui Guan.
\newblock Flow: Per-instance personalized federated learning.
\newblock In \emph{Thirty-seventh Conference on Neural Information Processing Systems}, 2023{\natexlab{b}}.

\bibitem[Zhang et~al.(2022)Zhang, Roller, Goyal, Artetxe, Chen, Chen, Dewan, Diab, Li, Lin, Mihaylov, Ott, Shleifer, Shuster, Simig, Koura, Sridhar, Wang, and Zettlemoyer]{zhang2022opt}
Susan Zhang, Stephen Roller, Naman Goyal, Mikel Artetxe, Moya Chen, Shuohui Chen, Christopher Dewan, Mona Diab, Xian Li, Xi~Victoria Lin, Todor Mihaylov, Myle Ott, Sam Shleifer, Kurt Shuster, Daniel Simig, Punit~Singh Koura, Anjali Sridhar, Tianlu Wang, and Luke Zettlemoyer.
\newblock Opt: Open pre-trained transformer language models.
\newblock \emph{arXiv 2205.01068}, 2022.

\bibitem[Liu et~al.(2019)Liu, Ott, Goyal, Du, Joshi, Chen, Levy, Lewis, Zettlemoyer, and Stoyanov]{liu2019roberta}
Yinhan Liu, Myle Ott, Naman Goyal, Jingfei Du, Mandar Joshi, Danqi Chen, Omer Levy, Mike Lewis, Luke Zettlemoyer, and Veselin Stoyanov.
\newblock Roberta: {A} robustly optimized {BERT} pretraining approach.
\newblock \emph{arxiv 1907.11692}, 2019.

\bibitem[Devlin et~al.(2018)Devlin, Chang, Lee, and Toutanova]{devlin2018bert}
Jacob Devlin, Ming{-}Wei Chang, Kenton Lee, and Kristina Toutanova.
\newblock {BERT:} pre-training of deep bidirectional transformers for language understanding.
\newblock \emph{arXiv 1810.04805}, 2018.

\bibitem[Sanh et~al.(2019)Sanh, Debut, Chaumond, and Wolf]{sanh2019distilbert}
Victor Sanh, Lysandre Debut, Julien Chaumond, and Thomas Wolf.
\newblock Distilbert, a distilled version of bert: smaller, faster, cheaper and lighter.
\newblock \emph{arXiv 1910.01108}, 2019.

\bibitem[Lan et~al.(2019)Lan, Chen, Goodman, Gimpel, Sharma, and Soricut]{lan2019albert}
Zhenzhong Lan, Mingda Chen, Sebastian Goodman, Kevin Gimpel, Piyush Sharma, and Radu Soricut.
\newblock {ALBERT:} {A} lite {BERT} for self-supervised learning of language representations.
\newblock \emph{arXiv 1909.11942}, 2019.

\bibitem[Beutel et~al.(2020)Beutel, Topal, Mathur, Qiu, Parcollet, and Lane]{beutel2020flower}
Daniel~J Beutel, Taner Topal, Akhil Mathur, Xinchi Qiu, Titouan Parcollet, and Nicholas~D Lane.
\newblock Flower: A friendly federated learning research framework.
\newblock \emph{arXiv preprint 2007.14390}, 2020.

\bibitem[aut(2024)]{autogptq}
{AutoGPTQ}, 2024.
\newblock URL \url{https://github.com/AutoGPTQ/AutoGPTQ}.

\bibitem[Burden and Faires(2005)]{burden2005numerical}
Richard~L. Burden and J.~Douglas. Faires.
\newblock \emph{Numerical analysis / Richard L. Burden, J. Douglas Faires.}
\newblock Thomson Brooks/Cole, 8th ed. edition, 2005.
\newblock ISBN 0534392008.

\bibitem[Jord{\'a}n(1965)]{jordan1965calculus}
K{\'a}roly Jord{\'a}n.
\newblock \emph{Calculus of finite differences}.
\newblock American Mathematical Soc., 1965.

\bibitem[Fang et~al.(2022)Fang, Yu, Jiang, Shi, Jones, and Zhou]{fang2022fedzo}
Wenzhi Fang, Ziyi Yu, Yuning Jiang, Yuanming Shi, Colin~N. Jones, and Yong Zhou.
\newblock Communication-efficient stochastic zeroth-order optimization for federated learning.
\newblock \emph{IEEE Transactions on Signal Processing}, 2022.

\bibitem[Ye et~al.(2024)Ye, Wang, Chai, Li, Li, Xu, Du, Wang, and Chen]{ye2024openfedllm}
Rui Ye, Wenhao Wang, Jingyi Chai, Dihan Li, Zexi Li, Yinda Xu, Yaxin Du, Yanfeng Wang, and Siheng Chen.
\newblock Openfedllm: Training large language models on decentralized private data via federated learning.
\newblock \emph{arXiv 2402.06954}, 2024.

\bibitem[Fan et~al.(2023)Fan, Kang, Ma, Chen, Wei, Fan, and Yang]{fan2023fatellm}
Tao Fan, Yan Kang, Guoqiang Ma, Weijing Chen, Wenbin Wei, Lixin Fan, and Qiang Yang.
\newblock Fate-llm: A industrial grade federated learning framework for large language models.
\newblock \emph{arXiv 2310.10049}, 2023.

\bibitem[Lai et~al.(2022)Lai, Dai, Singapuram, Liu, Zhu, Madhyastha, and Chowdhury]{lai2022fedscale}
Fan Lai, Yinwei Dai, Sanjay~S. Singapuram, Jiachen Liu, Xiangfeng Zhu, Harsha~V. Madhyastha, and Mosharaf Chowdhury.
\newblock Fedscale: Benchmarking model and system performance of federated learning at scale.
\newblock \emph{arXiv 2105.11367}, 2022.

\bibitem[Ro et~al.(2022)Ro, Breiner, McConnaughey, Chen, Suresh, Kumar, and Mathews]{ro2022scaling}
Jae Ro, Theresa Breiner, Lara McConnaughey, Mingqing Chen, Ananda Suresh, Shankar Kumar, and Rajiv Mathews.
\newblock Scaling language model size in cross-device federated learning.
\newblock In \emph{Proceedings of the First Workshop on Federated Learning for Natural Language Processing (FL4NLP 2022)}. Association for Computational Linguistics, 2022.

\bibitem[Malaviya et~al.(2023)Malaviya, Shukla, and Lodha]{malaviya2023reducing}
Shubham Malaviya, Manish Shukla, and Sachin Lodha.
\newblock Reducing communication overhead in federated learning for pre-trained language models using parameter-efficient finetuning.
\newblock In \emph{Conference on Lifelong Learning Agents}. PMLR, 2023.

\bibitem[Zhang et~al.(2023)Zhang, Yang, Dai, Wang, Yu, Qu, and Xu]{zhang2023fedpetuning}
Zhuo Zhang, Yuanhang Yang, Yong Dai, Qifan Wang, Yue Yu, Lizhen Qu, and Zenglin Xu.
\newblock {F}ed{PET}uning: When federated learning meets the parameter-efficient tuning methods of pre-trained language models.
\newblock In \emph{Findings of the Association for Computational Linguistics: ACL 2023}. Association for Computational Linguistics, 2023.

\bibitem[Park et~al.(2023)Park, Shin, Chung, and Lee]{park2023fedfwd}
Seonghwan Park, Dahun Shin, Jinseok Chung, and Namhoon Lee.
\newblock Fedfwd: Federated learning without backpropagation.
\newblock \emph{arXiv 2309.01150}, 2023.

\bibitem[Hinton(2022)]{hinton2022forwardforward}
Geoffrey Hinton.
\newblock The forward-forward algorithm: Some preliminary investigations.
\newblock \emph{arXiv 2212.13345}, 2022.

\bibitem[Ding et~al.(2023)Ding, Qin, Yang, Wei, Yang, Su, Hu, Chen, Chan, Chen, et~al.]{ding2023peftsurvey}
Ning Ding, Yujia Qin, Guang Yang, Fuchao Wei, Zonghan Yang, Yusheng Su, Shengding Hu, Yulin Chen, Chi-Min Chan, Weize Chen, et~al.
\newblock Parameter-efficient fine-tuning of large-scale pre-trained language models.
\newblock \emph{Nature Machine Intelligence}, 2023.

\bibitem[Han et~al.(2024)Han, Gao, Liu, Zhang, and Zhang]{han2024peftsurvey}
Zeyu Han, Chao Gao, Jinyang Liu, Jeff Zhang, and Sai~Qian Zhang.
\newblock Parameter-efficient fine-tuning for large models: A comprehensive survey.
\newblock \emph{arXiv 2403.14608}, 2024.

\bibitem[Dettmers et~al.(2023)Dettmers, Pagnoni, Holtzman, and Zettlemoyer]{dettmers2023qlora}
Tim Dettmers, Artidoro Pagnoni, Ari Holtzman, and Luke Zettlemoyer.
\newblock {QL}o{RA}: Efficient finetuning of quantized {LLM}s.
\newblock In \emph{Thirty-seventh Conference on Neural Information Processing Systems}, 2023.

\end{thebibliography}
}

\newpage
\appendix

\clearpage
\section{Related Work}
\label{sec:related-work}
\projectname{} uses first-order forward gradients for federated finetuning of language models.
Hence, we review the literature on estimating gradients with low memory consumption and show how \projectname{} represents a significant advancement in finetuning language models in FL. 

\paragraph{Zero-order Gradients.}
Gradients derived from finite difference~\cite{burden2005numerical, richardson1955finitedifferences, jordan1965calculus} methods are called \emph{zero-order gradients} since they don't involve Taylor expansion of the objective function $f$.
\textsc{MeZO}~\cite{malladi2023mezo} has shown that finite difference with one perturbation per batch does not reach convergence on its own without additional tricks like prompt-based finetuning, which are highly specific to the tasks.
For a more accurate gradient approximation,
an average of zero-order gradients derived from multiple ($\sim$10 to 100) random perturbations on the same input batch is required~\cite{baydin2017autodiff}, leading to slow convergence.
\textsc{Baffle}~\cite{feng2023baffle} and \textsc{FedZO}~\cite{fang2022fedzo} utilize zero-order gradients in federated settings.
To train vision models (of parameter count $\leq$ 13M), \textsc{Baffle} requires 
(a)~$\sim$100-500 perturbations per batch for each client respectively, and 
(b)~per-iteration communication among clients like \textsc{FedSgd}~\cite{macmahan2017aFedLearning}.
\textsc{FedZO} also requires $\sim$20 perturbations per batch for a vision model of size $\leq$ 25M.
Besides, numerical errors associated with finite difference make \textsc{MeZO}, \textsc{Baffle}, and \textsc{FedZO} suffer from sub-optimal predictions compared to backpropagation-based counterparts. 
\projectname{}, using Forward-mode AD,  computes gradients more accurately in a single forward pass compared to averaged gradients obtained through finite difference methods.
This higher accuracy is achieved without needing modifications to model architectures or task structures, while also maintaining a memory footprint similar to that of finite difference methods.

\paragraph{First-order Forward Gradients.}
Gradients derived from Forward-mode Auto-Differentiation (AD) are considered first-order since it involves computing partial derivatives of the intermediate activations with respect to the input.
We guide interested readers to the survey on different modes of automatic differentiation~\cite{baydin2017autodiff}. 
\textsc{Fgd}~\cite{baydin2022gradients} shows preliminary results on the speedup and comparable accuracy achieved by Forward-mode AD against backpropagation.
The challenge that makes Forward-mode AD less popular than backpropagation is that gradients derived from forward mode require more \texttt{jvp} column-by-column evaluations per input batch as the number of trainable weights increases. 
Moreover, evaluation of \textsc{Fgd} is limited to a multi-layer perceptron of size $d\approx$1.8M.
Direct use of \textsc{Fgd} to finetune language models leads to slow or no convergence.
\projectname{} splits the trainable layers of a large language model across multiple clients in FL, letting each client finetune only a small subset of weights through forward gradients. 

\paragraph{Training or Finetuning Language Models in Federated Learning.}
In recent years, several frameworks have been proposed to train or finetune LLMs in FL~\cite{ye2024openfedllm, fan2023fatellm, lai2022fedscale}. 
The backpropagation-based methods~\cite{tian2022fedbert, ro2022scaling}, even with parameter efficient finetuning (PEFT) and quantization~\cite{cai2023fedadapter, malaviya2023reducing, zhang2023fedpetuning}, have large memory footprints due to the overhead related to activations, gradients, and gradient history storage for adaptive optimizers~\cite{malladi2023mezo}. 

\textsc{FedFwd}~\cite{park2023fedfwd} applies \textsc{FwdFwd}~\cite{hinton2022forwardforward} (which measures ``goodness'' of forward pass activations to judge which perturbations are useful) in FL, but \textsc{FwdFwd} struggles as model size scales up. 
\textsc{FwdLLM} uses zero-order gradients to finetune language models. 
It samples $\sim$10 perturbations per batch. 
For each batch, it picks 1 perturbation that has the highest cosine similarity with the previous round's gradients.
Sampling new perturbations based on aggregated gradients from previous rounds during the initial stages can disrupt the learning trajectory.
\projectname{} requires 1 perturbation (without resampling) per batch to reach a higher prediction performance faster than \textsc{FwdLLM}.

\clearpage
\section{Datasets and Hyperparameters}
\label{adx:datasets-hyperparameters}
Here we provide details of the datasets and their corresponding training hyperparameters used in this work. 

\paragraph{Simulating Heterogeneity through Dirichlet Distribution.}
For each of the tasks, the class distribution each client gets depends on the Dirichlet distribution, where a parameter $\alpha$ regulates the concentration of samples of a particular class for a specific client.
Dir $\alpha=1.0$ means all clients have homogeneous datasets where each class is equally likely to be on each client.
With Dir $\alpha \rightarrow 0$, the datasets of each client get more heterogeneous, where the sample distribution of each class is more likely to be concentrated on only a subset of clients.

\paragraph{Default Hyperparameters.}
Unless otherwise mentioned in dataset-specific paragraphs, the default hyperparameters for each method and for all datasets are stated here.
For the backpropagation-based methods \textsc{FedAvg}, \textsc{FedYogi}, and \textsc{FedSgd}; we will fix the number of \textbf{epochs} to 1 since the goal of this work is to inch closer to backpropagation-like prediction performance while reducing the memory footprint. 
All the experiments have been run for 1500 \textbf{FL rounds}, except the experiments on OPT models, which are run for 600 FL rounds.  
Our observation from hyperparameter-tuning shows that the learning rate that gives the best performance is the same for all studied methods. 
\textsc{Baffle} and its memory-efficient improvement \textsc{Baffle+} made by us, can perform better as the number of perturbations per batch increases, but due to the scale of experiments with $10$-$100$ per round and up to 1500 rounds in the FL setting, we limit the total perturbations per batch of \textsc{Baffle+} to 20 perturbations per batch and fixed finite difference step size $\sigma=1$e-4.  
\textsc{FwdLLM+} samples 10 perturbations for each batch, finite difference step size of $\sigma=1$e-2.
\textsc{FedMeZO} samples 1 perturbation for each batch, with finite difference step size of $\sigma=1$e-3.
\textsc{FedMeZO} also requires 3-5 epochs for each client.
\projectname has 1 perturbation per batch for each client. 
For \projectname and its zero-order counterparts (\textsc{Baffle+}, \textsc{FwdLLM+}, and \textsc{FedMeZO}), perturbations are \textbf{sampled for a normal distribution} with 0 mean and 1 variance. 
Default \textsc{LoRA} {$r$ and $\alpha$} are 1 and 1, respectively.
All methods use \textsc{AdamW} as client-side optimizer.
Besides \textsc{FedAvg}, all methods use \textsc{FedYogi} as \textbf{server-side optimizer}.

\paragraph{AG News.}
AG News dataset~\cite{zhang2015agnewsyelpyahoo} has been derived from a corpus of 496,835 categorized news articles. 
A subset of the corpus is used that has 120,000 total training samples and 7,600 total testing samples spread equally across 4 classes. 
The news articles are classified into 4 classes: World, Sports, Business, and Sci/Tech. 
We split this data across 1000 clients. 
Each client gets an equal number of samples for train and test datasets.
This dataset is under Creative Commons CCZero(CC0) public domain dedication.

Learning rate for backpropagation-based (\textsc{FedAvg}, \textsc{FedYogi}, and \textsc{FedSgd}), zero-order-based (\textsc{FwdLLM}, \textsc{Baffle}, and \textsc{FedMeZO}), and first-order-based \projectname is $\{$1e-3, 5e-4, \textbf{1e-4}, 1e-5$\}$.
The batch size is set to $8$.
The max sequence length is 128.
All methods use \textsc{AdamW} as a client-side optimizer, while \projectname performs better with \textsc{Sgd}.
Variance threshold of \textsc{FwdLLM+} is 1e+1.

\paragraph{SST2.}
Stanford Sentiment Treebank Binary\cite{socher2013sst2} (or SST2) dataset is  for a binary sentiment classification task. 
The dataset has 11,855 sentences derived from a set of movie reviews. 
The corpus was parsed using the Stanford Parser into 215,154 discrete phrases annotated by 3 human judges. 
This dataset contains 67,349 training samples, 872 validation samples, and 1821 testing samples. 
This sentiment classification dataset has the following sentiments as classes: Positive, and Negative.
These samples are equally split between 100 clients, depending on the Dirichlet distribution.
This dataset is under Creative Commons CCZero public domain dedication.

Learning rate for backpropagation-based (\textsc{FedAvg}, \textsc{FedYogi}, and \textsc{FedSgd}), zero-order-based (\textsc{FwdLLM}, \textsc{Baffle}, and \textsc{FedMeZO}), and first-order-based \projectname is $\{$1e-3, 5e-4, \textbf{1e-4}, 1e-5$\}$.
The batch size is set to $8$.
The max sequence length is 64.
Variance threshold of \textsc{FwdLLM+} is 5e+0.

\paragraph{Yelp.}
The Yelp reviews dataset\cite{zhang2015agnewsyelpyahoo} is a binary classification dataset gathered during the 2015 Yelp Dataset Challenge. 
The full dataset has 1,569,264 total samples, and it defines 2 classification tasks. 
We use the polarity classification task, which is a binary classification problem. 
It considers 1-2 stars negative polarity and 3-4 stars positive polarity.  This dataset has 280,000 training samples and 19,000 test samples in each polarity. 
We split this data into 1000 clients.
This dataset is under the Apache License, Version 2.0.

Learning rate for backpropagation-based (\textsc{FedAvg}, \textsc{FedYogi}, and \textsc{FedSgd}), zero-order-based (\textsc{FwdLLM}, \textsc{Baffle}, and \textsc{FedMeZO}), and first-order-based \projectname is $\{$1e-3, 5e-4, 1e-4, \textbf{5e-5}, 1e-5$\}$.
The batch size is set to $8$.
The max sequence length is 128.
Variance threshold of \textsc{FwdLLM+} is 5e+1.

\paragraph{Yahoo.}
Yahoo! Answers Comprehensive Questions and Answers dataset \cite{zhang2015agnewsyelpyahoo} was gathered via the Yahoo! webscope program.
The corpus itself contains 4,483,032 question/answer pairs, which were then formulated as a 10-class classification task. 
Each class in this dataset has 140,000 training and 5,000 testing samples.are  
The question/answer pairs are split into the following classes: 1. Society \& Culture, 2. Science \& Mathematics, 3. Health, 4. Education \& Reference, 5. Computers \& Internet, 6. Sports, 7. Business \& Finance, 8. Entertainment \& Music, 9. Family \& Relationships, 10. Politics \& Government. 
We split this dataset between 1000 clients, where each client gets an equal amount of data samples, where the data distribution is set by changing the Dirichlet distribution $\alpha$ to range from most homogeneous ($\alpha=1$) to least homogeneous ($\alpha=0$).
This dataset is under the Apache License, Version 2.0.

Learning rate for backpropagation-based (\textsc{FedAvg}, \textsc{FedYogi}, and \textsc{FedSgd}), zero-order-based (\textsc{FwdLLM}, \textsc{Baffle}, and \textsc{FedMeZO}), and first-order-based \projectname is $\{$1e-3, 5e-4, \textbf{1e-4}, 1e-5$\}$.
The batch size is set to $8$.
The max sequence length is 128.
Variance threshold of \textsc{FwdLLM+} is 5e+1.

\paragraph{SNLI.}
The Stanford Natural Language Inference corpus \cite{bowman2015snli} has 570,152 total sentence pairs. 
It is a natural language inference dataset, where the task is identifying if one sentence infers another. 
It has 550,152 training samples, 10,000 testing samples, and 10,000 evaluation samples.
This dataset is split among 1,000 clients.
The dataset has 3 classes:
1) The first sentence entails the second sentence, 
2) The first sentence is neutral to the second sentence and 
3) The first sentence contradicts the second sentence.
This dataset is under CC BY-SA 4.0. 

Learning rate for backpropagation-based (\textsc{FedAvg}, \textsc{FedYogi}, and \textsc{FedSgd}), zero-order-based (\textsc{FwdLLM}, \textsc{Baffle}, and \textsc{FedMeZO}), and first-order-based \projectname is $\{$1e-3, 5e-4, 1e-4, \textbf{5e-5}, 1e-5$\}$.
The batch size is set to $8$.
The max sequence length is 80.
Variance threshold of \textsc{FwdLLM+} is 1e+2.

\paragraph{MNLI.}
The Multi-Genre Natural Language Inference (MNLI) \cite{williams2018mnli} corpus contains 432,702 sentence pairs which were crowd-sourced and then annotated with textual entailment information. 
This dataset is also a Natural Language Inference dataset as with SNLI. 
This dataset has 392,702 training, 20,000 evaluation, and 20,000 testing samples. 
These samples are split among 1,000 clients.
This dataset draws from multiple sources, most of which are under the Open American National Corpus (OANC) license. 
The rest are under either the CC BY 3.0 Unported licenses or the CC BY-SA 3.0 licenses.

Learning rate for backpropagation-based (\textsc{FedAvg}, \textsc{FedYogi}, and \textsc{FedSgd}), zero-order-based (\textsc{FwdLLM}, \textsc{Baffle}, and \textsc{FedMeZO}), and first-order-based \projectname is $\{$1e-3, 5e-4, 1e-4, \textbf{5e-5}, 1e-5$\}$.
The batch size is set to $8$.
The max sequence length is 80.
Variance threshold of \textsc{FwdLLM+} is 1e+2.

\paragraph{SQuADv2.}
The Stanford Question Answering Dataset (SQuADv2) \cite{rajpurkar2018squad} consists of crowd-sourced questions about a set of Wikipedia articles. 
It is a reading comprehension dataset, where the answer to a question is a section (or a span) from the passage. 
It is also possible for the question to be unanswerable. 
The dataset contains 100,000 answerable and 50,000 unanswerable questions.
This dataset was split into 500 clients.
The heterogeneity is generated based on the topic labels (or ``titles'') associated with each question in the dataset.
There are 35 titles available.
This dataset is under the CC BY-SA 4.0 license.

Learning rate for backpropagation-based (\textsc{FedAvg}, \textsc{FedYogi}, and \textsc{FedSgd}), zero-order-based (\textsc{FwdLLM}, \textsc{Baffle}, and \textsc{FedMeZO}), and first-order-based \projectname is $\{$1e-3, 5e-4, 1e-4, \textbf{1e-5}$\}$.
The batch size is set to $8$ for Forward-mode AD and zero-order differentiation-based methods.
Backpropagation-based methods use a batch size of $4$.
The max sequence length is 400.
Variance threshold of \textsc{FwdLLM+} is 5e+1.

\paragraph{MultiRC.} 
Multi-sentence Reading Comprehension (MultiRC) corpus is a dataset that contains short paragraphs with questions, whose answers can be found in the paragraph itself. 
We consider a task where we classify if the input question given as input is correct or false.
It has 6k multi-sentence questions about 800+ paragraphs.
Due to the scale of the dataset, we split it into 100 clients.
This dataset is under the MIT license. 

Learning rate for backpropagation-based (\textsc{FedAvg}, \textsc{FedYogi}, and \textsc{FedSgd}), zero-order-based (\textsc{FwdLLM}, \textsc{Baffle}, and \textsc{FedMeZO}), and first-order-based \projectname is $\{$1e-3, 5e-4, \textbf{1e-4}, 1e-5$\}$.
The batch size is set to $8$.
The max sequence length is 256.
Variance threshold of \textsc{FwdLLM+} is 1e+2.

\section{Limitations and Future Work}
\label{adx:limitations}
\projectname achieves a remarkable drop in memory consumption due to Forward-mode AD not having to store activations during the forward pass.
However, the current implementation of Forward-mode AD by PyTorch is still suboptimal in terms of computation time.
The high computation time is attributed to the column-by-column computation of the intermediate results of \texttt{jvp}.
Improving the computation of \texttt{jvp} such that it takes significantly less time (almost half) than backpropagation remains an open and interesting problem.
Moreover, there's room for improvement in reducing the memory consumption to that of zero-order methods.
In zero-order methods, the weights are perturbed and a forward pass is computed on the perturbed weights.
However, with the current implementation of Forward-mode AD, perturbations create a separate copy from the original weights, which accrues additional overhead. 
To further utilize the computation capacity of clients in FL, device-heterogeneity-aware strategies on splitting and mapping layers to clients can be explored for \projectname, e.g., layer selection could happen on the client-side according to their data distributions.

\section{Broader Impact}
\label{adx:broader-impact}
Through \projectname, we aspire to bring impact in terms of data privacy and accessible finetuning of large language models (LLMs) on edge devices.

For small and medium organizations and individual users, the cost to finetune these language models can be prohibitively expensive as the size of the trainable weights increases.
With \projectname, we enable finetuning LLMs on resource-constrained edge devices with low memory consumption, making it feasible for a wider range of users.
Moreover, the federated setting also provides benefits of data privacy. 
This can be ideal for use cases where LLMs can bring improved performance for a plethora of personalized downstream tasks, but the finetuning data containing sensitive and confidential information is never shared with a third party. 
With \projectname, such data remains on the user's device, ensuring privacy while still allowing for effective finetuning of the LLM.

However, making LLM finetuning accessible to a broader range of organizations and individuals comes with its own challenges like a spread of biases or misinformation. 
Without preprocessing and filtering client data on their devices, the LLMs can be fed harmful and misleading information.
Hence, it is necessary to develop guardrails on what kind of data should be filtered in, to finetune LLMs with crowd-sourced compute resources.

\clearpage
\section{\projectname Pseudocode}
\label{adx:detailed-pseudocode}
Algorithm \ref{algo:spry} shows the workflow of \projectname{}. 
\RestyleAlgo{ruled}
\begin{algorithm}
\footnotesize
\caption{\text{\projectname}}
\label{algo:spry}
\KwData{
$R$: Total number of rounds,
$r \in [R]$: Round index,
$M$: Number of clients per round,
$m \in [M]$: Client index,
$\cM$: Set of available clients,
$s$: Client sampling rate,
$\cD_m$: Dataset of client $m$,
$\eta_\ell$: Local learning rate,
$\w{r}$: Model weights at $r^{th}$ round,
$\wmtrain{r}{m}$: Subset of assigned trainable weights for client $m$ at $r^{th}$ round,
$f$: Objective function.  
} 
\KwResult{$\w{R+1}$: Model at the end of the training} 
\Main{\textsc{Spry}}{
    Server loads pre-trained LM $\w{1}$ with \textsc{PEFT}\\
    \For{$r \in [R]$ round}{
        Sample $M$ clients from $\cM{}$ at rate of $s$\\
        $\cL \gets$ list of trainable \textsc{PEFT} param names\\
        \texttt{client\_layer\_mapping} $\gets$ \textsc{MapLayersToClients}($\cL, M$)\\
        \For{$m \in [M]$ in parallel}{
         $\wmtrain{r}{m}$ $\gets$ \textsc{ClientTrain}($\w{r}$, \texttt{client\_layer\_mapping}$[m]$)
        }
        Build $\wprime{r}$ with $\wmtrain{r}{m}$ using $\texttt{client\_layer\_mapping}[m]$, $\forall m \in [M]$\\
        Use adaptive optimizer like \textsc{FedYogi} to build $\w{r+1}$ based on the aggregated $\wprime{r}$
    }
}
\begin{multicols}{2}
\Function{\textsc{MapLayersToClients}($\cL, M$) \label{fn:map-layers-to-clients}}{
    \texttt{client\_idx} $\gets$ 0; $\;$
    \texttt{mapping} $\gets$ \{\}\\
    \For{\text{name} $\in \cL$}{
        \texttt{rollover\_idx} $\gets$ \texttt{client\_idx} \% $M$\\
        \texttt{mapping}[\texttt{rollover\_idx}].\text{update}(\textit{name})\\
        \texttt{client\_idx} $\gets$ \texttt{client\_idx} + 1\\
    }
    \Return \texttt{mapping}
}
\columnbreak
\Function{\textsc{ClientTrain}($\wm{r}{m}$, \texttt{trainable\_layers})\label{fn:client-train}}{
    Freeze $\wm{r}{m}$ parameters $\notin$ \texttt{trainable\_layers} \\
    Generate perturbations $\pmb{v} \gets \cN(0, \mathbf{I}_{\wmtrain{r}{m}.\text{shape}})$; \phantom{phantomphantomphantom} $\forall \text{ trainable } \wmtrain{r}{m}$ \\ 
    \texttt{jvp} $\gets f$.\textsc{ForwardAD}$(\wmtrain{r}{m}, \pmb{v}; \cD_m)$ \\
    $\wmtrain{r}{m} \gets \wmtrain{r}{m} - \eta_\ell (\pmb{v} \cdot \texttt{jvp})$\\
    \Return all trainable $\wmtrain{r}{m}$
}
\end{multicols}
\end{algorithm}\\
The function \textsc{MapLayersToClients} on Line \ref{fn:map-layers-to-clients} in Algorithm \ref{algo:spry} shows how we have assigned only a few trainable layers to each client in FL, to make Forward-mode AD more effective at generating better estimation of the gradients.
In function \textsc{ClientTrain} on Line \ref{fn:client-train} of Algorithm \ref{algo:spry}; each client $m$ gets a copy of the entire language model $\wm{r}{m}$ and a list \texttt{trainable\_layers} of parameter names the client $m$ has to train.
A client $m$ freezes the parameters that are not included in its \texttt{trainable\_layers}. 
And for each of the parameter $\wmtrainparam{r}{m}$ which need to be trained, \projectname{} generates a corresponding random perturbation $v$ using a normal distribution $\cN(0, \mathbf{I}_{\wmtrainparam{r}{m}.\text{shape}})$.

Once a client $m$ has obtained forward gradients of all the trainable parameters $\wmtrain{r}{m}$, those parameters are locally updated with optimizers like \textsc{SGD} or \textsc{Adam}. 
The updated trainable parameters $\wmtrain{r}{m}$ are sent back to the server.  
The server has a mapping of parameter names to client IDs, and hence it builds $\wprime{r}$ by using $\wmtrain{r}{m} \; \forall m \in [M]$.
If there are multiple clients mapped to the same parameter, then we take a weighted average (similar to \textsc{FedAvg}) of all the parameters to build $\wprime{r}$.
\projectname{} uses adaptive optimizers like \textsc{FedYogi} at server-side on effective gradients $\Delta = \wprime{r} - \w{r}$ to reduce the noise of forward gradients.

\clearpage
\section{Communication and Computation Costs}
\label{adx:communication-and-computation-costs}
\subsection{Communication Costs}
Table~\ref{tbl:communication-cost} illustrates communication costs of \projectname and its backpropagation- and finite difference- based baselines. 
A discussion on communication modes of \projectname is also given in Section 3.2, “Per-Epoch Communication” and “Per-Iteration Communication”. 
\begin{table}[h]
\caption{Communication cost of \textsc{Spry} and all its baselines. $M$ is the count of participation clients. Total count of trainable parameters of a global model is $w_g = w_\ell L$ (for simplicity, we assume that each layer has $w_\ell$ parameters).}
\label{tbl:communication-cost}
\centering
\footnotesize
\begin{tabular}{cccc} \toprule
Gradient computation & \makecell{Method \\ (Comm. frequency)} & \makecell{Communication cost \\ (in parameter count) \\ \textbf{from each client} \\ \textbf{to server} for each\\ communication round} & \makecell{Communication cost\\ (in parameter count) \\\textbf{from server to all} \\ \textbf{clients} for each \\ communication round} \\ \midrule
\multirow{2}{*}{\begin{tabular}[c]{@{}c@{}}Backpropagation\\(First-order gradients)\end{tabular}} & \makecell{\textsc{FedAvg} / \textsc{FedYogi} \\ (Per-epoch)} & $w_g$ & $w_g \times M$ \\ \cmidrule{2-4}
 & \makecell{\textsc{FedSgd} \\ (Per-iteration) } & $w_g$ & $w_g \times M$ \\ \midrule
\multirow{2}{*}{\begin{tabular}[c]{@{}c@{}}Finite differences\\(Zero-order gradients)\end{tabular}} & \makecell{\textsc{FedMeZO} / \textsc{FwdLLM} / \\\textsc{Baffle} (Per-epoch)} & $w_g$ & $w_g \times M$ \\ \cmidrule{2-4}
 & \makecell{\textsc{FedMeZO} / \textsc{FwdLLM}\\\textsc{Baffle} (Per-iteration)} & \begin{tabular}[c]{@{}c@{}}1 (of finite \\ difference scalar)\end{tabular} & \begin{tabular}[c]{@{}l@{}} \makecell{$(w_g + 1) \times M$ \\ (``1'' is for \\ perturbation seed)}\end{tabular} \\ \midrule
\multirow{2}{*}{\begin{tabular}[c]{@{}c@{}}Forward-mode AD\\(First-order gradients)\end{tabular}} & \makecell{\textsc{Spry}\\(Per-epoch)} & \makecell{$w_\ell \times \max(L / M, 1)$ \\
(Assuming $L \% M = 0$ \\ for each of exposition) }& \makecell{$w_\ell \times \max(L / M, 1) \times M$ \\ = $w_\ell \times \max(L, M)$} \\ \cmidrule{2-4}
 & \makecell{\textsc{Spry}\\(Per-iteration)} & 1 (of \texttt{jvp} scalar) & \makecell{($w_\ell \times \max(L / M, 1) \times M)$ \\ $+ (1 \times M)$\\$= w_\ell \times \max(L, M) + M$} \\ \bottomrule
\end{tabular}
\end{table}

Here we discuss the costs related to those communication modes:
\paragraph{Per-epoch Communication.}
\projectname's client-to-server communication cost does not scale linearly with clients like in its backpropagation and finite-difference counterparts, but instead decreases or stays constant for $L$ as more clients are present. 
Server-to-client communication cost is lower in \projectname due to only sending one layer per client when $M > L$ or $\frac{L}{M}$ layers per client otherwise. This result follows from the below observation:

Backpropagation-based and finite-difference-based methods have a communication cost of $w_g$, where $w_g$ represents the global model size. Each client in $[M]$ (set of participating clients) receives all trainable parameters from the server, requiring the server to send a total of $w_g \times M$ parameters each round.

\projectname's communication cost per epoch is $w_\ell \max(\frac{L}{M}, 1)$, where $L$ is the layer count and $w_\ell$ is the count of parameters for each layer. 
Each client sends a subset of trainable parameters, incurring a communication cost of $\frac{w_\ell L}{M}$ parameters for $L > M$, and $w_\ell$ for $L \leq M$.
When $L \leq M$, each client gets 1 layer, hence the communication for each client is $w_\ell$.
 
\paragraph{Per-iteration Communication.}
\projectname accrues lower communication cost than the finite difference and backpropagation counterparts due to the layer splitting strategy, and the server’s ability to compute gradients based on the \texttt{jvp} value. This is because:

The communication cost from client to server for forward-mode AD and finite differences is 1. 
This is due to an FL round that involves (1) server selecting a random seed, (2) server sending it with trainable parameters to clients, (3) clients generating perturbations based on the seed, (4) deriving and sending back a scalar or finite difference scalar to the server, and then (5) server computing gradients by multiplying the derived perturbations with the seed.

The server to client communication is $(w_g + 1)\times M$, where the ``+1'' is due the randomness seed.

\subsection{Computation Costs}
Table~\ref{tbl:computation-cost} shows the computation costs of \projectname and its baselines, where the client-side cost is for each iteration, and the server-side cost is for each round.

Briefly, \projectname{}’s client-side computation cost is traded off by a faster convergence to higher accuracy through better gradient approximations compared to finite difference-based methods.
Furthermore, \projectname is the least computationally expensive on the server side due to needing to aggregate fewer parameters from the clients.

Table~\ref{tbl:computation-cost} assumes that matrix multiplication costs $c$ for each layer, resulting in a forward pass cost of $c$. 
The cost of backpropagation is $2c$ because the computation of the current layer’s weight gradient is $c$, and the cost of computing the previous layer’s activation gradient is another $c$. 
\texttt{jvp} computation in \projectname takes additional cost of $c$ for each layer. 
Moreover, since \texttt{jvp} calculation happens through column-by-column vector multiplications (Sec 3.1 of~\cite{baydin2017autodiff}), the related overhead is quantified by $v$.

\begin{table}[h]
\caption{Computation cost of \textsc{Spry} and all its baselines. The client-side cost is for each iteration, and the server-side cost is for each round. $L$ is the layer count, $M$ is the participating client count, $c$ is the cost of matrix multiplication for each layer. $v$ is the overhead related to column-by-column vector multiplications of \texttt{jvp}. $w_\ell$ is the size of each layer and hence size of each layer's perturbation too. $K$ is the perturbation count per iteration. $K=1$ for \textsc{Spry} and \textsc{FedMeZO}, and $\sim20$ for \textsc{Baffle} and \textsc{FwdLLM}.}
\centering
\footnotesize
\label{tbl:computation-cost}
\begin{tabular}{cccc}
\toprule
\makecell{Gradient computation} &
  \makecell{Method \\ (Comm. frequency)} &
  \makecell{Computation cost of \\ \textbf{each client for} \\\textbf{each iteration}} &
  \makecell{Computation cost of \\ \textbf{the server for} \\ \textbf{ each round}} \\
  \midrule
   \multirow{2}{*}{\begin{tabular}[c]{@{}c@{}}Backpropagation \\ (First-order \\ gradients)\end{tabular}} &
  \makecell{\textsc{FedAvg} / \textsc{FedYogi} \\ (Per-epoch)} &
  $3Lc$ &
  \makecell{(Aggregating $L$ layer \\ weights from $M$ clients) \\ $(M-1)  \times w_\ell L$} \\ \cmidrule{2-4}
 &
  \makecell{\textsc{FedSgd} \\ (Per-iteration)} &
  $3Lc$ &
  $(M-1) \times w_\ell L$ \\ \midrule
  \multirow{4}{*}{\begin{tabular}[c]{@{}c@{}}Finite differences \\ (Zero-order \\ gradients)\end{tabular}} &
  \makecell{\textsc{FedMeZO} \\ (Per-epoch)} &
  $L (2c + 3 w_\ell $) &
  $(M-1) \times w_\ell L$ \\ \cmidrule{2-4}
 &
  \makecell{\textsc{FwdLLM} / \textsc{Baffle} \\ (Per-epoch)} &
  $KL (2c + w_\ell $) &
  $(M-1) \times w_\ell L$ \\ \cmidrule{2-4}
 &
  \makecell{\textsc{FedMeZO} \\ (Per-iteration)} &
  $L (2c + 3w_\ell $) &
  \makecell{$ w_\ell L + w_\ell ML + w_\ell (M-1) L$ \\ $= 2 M w_\ell L$ \\ (perturbation generation \\ + gradient calculation \\ + weight update)
} \\ \cmidrule{2-4}
 &
  \makecell{\textsc{FwdLLM} / \textsc{Baffle} \\ (Per-iteration)} &
  $KL (2c + w_\ell $) &
  $  2 M w_\ell L$ \\ \midrule
\multirow{2}{*}{\begin{tabular}[c]{@{}c@{}}Forward-mode AD \\ (First-order \\ gradients)\end{tabular}} &
  \makecell{\textsc{Spry} \\ (Per-epoch)} &
  \makecell{$2 \times \max(\frac{L}{M} , 1)$ \\ $\times (c + v) + w_\ell L$} &
  \makecell{$\sum_{\mathcal{M} \subset [M]} \Big( (|\mathcal{M}|-1) w_\ell \max{(\frac{L}{M},1)} \Big)$ \\ (usually, $|\mathcal{M}| = \max{(\frac{M}{L},1)}$)} \\ \cmidrule{2-4}
 &
  \makecell{\textsc{Spry} \\ (Per-iteration)} &
  \makecell{$2 \times \max(\frac{L}{M} , 1) $ \\ $\times (c + v) + w_\ell L$} &
  \makecell{$\sum_{\mathcal{M} \subset [M]} \Big( 2|\mathcal{M}| w_\ell \max{(\frac{L}{M},1)}  \Big)$ }\\
  \bottomrule
\end{tabular}
\end{table}
\paragraph{Client-side per-iteration computation cost.}
Backpropagation needs 3 matrix multiplication operations per layer. 
For zero-order methods, there are 2 matrix multiplications (incurred due to two forward passes) per layer, and per perturbation within a training iteration; and additional overhead $w_\ell K L$ for perturbation generation. 
\textsc{MeZO} requires generation of perturbations thrice for the same seed (Algorithm 1 in \textsc{MeZO}~\cite{malladi2023mezo}).

\projectname’s computation cost is $2\times \max(\frac{L}{M}, 1)\times (c + v) + w_\ell L$. 
Since \projectname allocates at most $\frac{L}{M}$ layers to each client, the computation cost only scales with $\max(\frac{L}{M}, 1)$, against its counterparts scaling with $L$.
However, forward-mode AD computes \texttt{jvp} column-wise, while its counterpart \texttt{vjp} in backpropagation is computed row-wise.
This results in time overhead (
) if the number of trainable parameters exceeds the output size (1 as loss is scalar), which is the case for neural networks. 
Therefore, \projectname's per-iteration computation cost is higher compared to other approaches.

Note that the per-iteration computation cost of \projectname is not the whole picture. 
It takes fewer communication rounds to reach higher accuracy due to better gradient approximation of forward-mode AD than finite difference methods. 
This is why "Time to Convergence" (Section~\ref{subsec:wall-clock-time}) discusses a fair comparison of \projectname's runtime and prediction performance.

\paragraph{Server-side per-round computation cost.}
On the server side, \projectname is the least computationally demanding. 
\projectname needs to aggregate a subset of layer weights from only the clients that were assigned to those layers, while its counterparts need to aggregate all layers from all clients.

Computation cost on the server-side changes based on the communication frequency per-iteration communication incurs an additional overhead of $w_\ell L (\frac{M}{L} + 1)$ and $w_\ell L (M + 1)$ (generation of perturbations at the server-side, and multiplying those perturbations with aggregate of the \texttt{jvp} values received from the clients) for \projectname and its zero-order counterparts respectively.

\clearpage
\section{Ablation Studies}
Here we will dive into various components of \projectname and see their impact on the performance of \projectname.
\label{adx:ablation-studies}
\paragraph{\projectname can Generalize to Different Language Model Architectures.} Table~\ref{tbl:acc-gen-pers-additional} shows generalized and personalized accuracies $Acc_g$ and $Acc_p$ for homogeneous data splits for language model architectures other than RoBERTa Large.
\begin{table}[t]
\begin{center}
\captionsetup{justification=centering}
\footnotesize
\tabcolsep=0.06cm
\caption{Generalized ($Acc_g$) and personalized ($Acc_p$) accuracies (the higher, the better) for \projectname{} and its backpropagation and zero-order based counterparts on various language model architectures.\\The datasets are split with Dir $\alpha=0.1$.
}
\label{tbl:acc-gen-pers-additional}
\resizebox{\textwidth}{!}{
\begin{tabular}{lcccccccc}
\toprule
 & \multicolumn{4}{c}{Backpropgation-based} & \multicolumn{2}{c}{\begin{tabular}[c]{@{}c@{}}Zero-order\\based Method\end{tabular}} & \multicolumn{2}{l}{\begin{tabular}[c]{@{}c@{}}First-order\\Forward-mode AD\end{tabular}} \\ \cmidrule{2-9}
 & \multicolumn{2}{c}{\textsc{FedAvg}} & \multicolumn{2}{c}{\textsc{FedYogi}} & \multicolumn{2}{c}{\textsc{FwdLLM+}} & \multicolumn{2}{c}{\projectname{}} \\ 
\multicolumn{1}{l}{} & $Acc_g$ & \multicolumn{1}{l}{$Acc_p$} & $Acc_g$ & \multicolumn{1}{l}{$Acc_p$} & $Acc_g$ & \multicolumn{1}{l}{$Acc_p$} & $Acc_g$ & $Acc_p$ \\ 
\midrule
\multicolumn{1}{l}{\begin{tabular}[c]{@{}l@{}}AG News on BERT Base\end{tabular}} & 93.00\% & \multicolumn{1}{l}{93.34\%} & 93.31\% & 93.88\% & 83.41\% & 83.42\% & 86.74\% & 92.42\% \\
\multicolumn{1}{l}{\begin{tabular}[c]{@{}l@{}}SST2 on DistilBERT Base \phantom{place}\end{tabular}} & 91.47\% & \multicolumn{1}{l}{95.28\%} & 87.95\% & 92.97\% & 79.09\% & 80.94\% & 84.90\% & 87.12\% \\
\multicolumn{1}{l}{\begin{tabular}[c]{@{}l@{}}SNLI on BERT Large\end{tabular}} & 85.79\% & \multicolumn{1}{l}{89.36\%} & 86.72\% & 90.33\% & 66.76\%  & 64.84\% & 77.01\% & 77.21\% \\
\multicolumn{1}{l}{\begin{tabular}[c]{@{}l@{}}Yahoo on DistilBERT Base\end{tabular}} & 69.13\% & \multicolumn{1}{l}{74.75\%} & 63.84\% & 71.25\% & 54.29\% & 55.87\% & 61.17\% & 61.47\% \\
\multicolumn{1}{l}{\begin{tabular}[c]{@{}l@{}}Yelp on AlbertV2 Large\end{tabular}} & 90.17\% & \multicolumn{1}{l}{92.78\%} & 90.24\% & 94.00\% & 82.65\% & 83.25\% & 85.80\% & 86.00\% \\
\bottomrule
\end{tabular}
}
\end{center}
\end{table}

For zero-order gradients, we show the results of the best-performing method, \textsc{FwdLLM+}.
We see a similar trend of \projectname outperforming \textsc{FwdLLM+} by 3.15-10.25\% for generalized accuracy, and by 2.75-12.37\% for personalized accuracy, exhibiting how \projectname is independent of model architectures. 
\projectname also comes as close as 4.44-9.71\% to the best-performing backpropagation-based method.

\begin{figure}[t]
     \centering
     \begin{subfigure}[b]{0.305\textwidth}
         \centering
         \includegraphics[width=\textwidth]{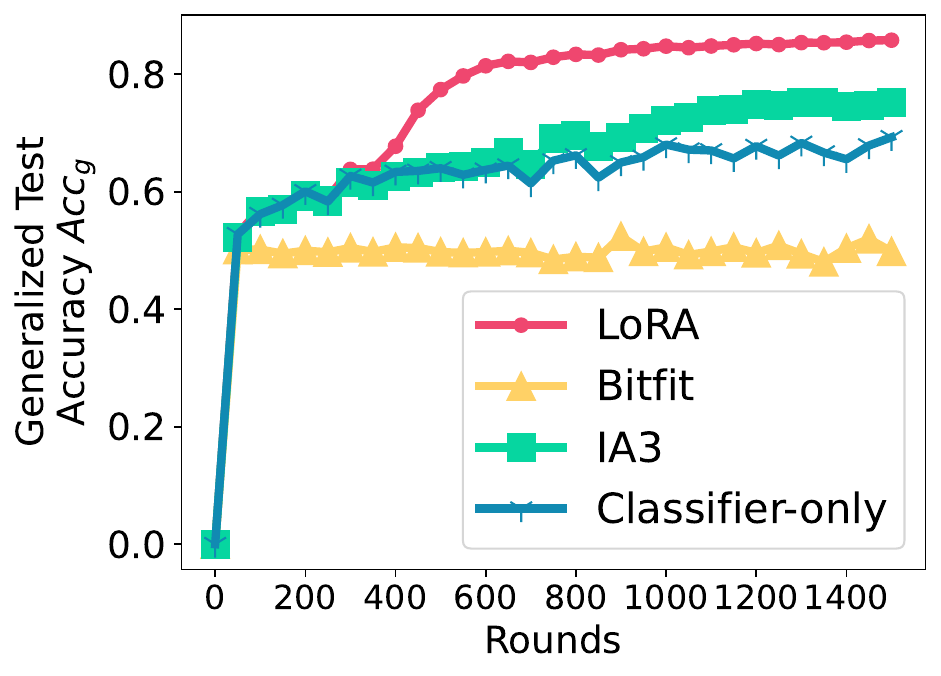}
         \caption{Effect of different \textsc{PEFT} methods on \projectname{}}
         \label{fig:ablation-peft}
     \end{subfigure}
     \hfill
     \begin{subfigure}[b]{0.365\textwidth}
         \centering
    \includegraphics[width=\textwidth]{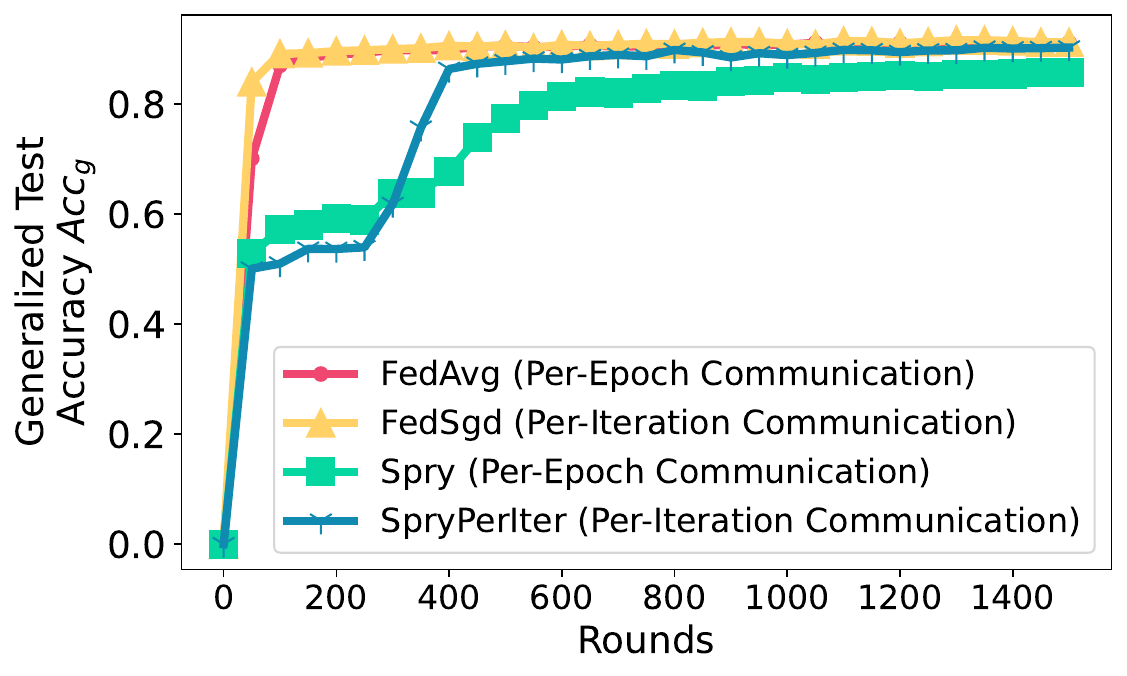}
         \caption{Effect of per-epoch and per-iteration communication}
         \label{fig:ablation-per-iteration}
     \end{subfigure}
     \hfill
     \begin{subfigure}[b]{0.305\textwidth}
         \centering
         \includegraphics[width=\textwidth]{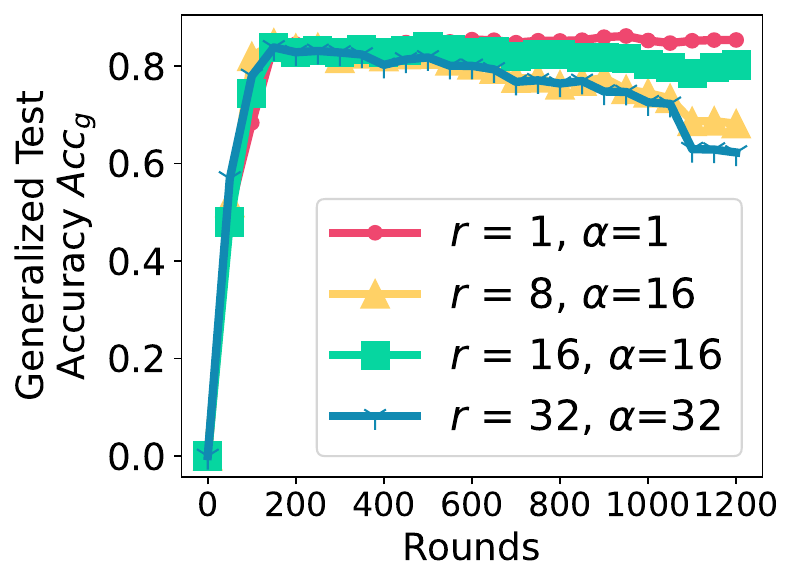}
         \caption{Changing \textsc{LoRA} $r$ and $\alpha$ for \projectname{}}
         \label{fig:ablation-lora-r}
     \end{subfigure}
        \caption{Ablation studies on PEFT methods, communication frequency, and \textsc{LoRA} hyperparameters.}
        \label{fig:ablation-peft-communication-lora}
\end{figure}

\paragraph{\projectname Supports Other PEFT Methods.}
Figure~\ref{fig:ablation-peft} shows the generalized accuracy of \projectname using three different PEFT methods: \textsc{LoRA}, \textsc{IA3}, and \textsc{Bitfit}. 
We also experiment with finetuning only classifier layers, calling it \textsc{Classifier-Only}.
\textsc{LoRA} (with 0.3241\% of the total parameters of RoBERTa Large) outperforms \textsc{IA3} (with 0.3449\% of the total parameters) by 10.60\%, while \textsc{BitFit} fails to converge for all datasets.
Only training classifier layers is worse than finetuning \textsc{LoRA} weights by 16.53\%.
The observation on comparison of \textsc{LoRA} with \textsc{IA3} is consistent with the work benchmarking various PEFT methods~\cite{ding2023peftsurvey}.
\textsc{BitFit} has been observed to fail on LLMs~\cite{han2024peftsurvey}. 
Furthermore, unlike \textsc{IA3}, \textsc{LoRA} has been shown to be successful at finetuning quantized billion-sized models in \textsc{QLoRA}~\cite{dettmers2023qlora}, making it a strong candidate for our work.

\paragraph{Effects of Communication Frequency.}
One way to reduce the noise introduced by the random perturbations in gradient computation is, to communicate the gradients back to the server every iteration instead of every few epochs. 
Figure~\ref{fig:ablation-per-iteration} shows results of per-epoch and per-iteration communication variants of \projectname and \textsc{FedAvg}. 

By communicating each iteration, we see a boost of 4.47\% in the accuracy of \projectname, only 0.92\% and 0.96\% away from the accuracy of \textsc{FedAvg} and \textsc{FedSgd} respectively.
Furthermore, as shown in Figure \ref{fig:ablation-per-iteration}, the convergence speed of \projectname also improves by communicating each iteration.

As discussed in Section~\ref{sec:methodology}, to reduce the trade-off between performance gains and communication cost per iteration, each client can send only the \texttt{jvp} scalar to the server instead of transmitting all the assigned trainable weights~\cite{feng2023baffle}. 
With the seed value of the randomness, the server can then generate the random perturbation vector $\pmb{v}$, which was used by all the clients to generate their respective \texttt{jvp} values.
Then the random perturbation is multiplied with the received \texttt{jvp} values of all clients, to compute gradients.


\paragraph{Effects of the Number of Trainable Weights.}
Figure \ref{fig:ablation-lora-r} displays the effect of changing trainable \textsc{LoRA} parameter count on the prediction performance of \projectname.
For DistilBERT Base (total parameter count of 66M), \textsc{LoRA} reduces the trainable parameter count to 0.61M (0.91\%), 0.74M (1.11\%), 0.89M (1.33\%), and 1.18M (1.77\%) with \textsc{LoRA} hyperparameter settings of ($r$=1, $\alpha$=1), ($r$=8, $\alpha$=16), ($r$=16, $\alpha$=16), and ($r$=32, $\alpha$=32).

\projectname achieves the highest accuracy of 84.90\% with ($r$=1, $\alpha$=1) setting, which has the smallest trainable parameter count.  
The accuracy increases as the layer size decreases since fewer perturbed weights provide less noisy gradients. 

\begin{figure}[t]
     \centering
     \begin{subfigure}[b]{0.24\textwidth}
         \centering
         \includegraphics[width=\textwidth]{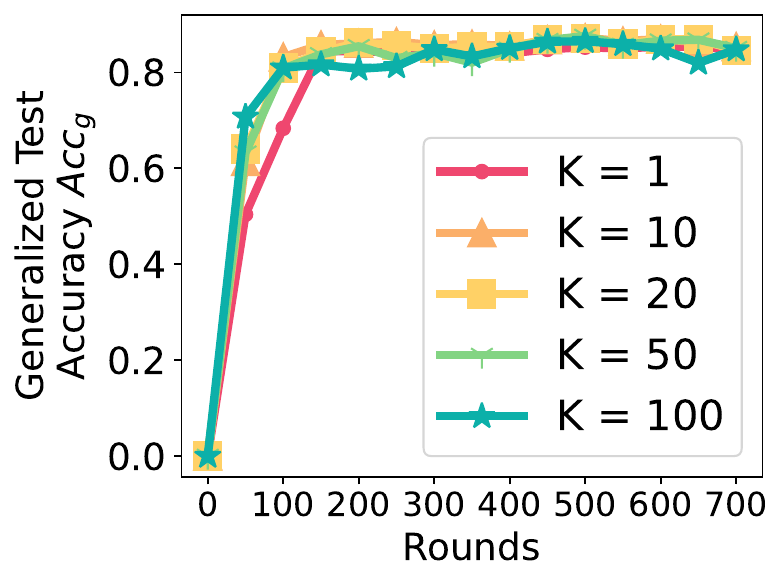}
         \caption{Changing $K$ (Forward-mode AD perturbation count per iteration)}
         \label{fig:ablation-perturbation-count}
     \end{subfigure}
     \hfill
     \begin{subfigure}[b]{0.24\textwidth}
         \centering
    \includegraphics[width=\textwidth]{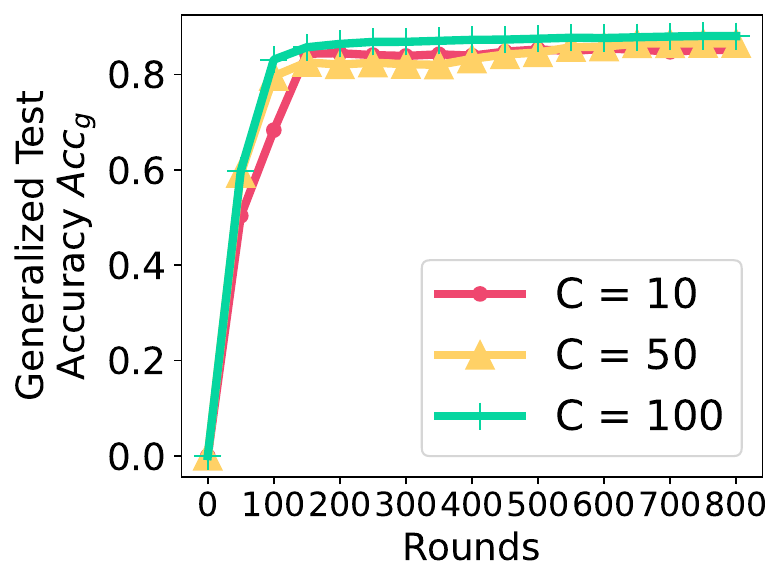}
         \caption{Changing $C$ (Per-round participating client count)\\\phantom{}}
         \label{fig:ablation-client-count}
     \end{subfigure}
     \hfill
     \begin{subfigure}[b]{0.50\textwidth}
         \centering
         \includegraphics[width=0.49\textwidth]{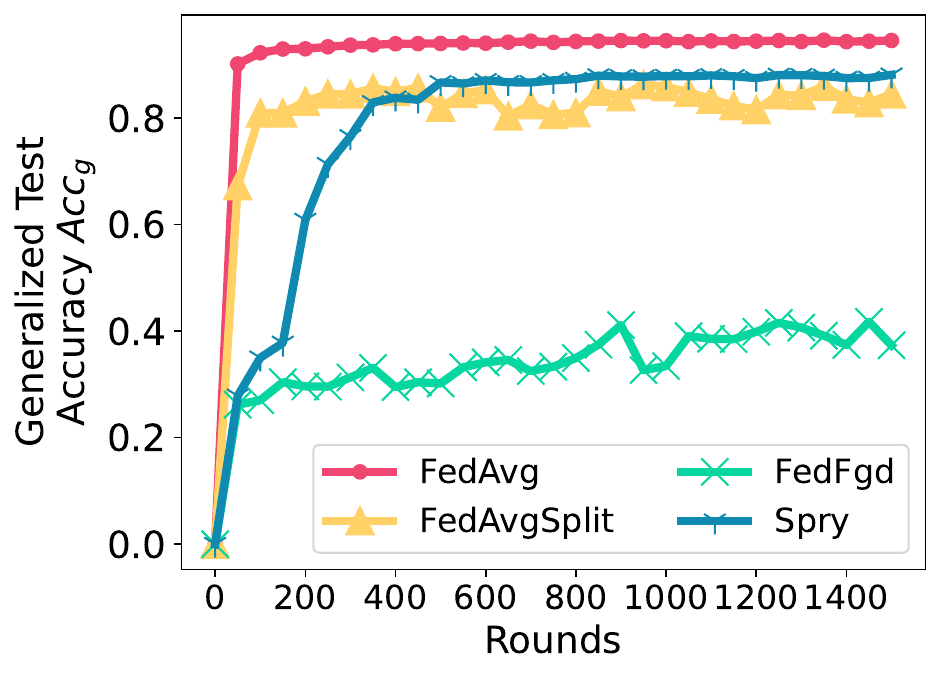}
         \hfill
         \includegraphics[width=0.49\textwidth]{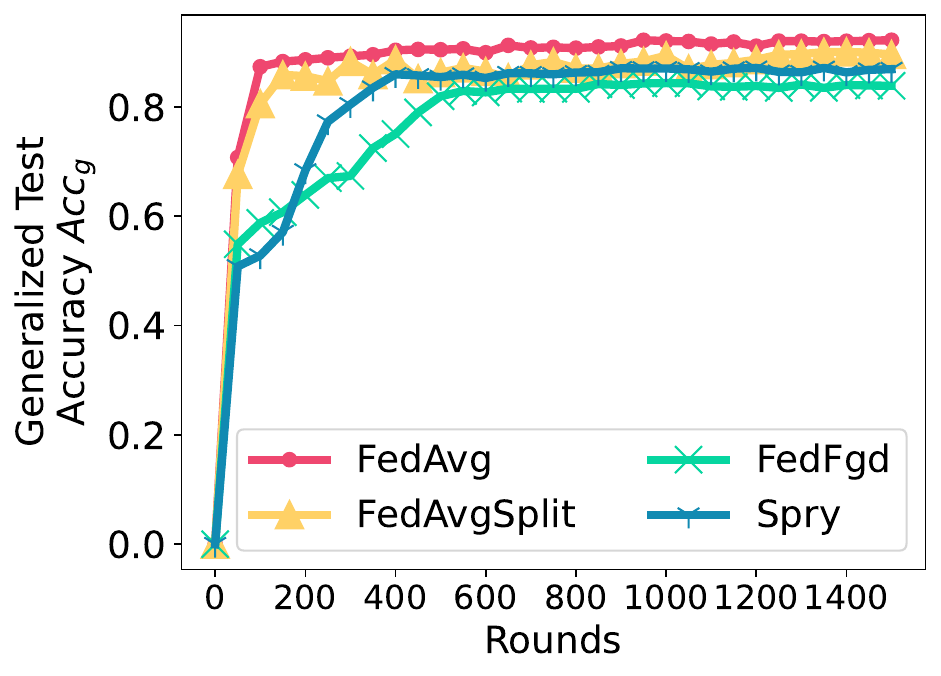}
         \caption{RoBERTa Large (Left) and BERT Base (Right) - Splitting parameters across clients on Backpropagation vs Forward-mode AD\\\phantom{}}
         \label{fig:ablation-splits}
     \end{subfigure}
        \caption{Ablation studies on perturbation counts, participating client counts, and layer splitting strategy.}
        \label{fig:ablation-perts-clients-splits}
\end{figure}

\paragraph{Effects of the Number of Perturbations per Batch.}
\label{subsubsec:perturbations-per-batch}
The effect of increasing the number of perturbations per batch and hence the number of \texttt{jvp} evaluations for a batch is shown in Figure \ref{fig:ablation-perturbation-count} for the SST2 dataset.
Here, gradients generated from each random perturbation and their corresponding \texttt{jvp} values are averaged to update the model.
We observe that increasing $K$ (perturbations per batch) for Forward-mode AD has little to no impact on the end prediction performance of \projectname, with $K=100$ improving the generalized test accuracy by 1.1\% over $K=1$. 
However, the benefits of increasing the perturbation count per batch are seen in the convergence speed. 
Setting $K=10$ achieves a steady state (of accuracy $\sim$86\%) around 200\textsuperscript{th} round, while the setting with $K=1$ takes 500 rounds.
The improvements in convergence speed are saturated for $K>10$.
This shows that more perturbations reduce the gradient estimation noise only to an extent.

\paragraph{Effects of the Number of Participating Clients per Round.} 
\label{subsubsec:ablation-client-count}
Figure \ref{fig:ablation-client-count} shows how changing the per-round number of participating clients $C$ influences \projectname on the SST2 dataset.
Increasing client count increases the prediction performance of \projectname.
With the total client count fixed to 100, the three settings $C=10$, $C=50$, and $C=100$ produce accuracies of $85.14\%$, $86.56\%$, and $88.08\%$, respectively.
Similar to the findings of Section \ref{subsubsec:perturbations-per-batch}, we also see an improvement in the convergence speed as the participating client count increases.
To achieve an accuracy of $\sim$85\%; $C=10$, $C=50$, $C=100$ require 500, 450, and 150 rounds respectively.
The performance gains and faster convergence are due to more clients training the weights of the same layers.

\paragraph{The Importance of Splitting Layers.}
\label{subsubsec:ablation-splitting-layers}
To understand the effects of splitting, we compare the results of the following two experiments:
(a)~With \textsc{FedAvgSplit}, we apply the strategy of splitting trainable layers across clients (Section~\ref{subsubsec:splitting-layers}) to backpropagation-based \textsc{FedAvg}, and 
(b)~With \textsc{FedFgd}, we omit the splitting strategy of \projectname.

Figure~\ref{fig:ablation-splits} shows the performance of \textsc{FedAvg} and \textsc{FedAvgSplit} against \textsc{FedFgd} and \projectname for two LMs: RoBERTa Large (355M) and BERT Base (110M).
We observe that \textsc{FedAvgSplit} fails to achieve similar accuracy for both models with a drop of 2.60\% and 10.00\%.
This is because in \textsc{FedAvgSplit}, fewer clients are training each subset of weights.
Moreover, we see a similar accuracy with an absolute difference of 2.70-3.61\% between \textsc{FedAvgSplit} and \projectname, since the trainable weight count per client is low. 
\textsc{FedYogiSplit} follows the same observation of not achieving similar accuracy to \textsc{FedYogi} if the trainable weights are split across clients.
On the contrary, \textsc{FedFgd} converges for the smaller model BERT base, albeit 150 rounds slower than \projectname, and with 2.87\% accuracy drop.
But as the size of trainable weights increases, e.g., for RoBERTa Large, \textsc{FedFgd} fails to converge.
This proves the necessity of splitting layers for Forward-mode AD so that each client has fewer trainable weights to perturb.

\clearpage
\section{Additional Results}
\label{adx:additional-results}

\subsection{Generalized Performance Curves}
\label{adx:performance-curves}
\begin{figure}[h]
     \centering
     \begin{subfigure}[b]{0.32\textwidth}
         \centering
         \includegraphics[width=\textwidth]{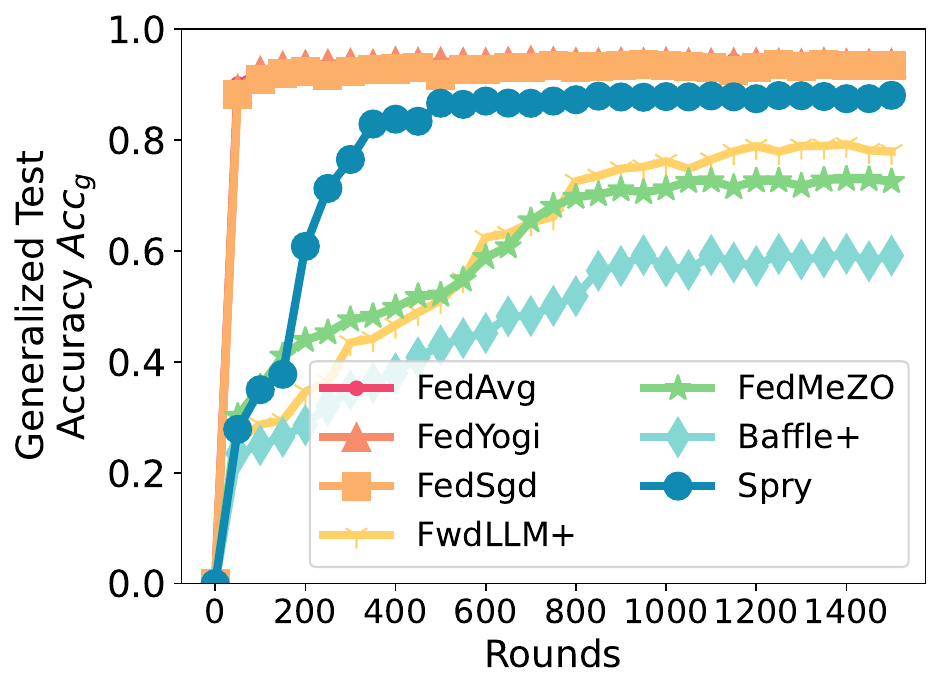}
         \caption{AG News with RoBERTa Large}
         \label{fig:ag-news-roberta-large-gen-homo}
     \end{subfigure}
     \hfill
     \begin{subfigure}[b]{0.32\textwidth}
         \centering
         \includegraphics[width=\textwidth]{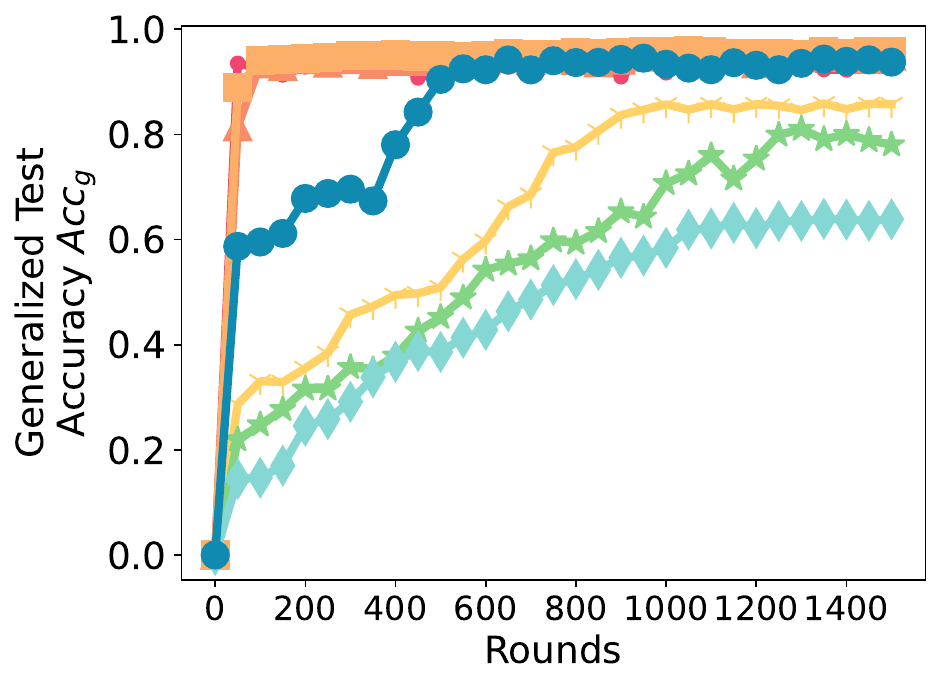}
         \caption{SST2 with RoBERTa Large}
         \label{fig:sst2-roberta-large-gen-homo}
     \end{subfigure}
     \hfill
     \begin{subfigure}[b]{0.32\textwidth}
         \centering
         \includegraphics[width=\textwidth]{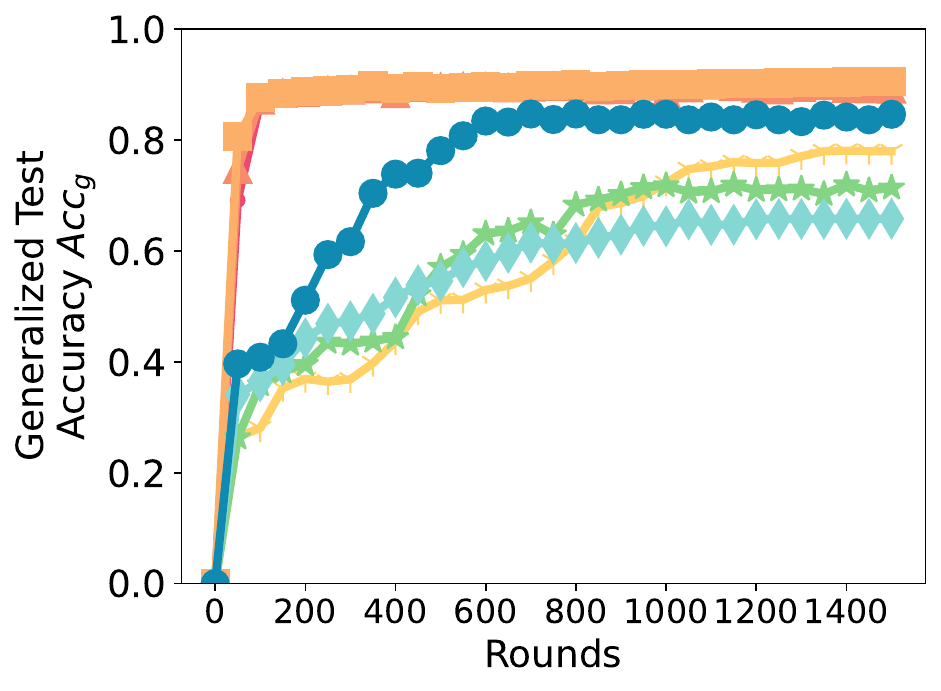}
         \caption{SNLI with RoBERTa Large}
         \label{fig:snli-roberta-large-gen-homo}
     \end{subfigure}
     \hfill
     \begin{subfigure}[b]{0.32\textwidth}
         \centering
         \includegraphics[width=\textwidth]{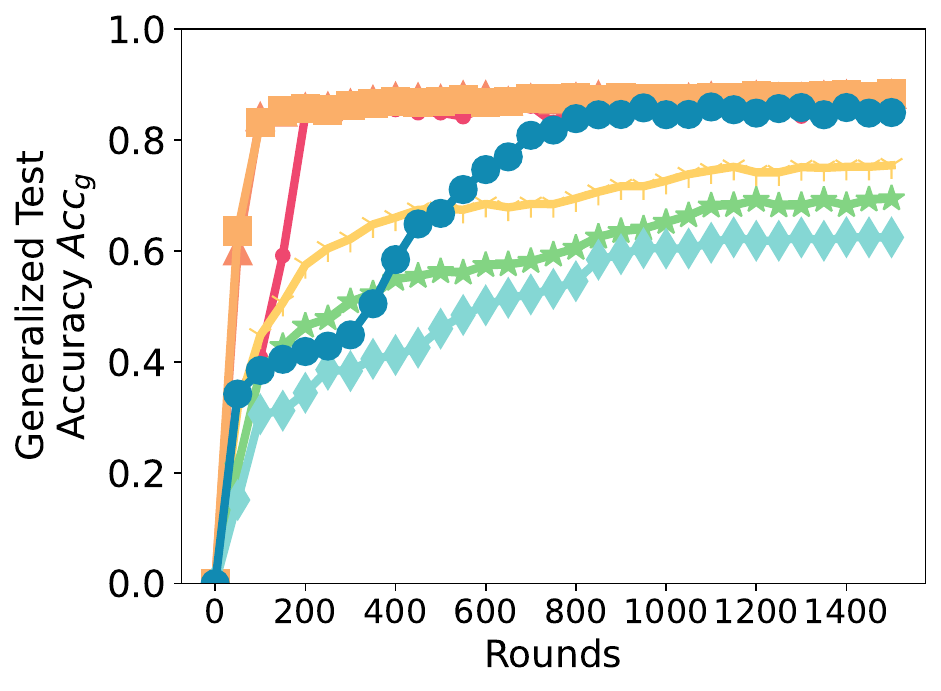}
         \caption{MNLI with RoBERTa Large}
         \label{fig:mnli-roberta-large-gen-homo}
     \end{subfigure}
     \hfill
     \begin{subfigure}[b]{0.32\textwidth}
         \centering
         \includegraphics[width=\textwidth]{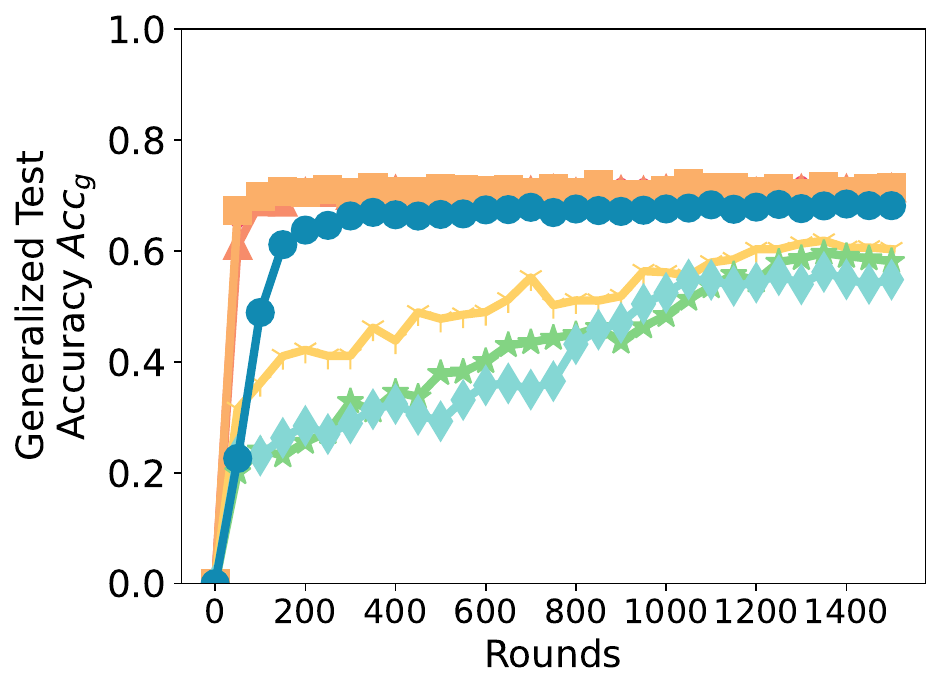}
         \caption{Yahoo with RoBERTa Large}
         \label{fig:yahoo-roberta-large-gen-homo}
     \end{subfigure}
     \hfill
     \begin{subfigure}[b]{0.32\textwidth}
         \centering
         \includegraphics[width=\textwidth]{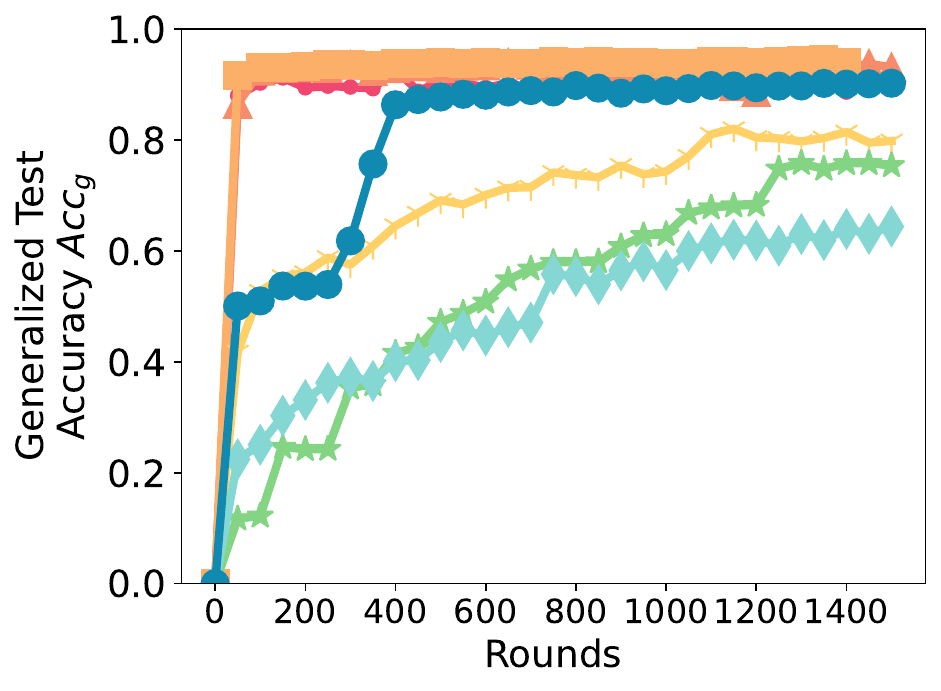}
         \caption{Yelp with RoBERTa Large}
         \label{fig:yelp-roberta-large-gen-homo}
     \end{subfigure}
     \hfill
     \begin{subfigure}[b]{0.32\textwidth}
         \centering
         \includegraphics[width=\textwidth]{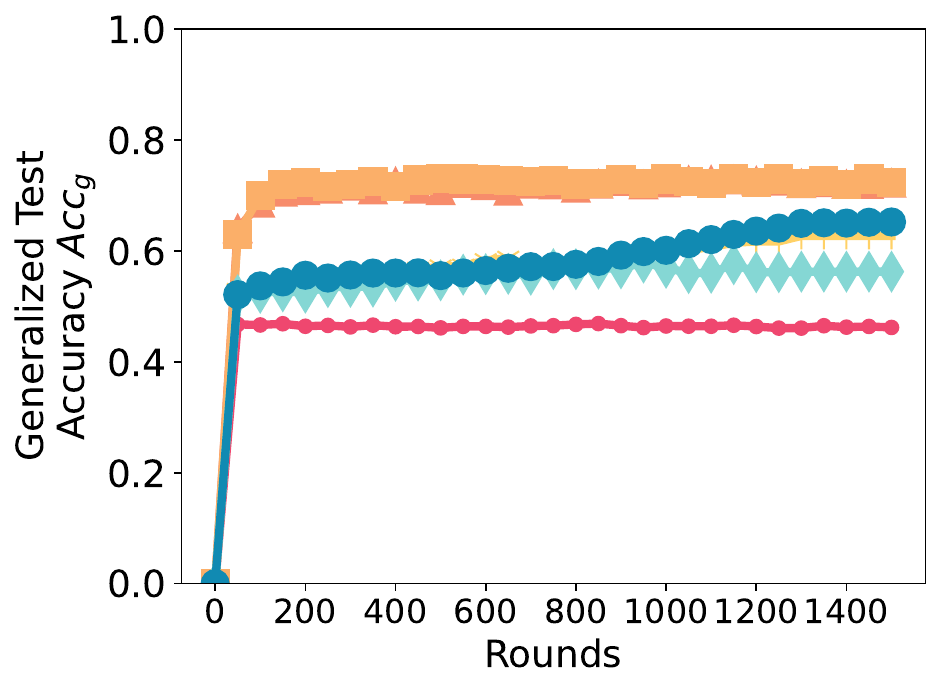}
         \caption{MultiRC with Llama2-7B}
         \label{fig:multirc-llama2-7B-gen-homo}
     \end{subfigure}
     \hfill
     \begin{subfigure}[b]{0.32\textwidth}
         \centering
         \includegraphics[width=\textwidth]{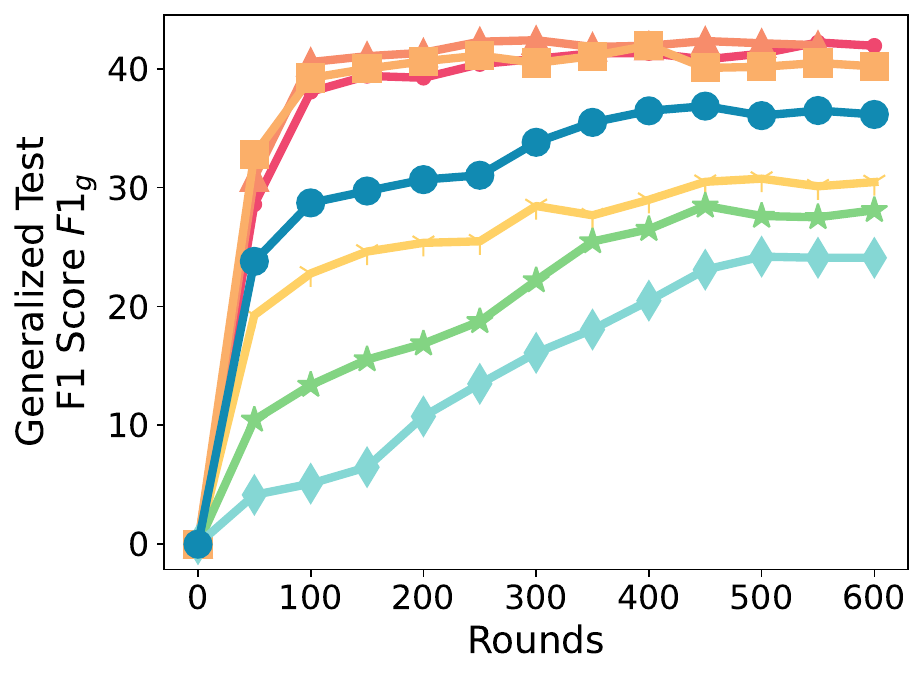}
         \caption{SQuADv2 with OPT6.7B}
         \label{fig:squadv2-opt-6.7B-gen-f1-homo}
     \end{subfigure}
     \hfill
     \begin{subfigure}[b]{0.32\textwidth}
         \centering
         \includegraphics[width=\textwidth]{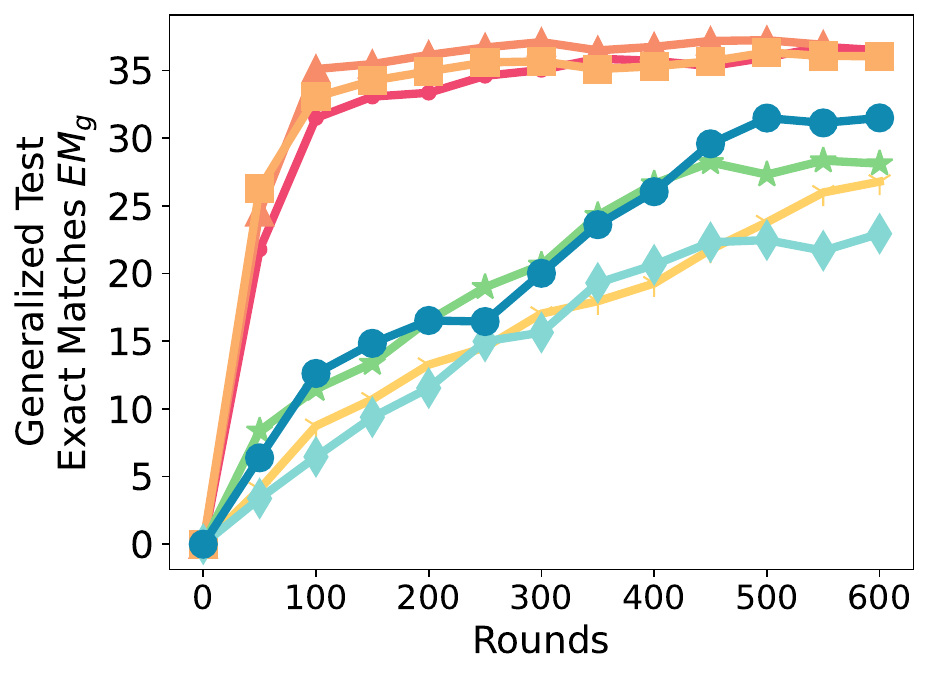}
         \caption{SQuADv2 with OPT6.7B}
         \label{fig:squadv2-opt-6.7B-gen-em-homo}
     \end{subfigure}
     \hfill
     \begin{subfigure}[b]{0.32\textwidth}
         \centering
         \includegraphics[width=\textwidth]{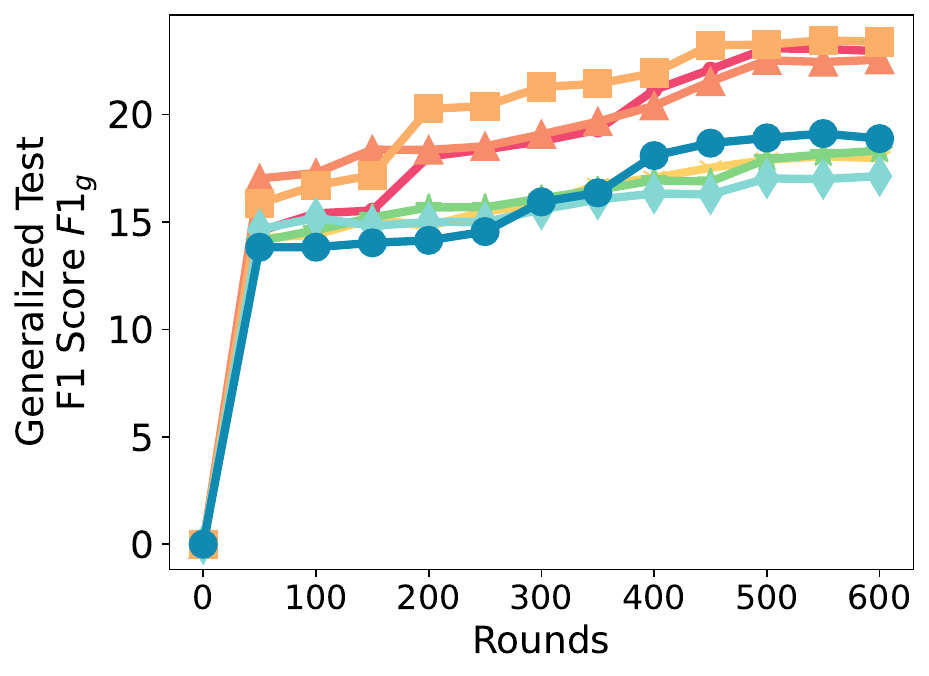}
         \caption{SQuADv2 with OPT13B}
         \label{fig:squadv2-opt-13B-gen-f1-homo}
     \end{subfigure}
     \hfill
     \begin{subfigure}[b]{0.32\textwidth}
         \centering
         \includegraphics[width=\textwidth]{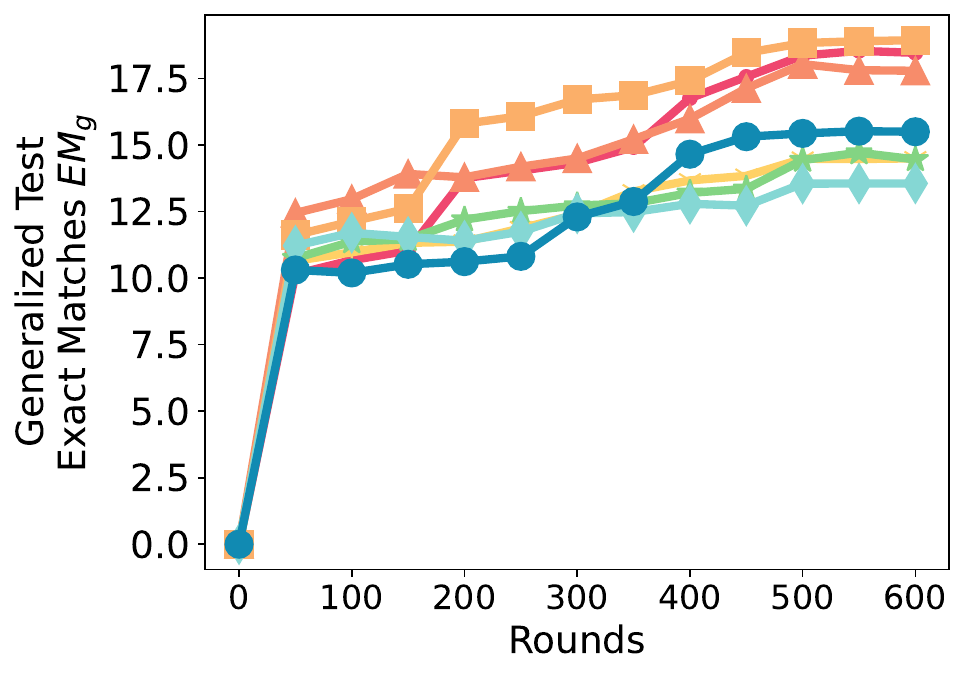}
         \caption{SQuADv2 with OPT13B}
         \label{fig:squadv2-opt-13B-gen-em-homo}
     \end{subfigure}
        \caption{Generalized accuracy / F1 score / Exact matches for\\ homogeneous clients (Dirichlet $\alpha=1.0$) setting}
        \label{fig:generalized-accu-homo}
\end{figure}

\begin{figure}[h]
     \centering
     \begin{subfigure}[b]{0.32\textwidth}
         \centering
         \includegraphics[width=\textwidth]{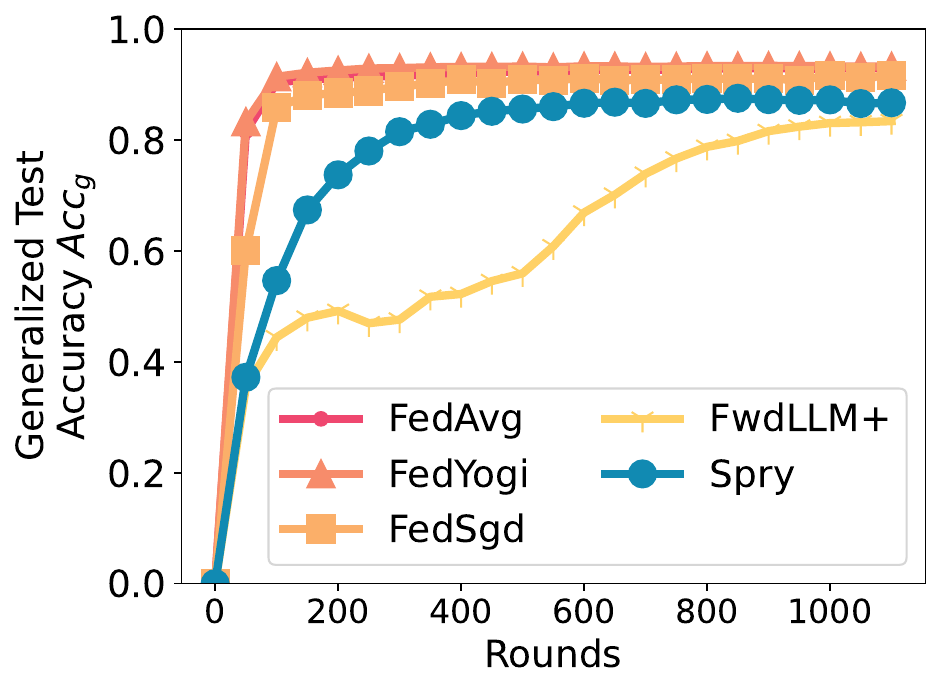}
         \caption{AG News with BERT Base}
         \label{fig:ag-news-bert-base-gen-homo}
     \end{subfigure}
     \hfill
     \begin{subfigure}[b]{0.32\textwidth}
         \centering
         \includegraphics[width=\textwidth]{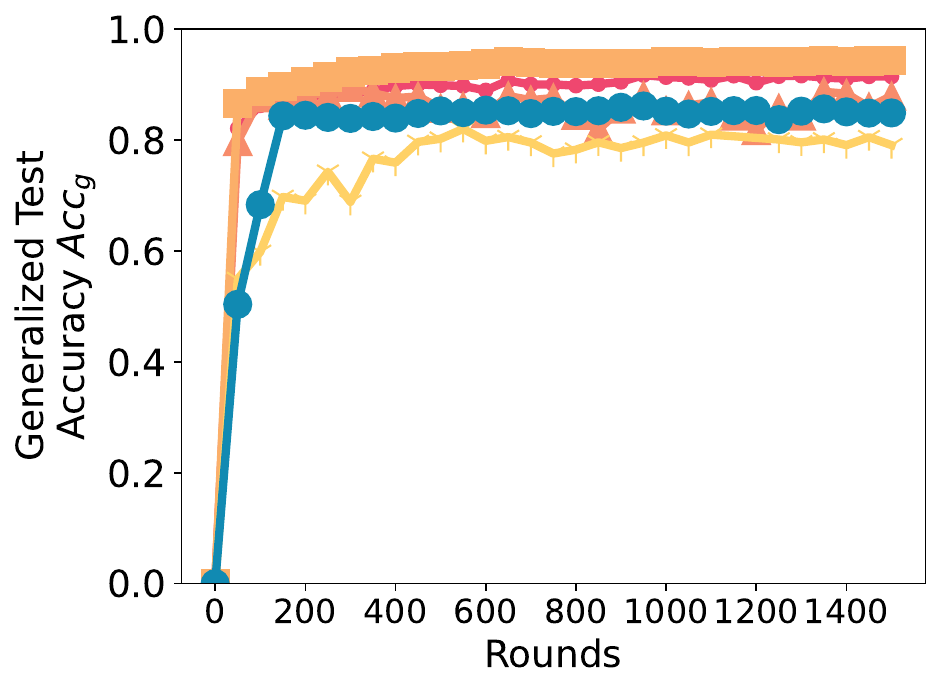}
         \caption{SST2 with DistilBERT Base}
         \label{fig:sst2-distilbert-base-gen-homo}
     \end{subfigure}
     \hfill
     \begin{subfigure}[b]{0.32\textwidth}
         \centering
         \includegraphics[width=\textwidth]{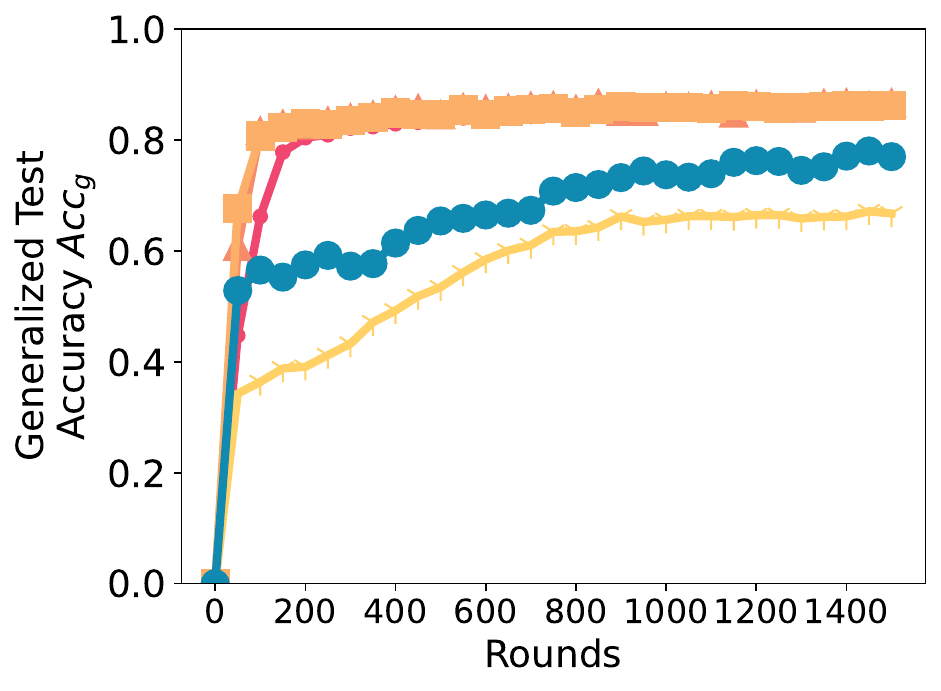}
         \caption{SNLI with BERT Large}
         \label{fig:snli-bert-large-gen-homo}
     \end{subfigure}
     \hfill
     \begin{subfigure}[b]{0.32\textwidth}
         \centering
         \includegraphics[width=\textwidth]{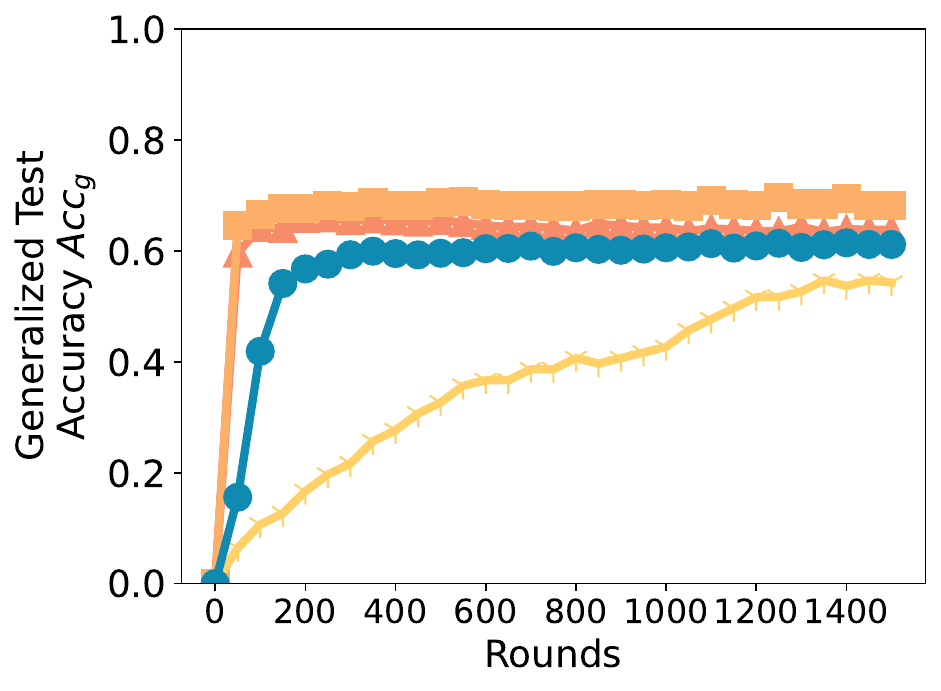}
         \caption{Yahoo with DistilBERT Base}
         \label{fig:yahoo-distilbert-base-gen-homo}
     \end{subfigure}
     \hfill
     \begin{subfigure}[b]{0.32\textwidth}
         \centering
         \includegraphics[width=\textwidth]{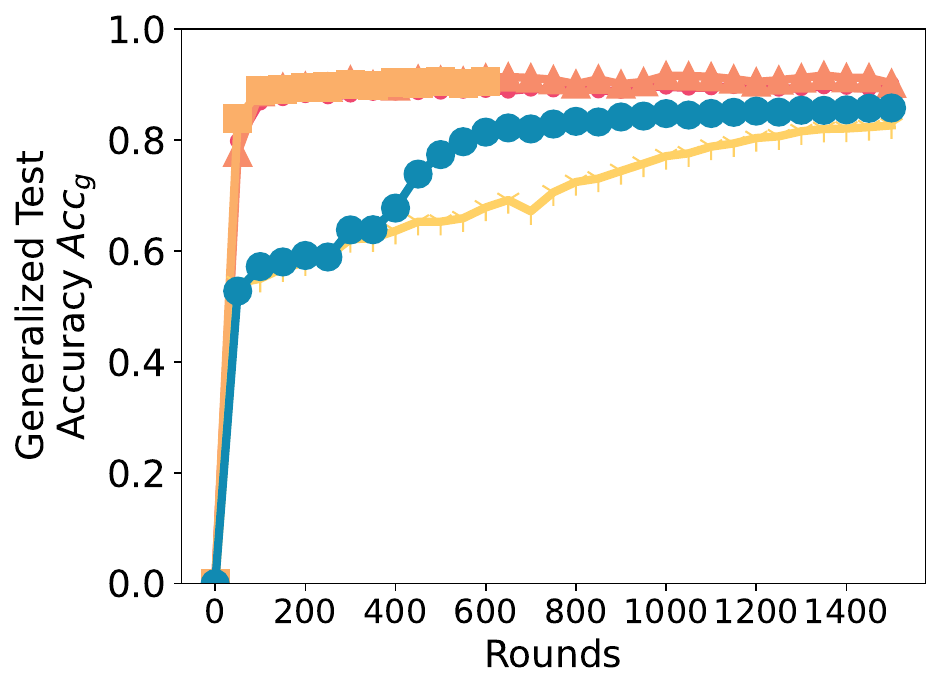}
         \caption{Yelp with Albert Large v2}
         \label{fig:yelp-albertv2-large-gen-homo}
     \end{subfigure}
        \caption{Generalized accuracy /F1 score / Exact matches for\\ homogeneous clients (Dirichlet $\alpha=1.0$) setting for a variety of language models}
        \label{fig:generalized-accu-homo-various-models}
\end{figure}

\begin{figure}[h]
     \centering
     \begin{subfigure}[b]{0.32\textwidth}
         \centering
         \includegraphics[width=\textwidth]{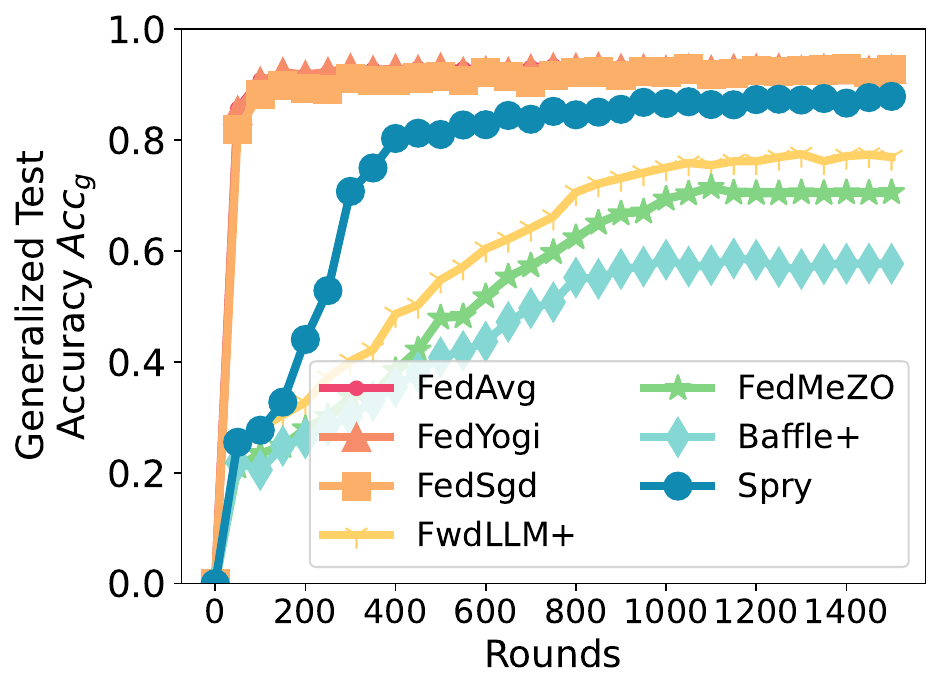}
         \caption{AG News with RoBERTa Large}
         \label{fig:ag-news-roberta-large-gen-hetero}
     \end{subfigure}
     \hfill
     \begin{subfigure}[b]{0.32\textwidth}
         \centering
         \includegraphics[width=\textwidth]{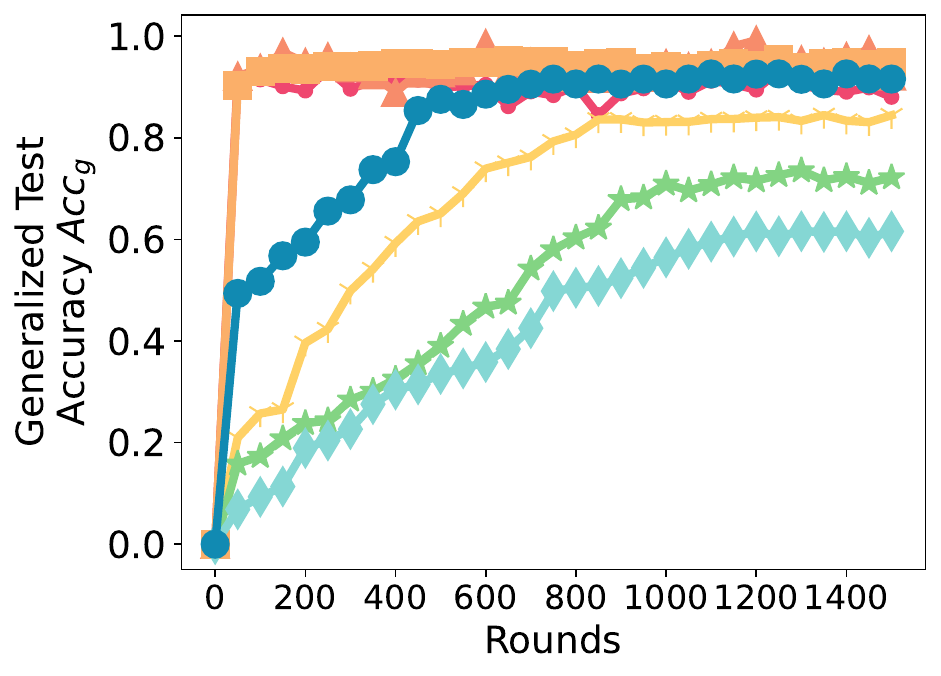}
         \caption{SST2 with RoBERTa Large}
         \label{fig:sst2-roberta-large-gen-hetero}
     \end{subfigure}
     \hfill
     \begin{subfigure}[b]{0.32\textwidth}
         \centering
         \includegraphics[width=\textwidth]{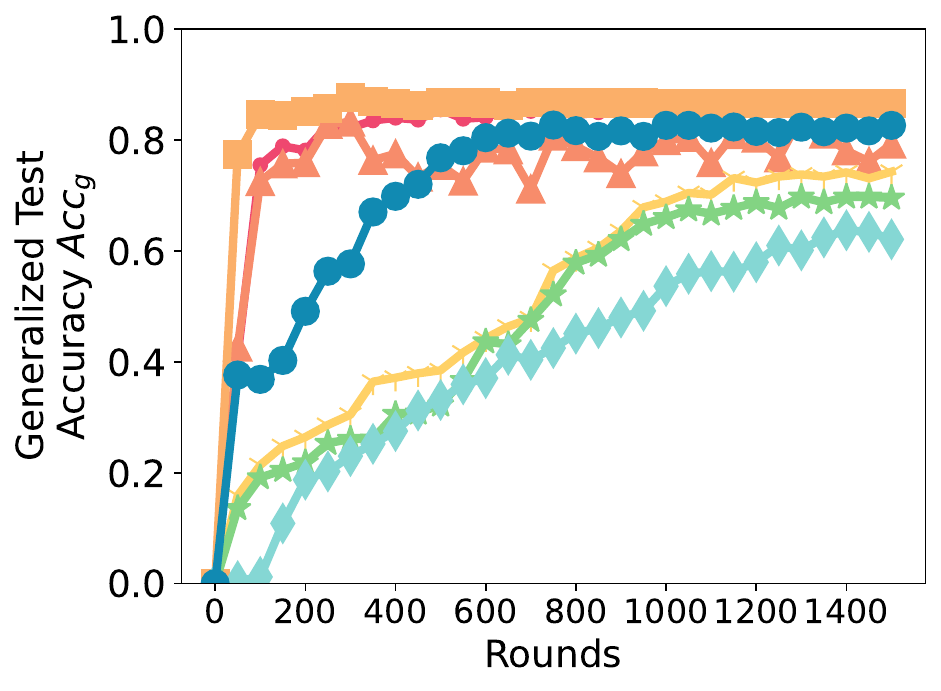}
         \caption{SNLI with RoBERTa Large}
         \label{fig:snli-roberta-large-gen-hetero}
     \end{subfigure}
     \hfill
     \begin{subfigure}[b]{0.32\textwidth}
         \centering
         \includegraphics[width=\textwidth]{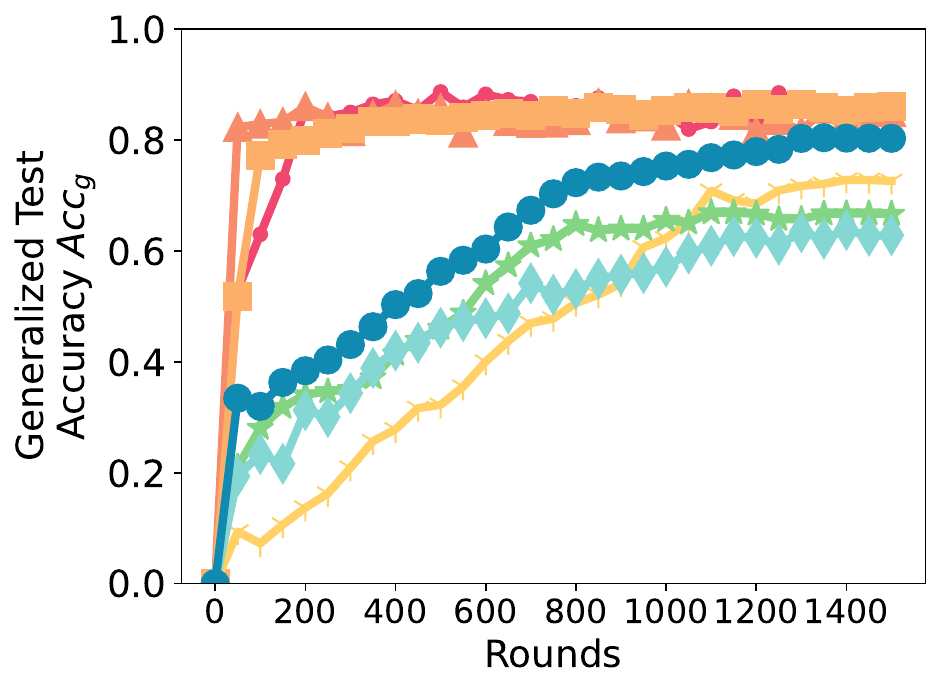}
         \caption{MNLI with RoBERTa Large}
         \label{fig:mnli-roberta-large-gen-hetero}
     \end{subfigure}
     \hfill
     \begin{subfigure}[b]{0.32\textwidth}
         \centering
         \includegraphics[width=\textwidth]{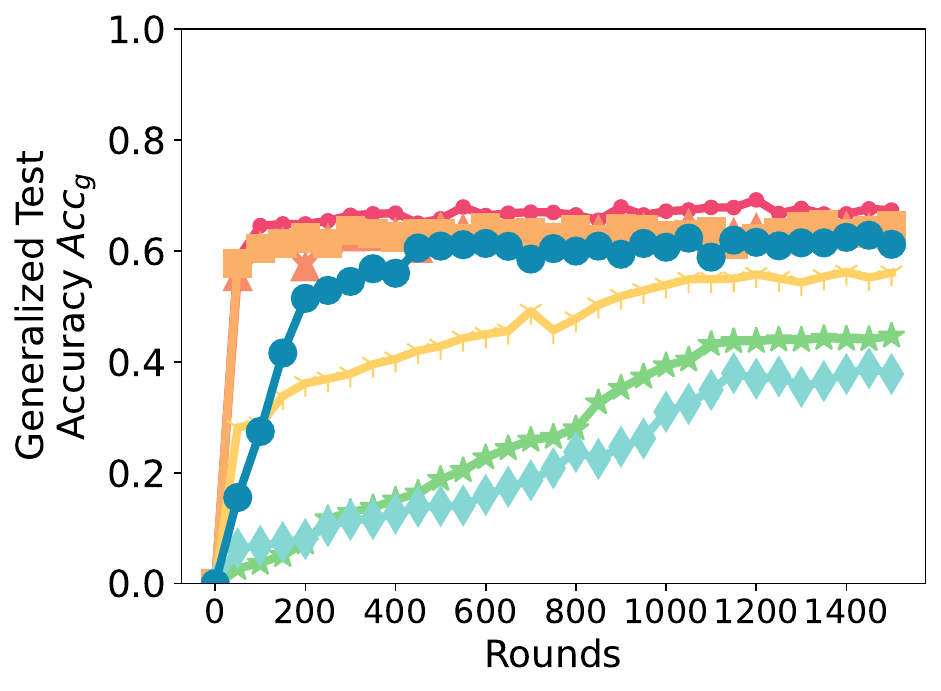}
         \caption{Yahoo with RoBERTa Large}
         \label{fig:yahoo-roberta-large-gen-hetero}
     \end{subfigure}
     \hfill
     \begin{subfigure}[b]{0.32\textwidth}
         \centering
         \includegraphics[width=\textwidth]{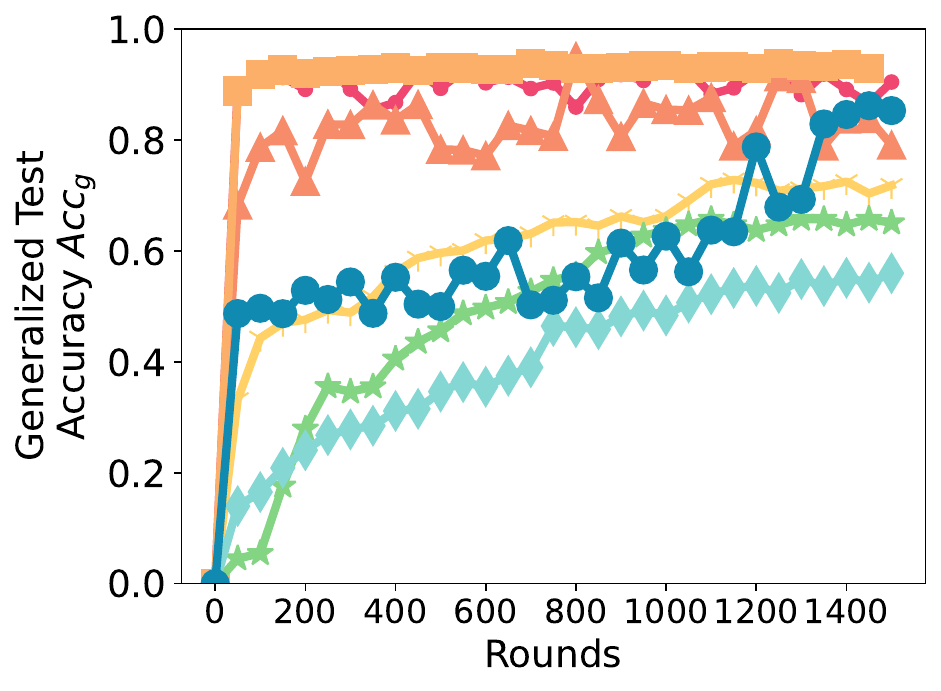}
         \caption{Yelp with RoBERTa Large}
         \label{fig:yelp-roberta-large-gen-hetero}
     \end{subfigure}
     \hfill
     \begin{subfigure}[b]{0.32\textwidth}
         \centering
         \includegraphics[width=\textwidth]{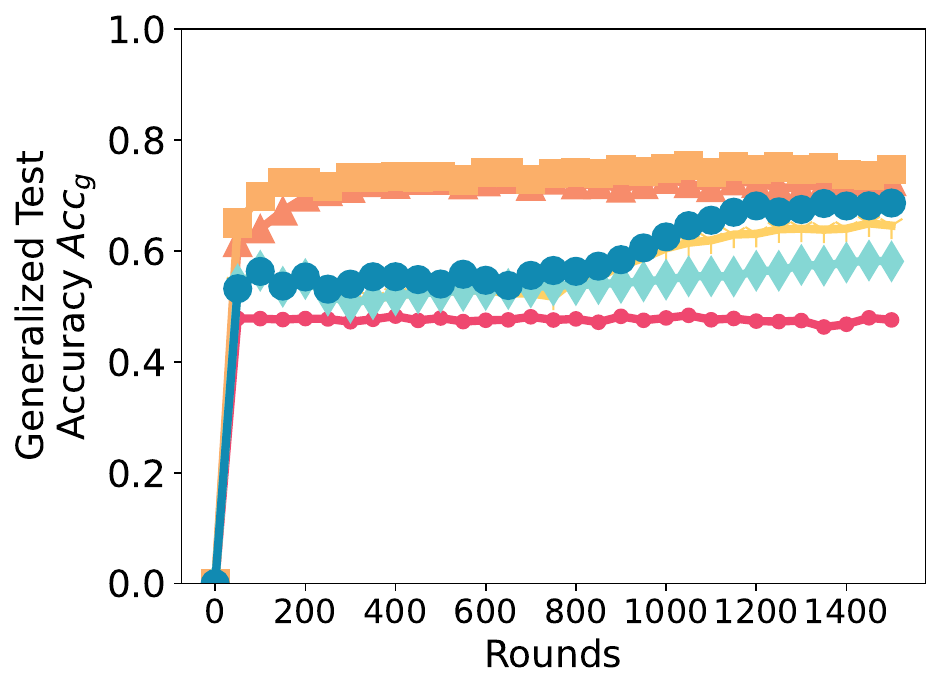}
         \caption{MultiRC with Llama2-7B}
         \label{fig:multirc-llama2-7B-gen-hetero}
     \end{subfigure}
     \hfill
     \begin{subfigure}[b]{0.32\textwidth}
         \centering
         \includegraphics[width=\textwidth]{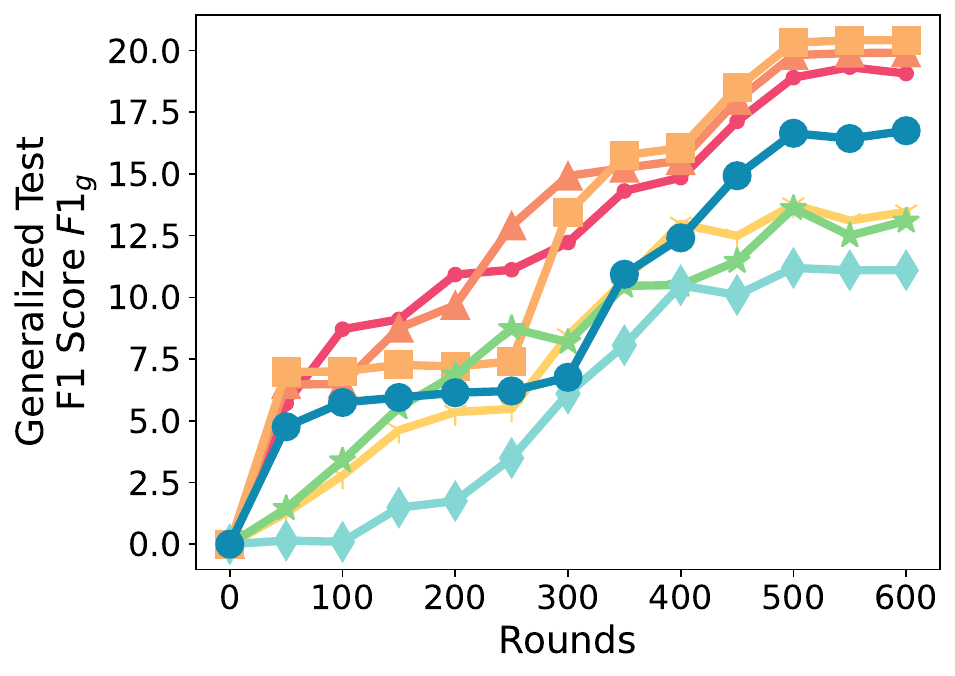}
         \caption{SQuADv2 with OPT6.7B}
         \label{fig:squadv2-opt-6.7B-gen-f1-hetero}
     \end{subfigure}
     \hfill
     \begin{subfigure}[b]{0.32\textwidth}
         \centering
         \includegraphics[width=\textwidth]{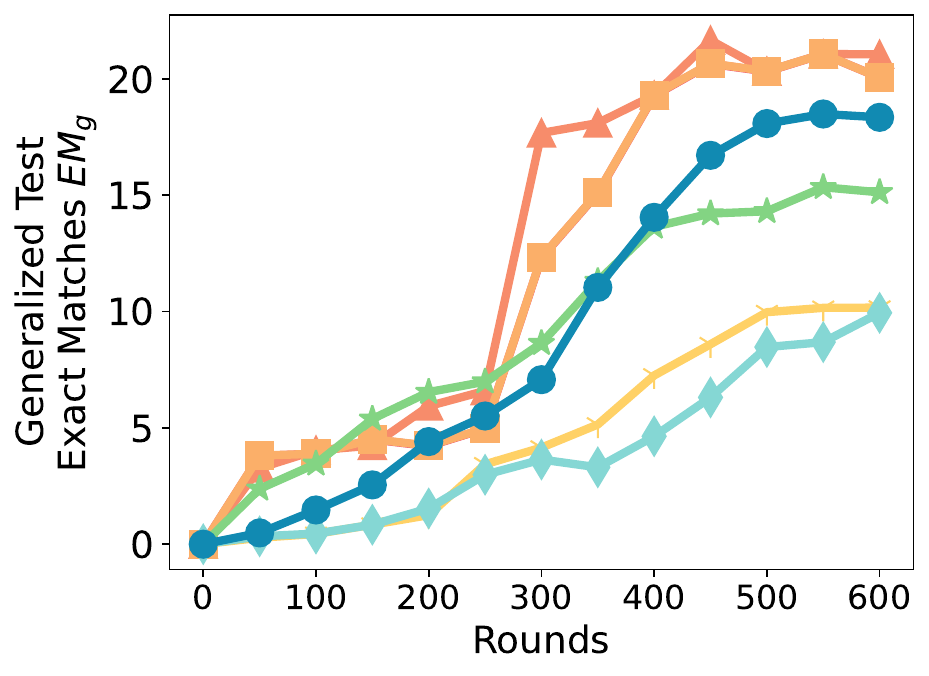}
         \caption{SQuADv2 with OPT6.7B}
         \label{fig:squadv2-opt-6.7B-gen-em-hetero}
     \end{subfigure}
     \hfill
     \begin{subfigure}[b]{0.32\textwidth}
         \centering
         \includegraphics[width=\textwidth]{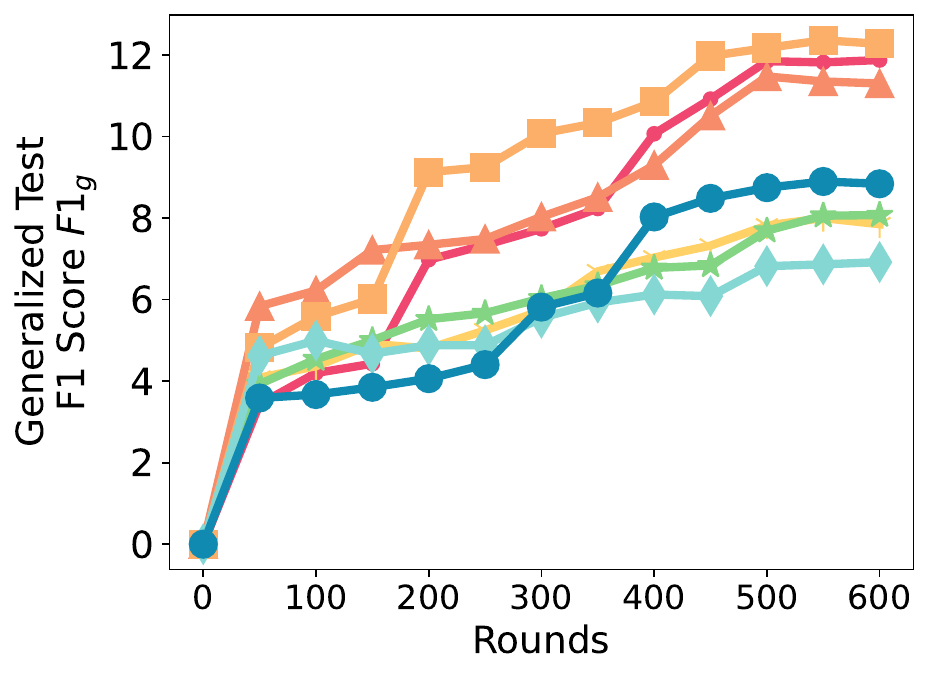}
         \caption{SQuADv2 with OPT13B}
         \label{fig:squadv2-opt-13B-gen-f1-hetero}
     \end{subfigure}
     \hfill
     \begin{subfigure}[b]{0.32\textwidth}
         \centering
         \includegraphics[width=\textwidth]{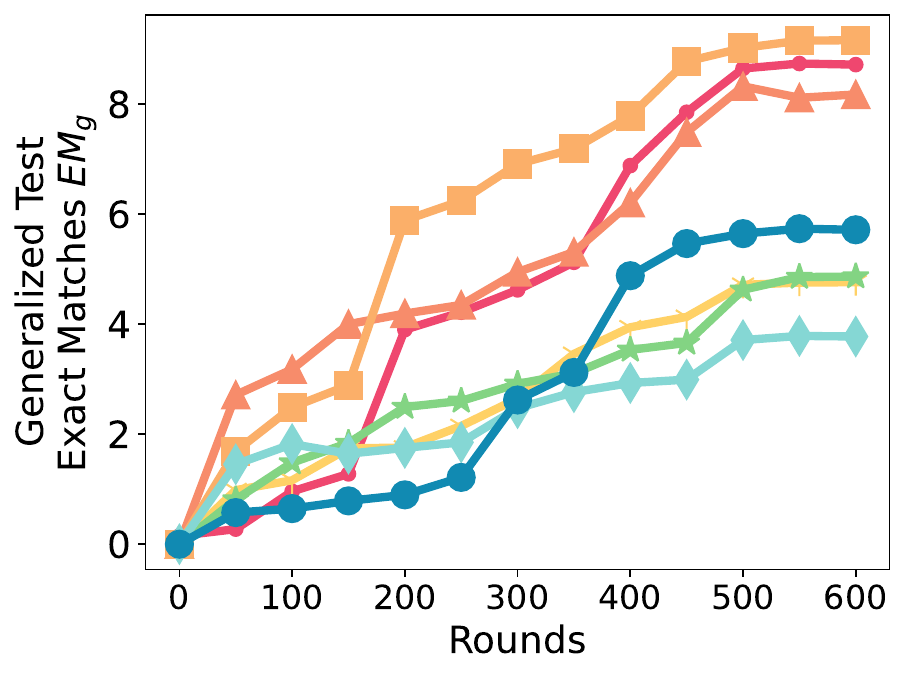}
         \caption{SQuADv2 with OPT13B}
         \label{fig:squadv2-opt-13B-gen-em-hetero}
     \end{subfigure}
        \caption{Generalized accuracy / F1 score / Exact matches for\\ heterogeneous clients (Dirichlet $\alpha=0.1$) setting}
        \label{fig:generalized-accu-hetero}
\end{figure}

Generalized results on homogeneous and heterogeneous clients with Dir $\alpha=1.0$ and $\alpha=0.1$ are shown in (a)~Figures~\ref{fig:generalized-accu-homo} and~\ref{fig:generalized-accu-hetero} for RoBERTa Large, Llama2-7B, OPT6.7B, OPT13B; and (b)~Figure~\ref{fig:generalized-accu-homo-various-models} for BERT Large, BERT Base, DistilBert Base, Albert Large v2.

\clearpage
\subsection{Personalized Performance Curves}
\begin{figure}[h]
     \centering
     \begin{subfigure}[b]{0.32\textwidth}
         \centering
         \includegraphics[width=\textwidth]{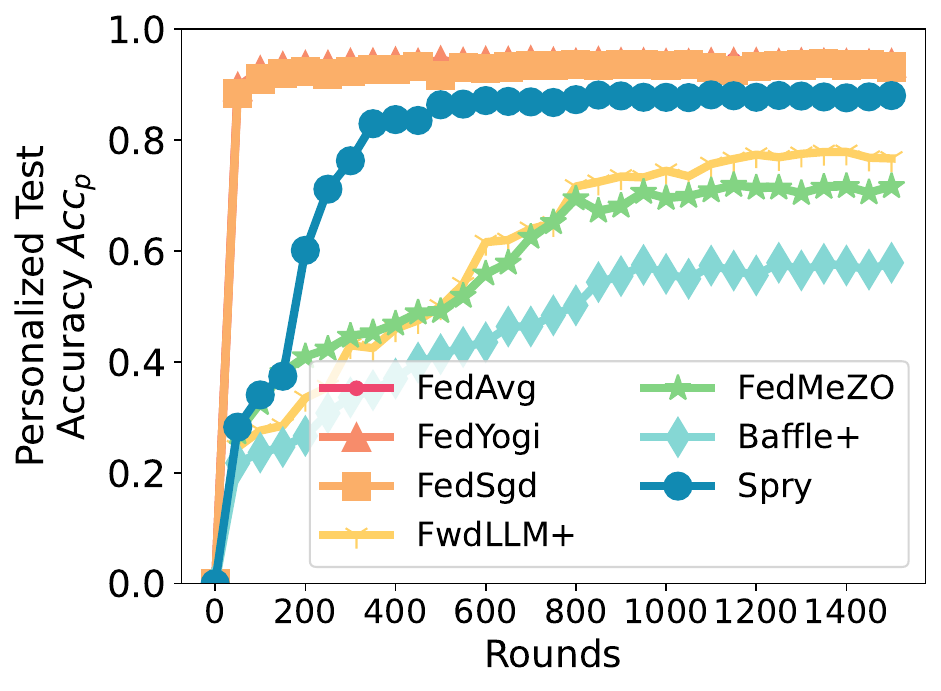}
         \caption{AG News with RoBERTa Large}
         \label{fig:ag-news-roberta-large-pers-homo}
     \end{subfigure}
     \hfill
     \begin{subfigure}[b]{0.32\textwidth}
         \centering
         \includegraphics[width=\textwidth]{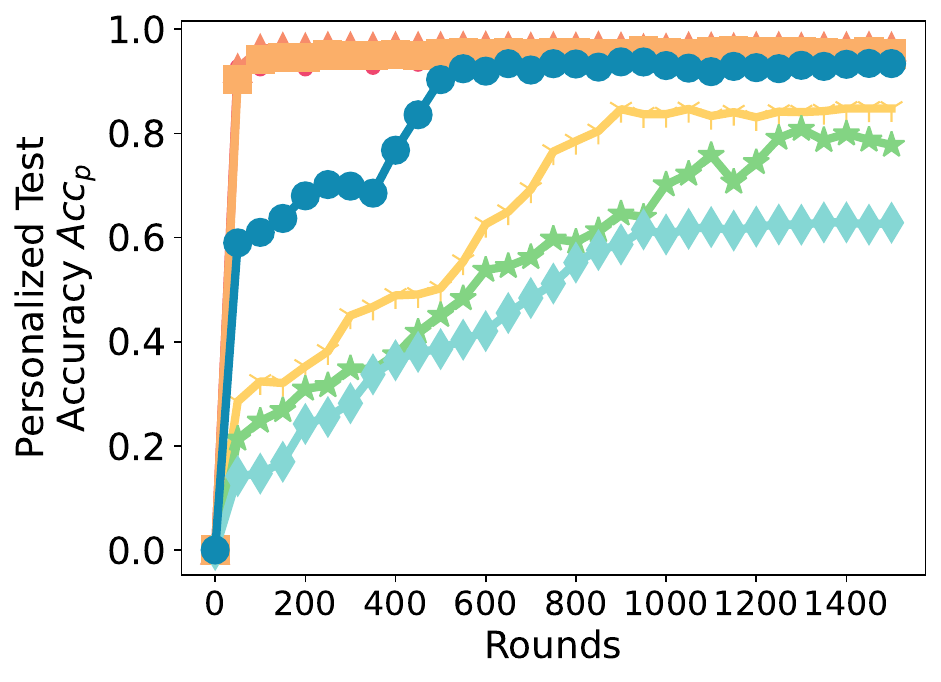}
         \caption{SST2 with RoBERTa Large}
         \label{fig:sst2-roberta-large-pers-homo}
     \end{subfigure}
     \hfill
     \begin{subfigure}[b]{0.32\textwidth}
         \centering
         \includegraphics[width=\textwidth]{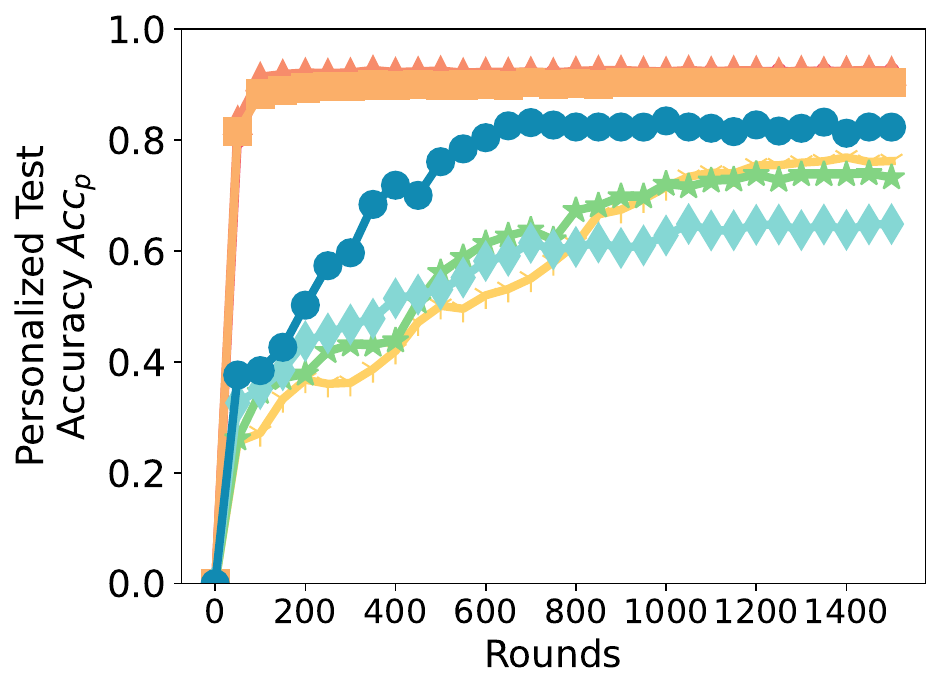}
         \caption{SNLI with RoBERTa Large}
         \label{fig:snli-roberta-large-pers-homo}
     \end{subfigure}
     \hfill
     \begin{subfigure}[b]{0.32\textwidth}
         \centering
         \includegraphics[width=\textwidth]{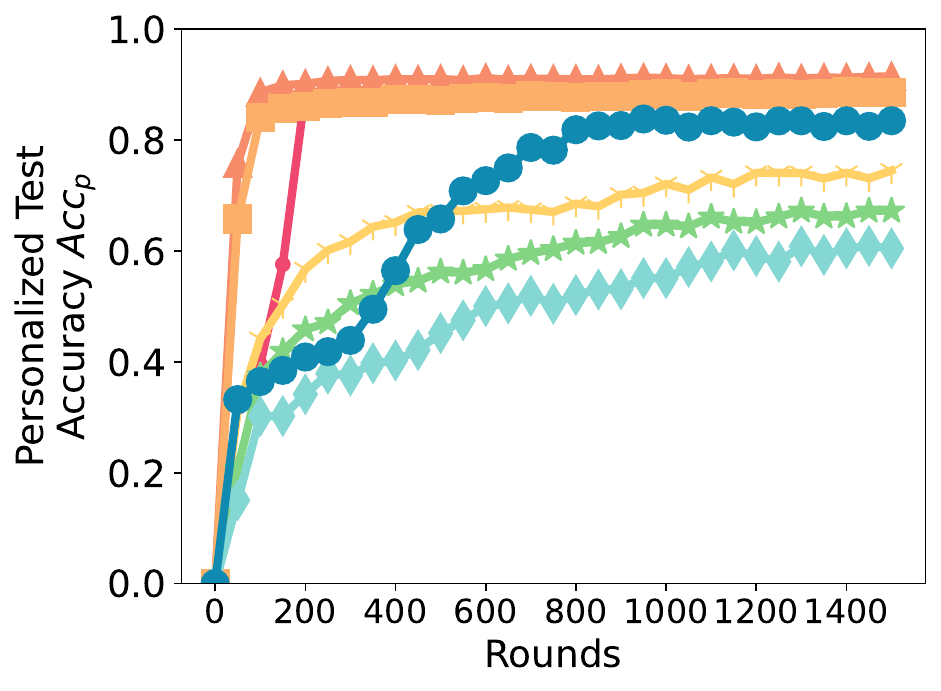}
         \caption{MNLI with RoBERTa Large}
         \label{fig:mnli-roberta-large-pers-homo}
     \end{subfigure}
     \hfill
     \begin{subfigure}[b]{0.32\textwidth}
         \centering
         \includegraphics[width=\textwidth]{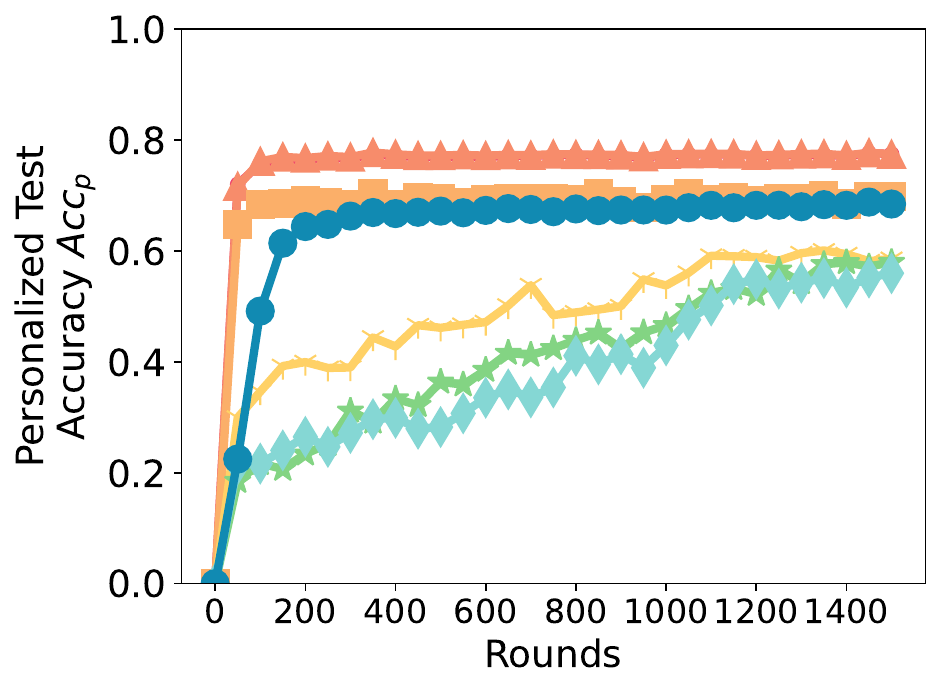}
         \caption{Yahoo with RoBERTa Large}
         \label{fig:yahoo-roberta-large-pers-homo}
     \end{subfigure}
     \hfill
     \begin{subfigure}[b]{0.32\textwidth}
         \centering
         \includegraphics[width=\textwidth]{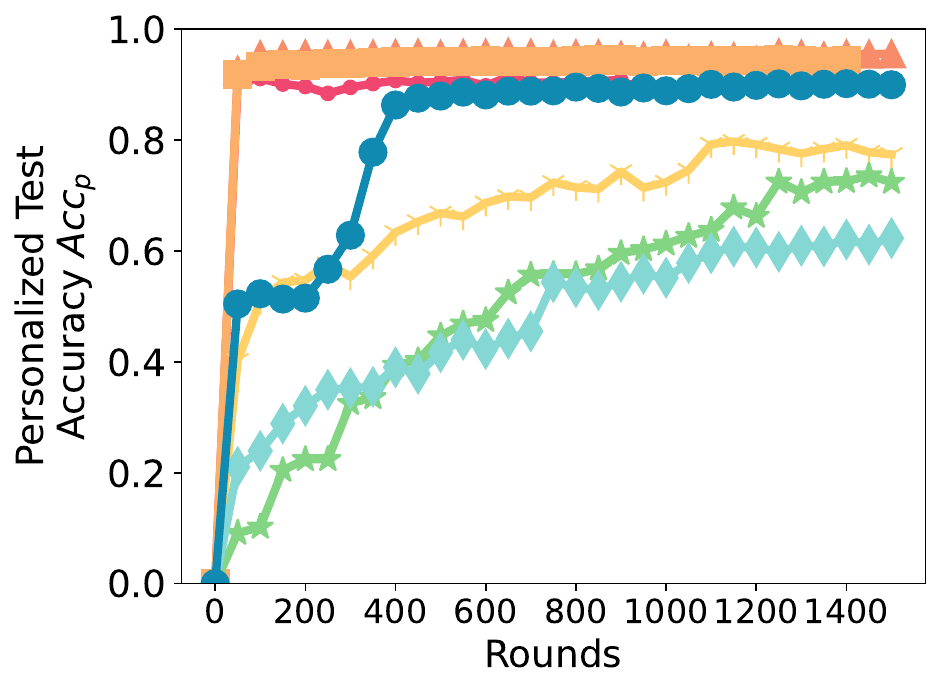}
         \caption{Yelp with RoBERTa Large}
         \label{fig:yelp-roberta-large-pers-homo}
     \end{subfigure}
     \hfill
     \begin{subfigure}[b]{0.32\textwidth}
         \centering
         \includegraphics[width=\textwidth]{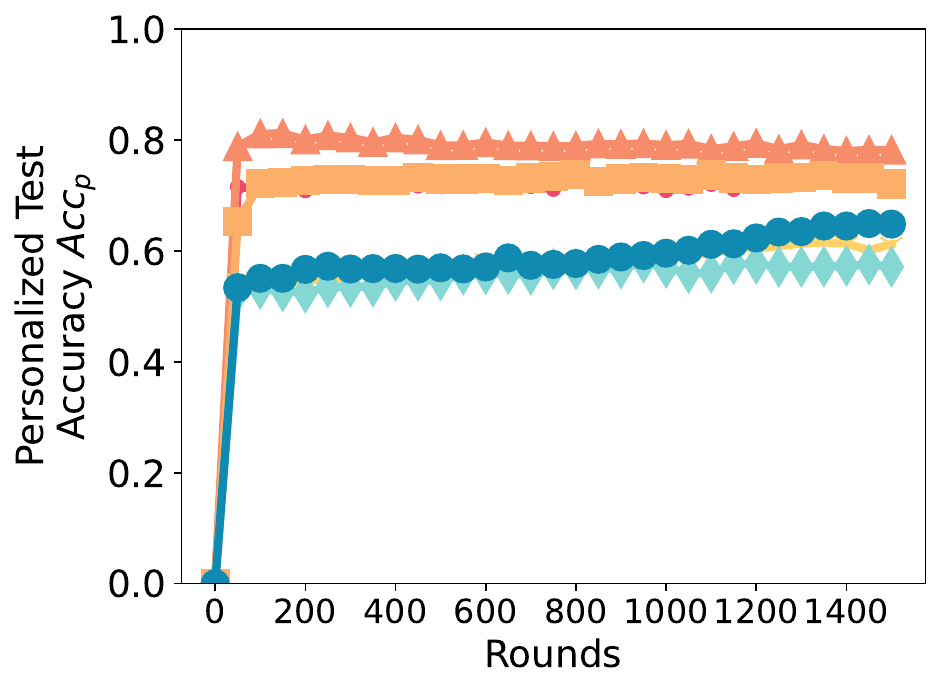}
         \caption{MultiRC with Llama2-7B}
         \label{fig:multirc-llama2-7B-pers-homo}
     \end{subfigure}
     \hfill
     \begin{subfigure}[b]{0.32\textwidth}
         \centering
         \includegraphics[width=\textwidth]{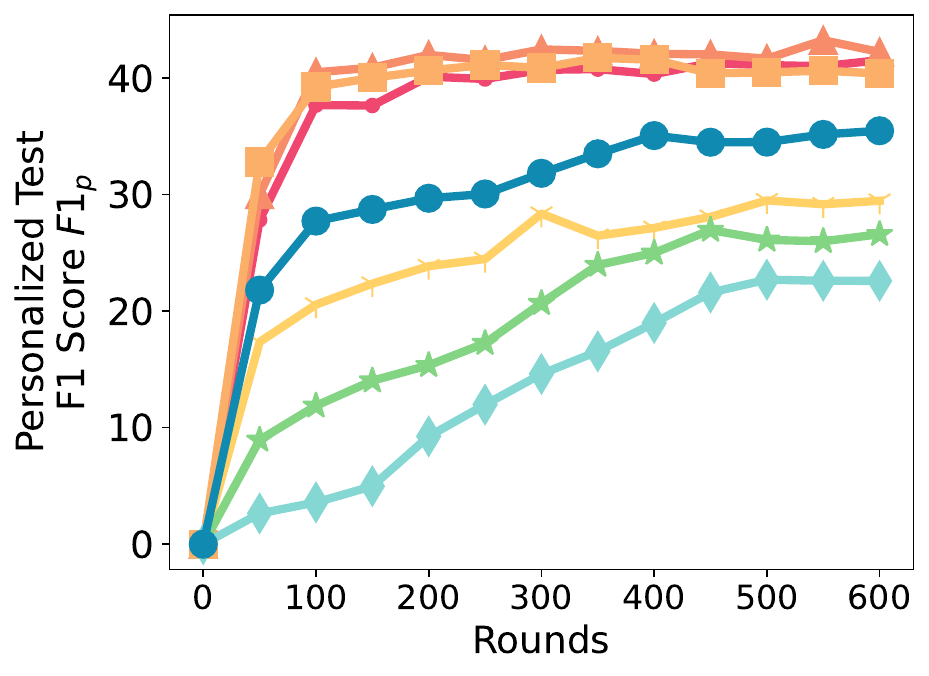}
         \caption{SQuADv2 with OPT6.7B}
         \label{fig:squadv2-opt-6.7B-pers-f1-homo}
     \end{subfigure}
     \hfill
     \begin{subfigure}[b]{0.32\textwidth}
         \centering
         \includegraphics[width=\textwidth]{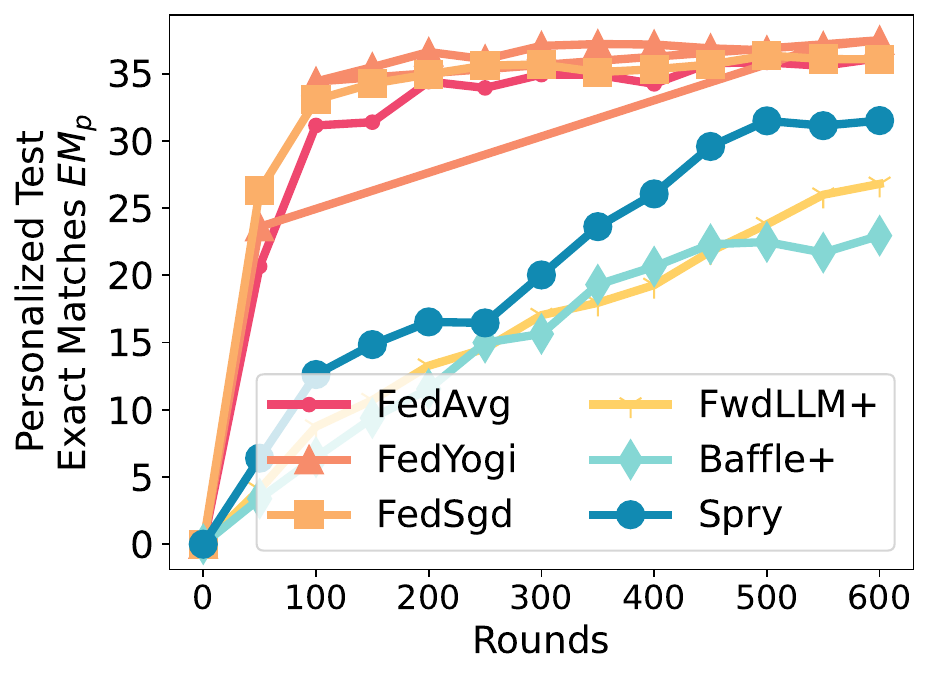}
         \caption{SQuADv2 with OPT6.7B}
         \label{fig:squadv2-opt-6.7B-pers-em-homo}
     \end{subfigure}
     \hfill
     \begin{subfigure}[b]{0.32\textwidth}
         \centering
         \includegraphics[width=\textwidth]{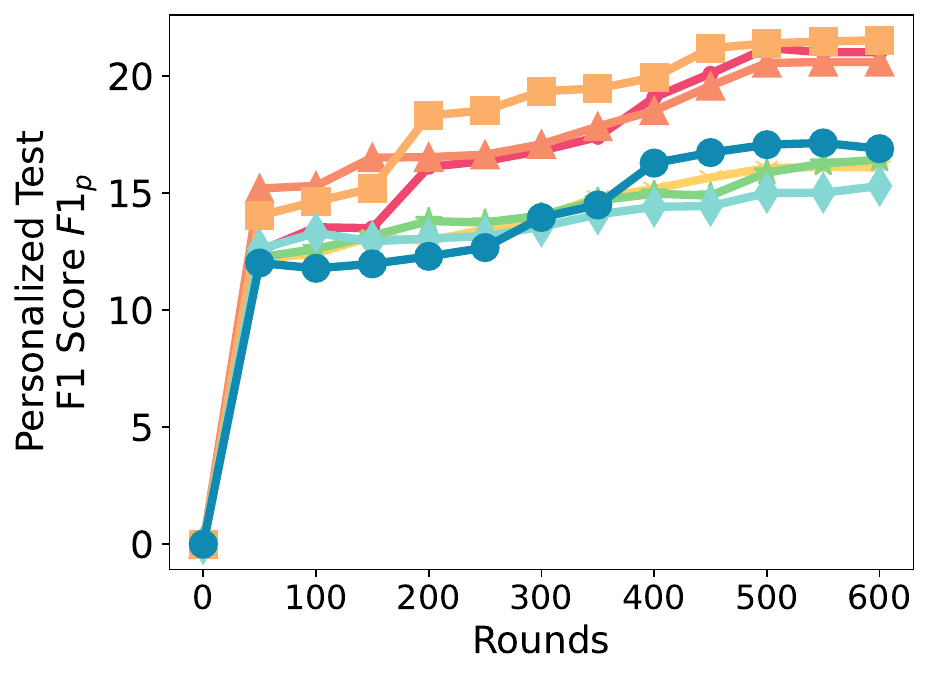}
         \caption{SQuADv2 with OPT13B}
         \label{fig:squadv2-opt-13B-pers-f1-homo}
     \end{subfigure}
     \hfill
     \begin{subfigure}[b]{0.32\textwidth}
         \centering
         \includegraphics[width=\textwidth]{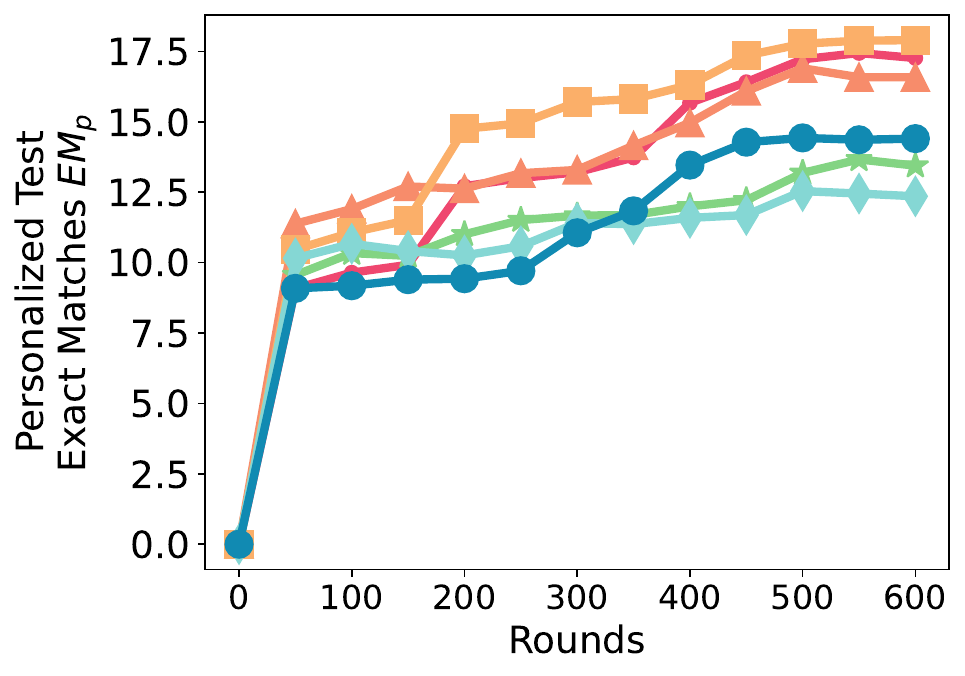}
         \caption{SQuADv2 with OPT13B}
         \label{fig:squadv2-opt-13B-pers-em-homo}
     \end{subfigure}
        \caption{Personalized accuracy / F1 score / Exact matches for\\ homogeneous clients (Dirichlet $\alpha=1.0$) setting}
        \label{fig:personalized-accu-homo}
\end{figure}

\begin{figure}[h]
     \centering
     \begin{subfigure}[b]{0.32\textwidth}
         \centering
         \includegraphics[width=\textwidth]{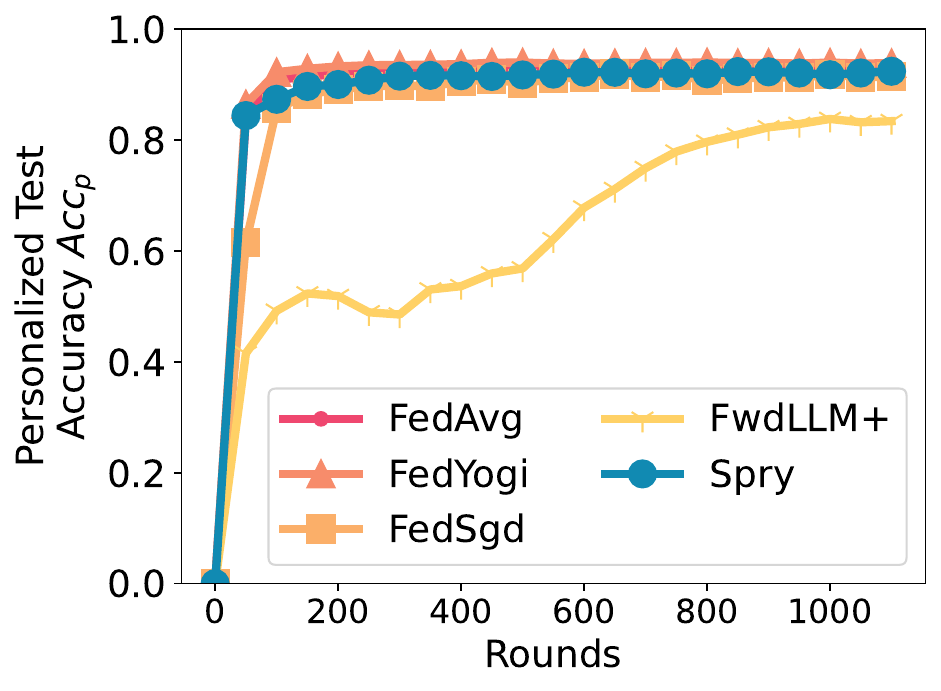}
         \caption{AG News with BERT Base}
         \label{fig:ag-news-bert-base-pers-homo}
     \end{subfigure}
     \hfill
     \begin{subfigure}[b]{0.32\textwidth}
         \centering
         \includegraphics[width=\textwidth]{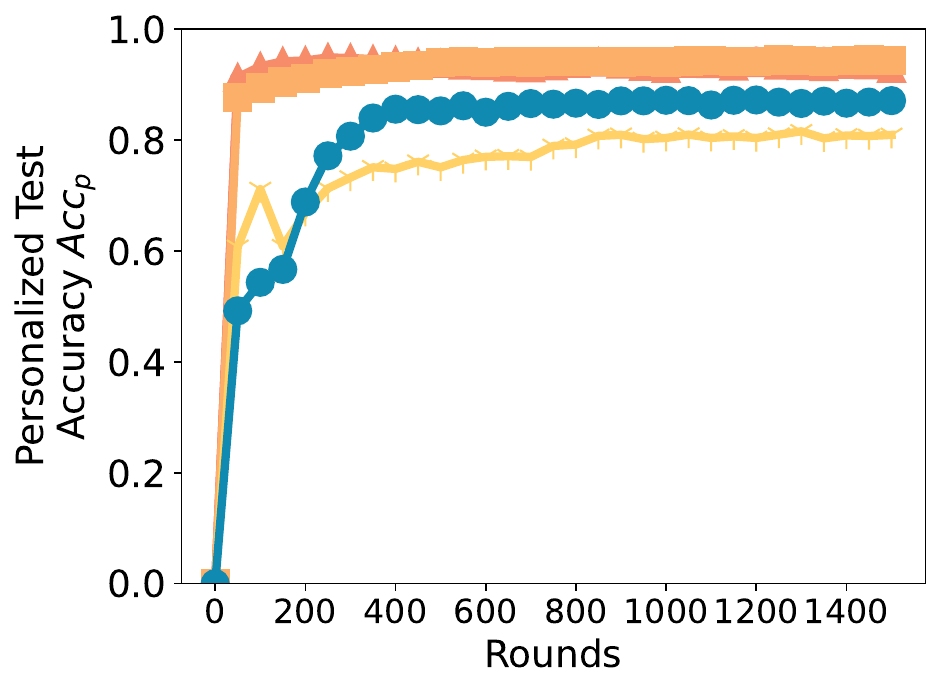}
         \caption{SST2 with DistilBERT Base}
         \label{fig:sst2-distilbert-base-pers-homo}
     \end{subfigure}
     \hfill
     \begin{subfigure}[b]{0.32\textwidth}
         \centering
         \includegraphics[width=\textwidth]{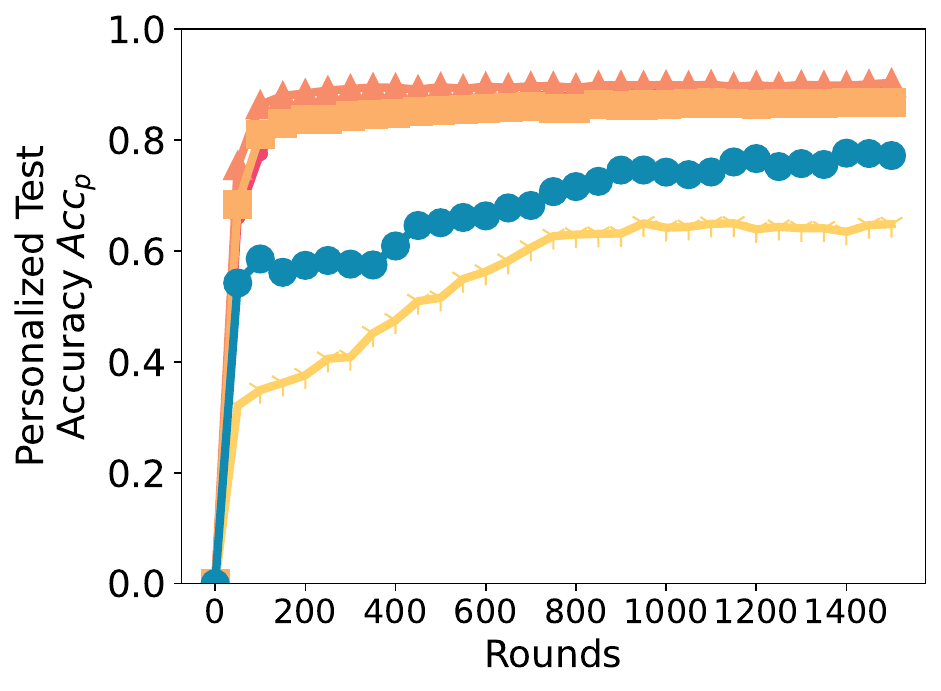}
         \caption{SNLI with BERT Large}
         \label{fig:snli-bert-large-pers-homo}
     \end{subfigure}
     \hfill
     \begin{subfigure}[b]{0.32\textwidth}
         \centering
         \includegraphics[width=\textwidth]{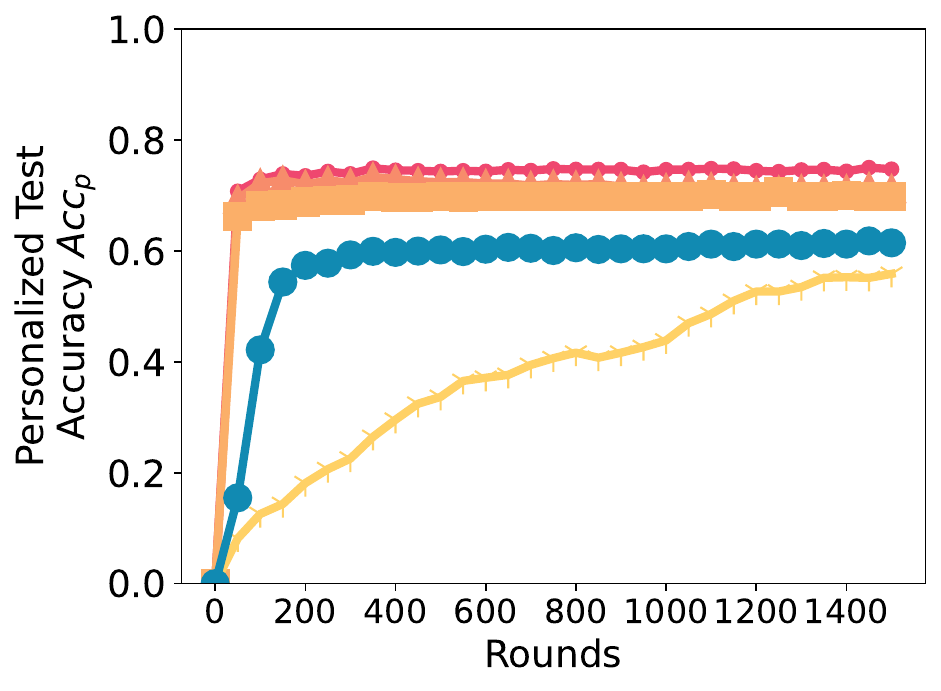}
         \caption{Yahoo with DistilBERT Base}
         \label{fig:yahoo-distilbert-base-pers-homo}
     \end{subfigure}
     \hfill
     \begin{subfigure}[b]{0.32\textwidth}
         \centering
         \includegraphics[width=\textwidth]{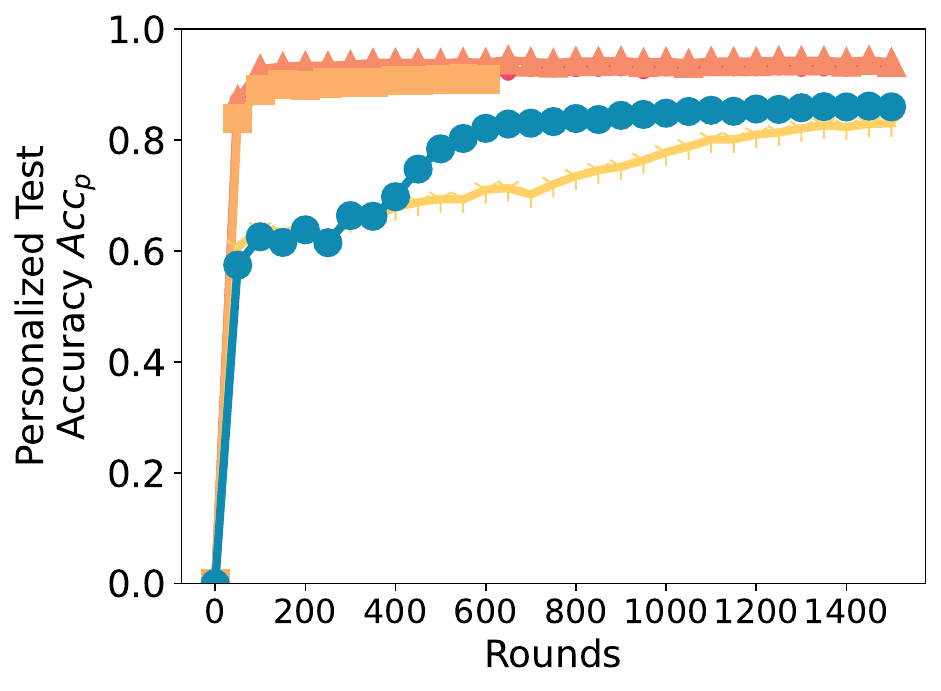}
         \caption{Yelp with Albert Large v2}
         \label{fig:yelp-albertv2-large-pers-homo}
     \end{subfigure}
        \caption{Personalized accuracy / F1 score / Exact matches for\\ homogeneous clients (Dirichlet $\alpha=1.0$) setting for a variety of language models}
        \label{fig:personalized-accu-homo-various-models}
\end{figure}

\begin{figure}[h]
     \centering
     \begin{subfigure}[b]{0.32\textwidth}
         \centering
         \includegraphics[width=\textwidth]{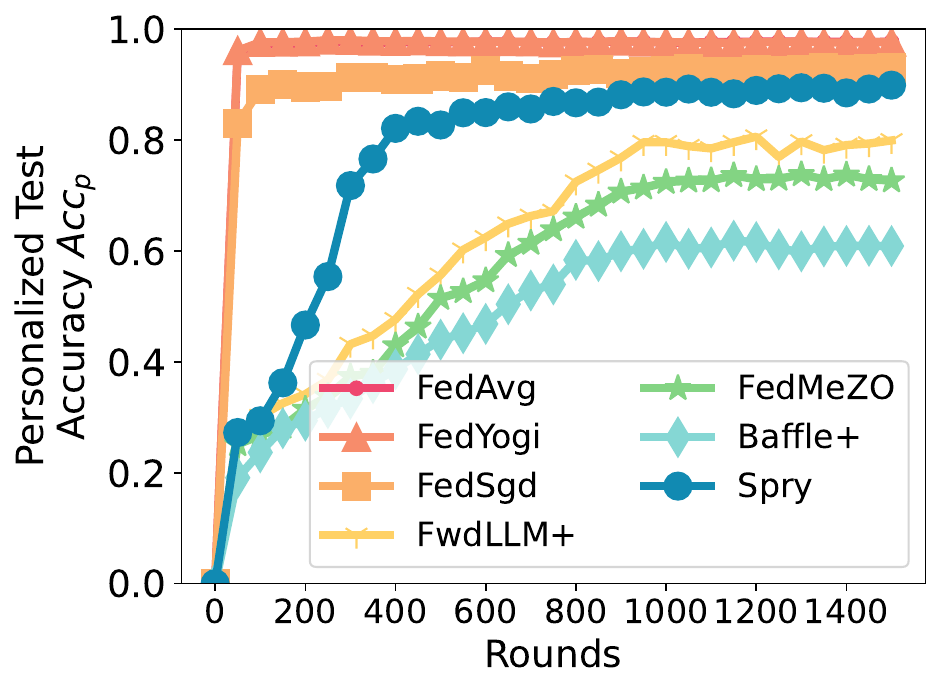}
         \caption{AG News with RoBERTa Large}
         \label{fig:ag-news-roberta-large-pers-hetero}
     \end{subfigure}
     \hfill
     \begin{subfigure}[b]{0.32\textwidth}
         \centering
         \includegraphics[width=\textwidth]{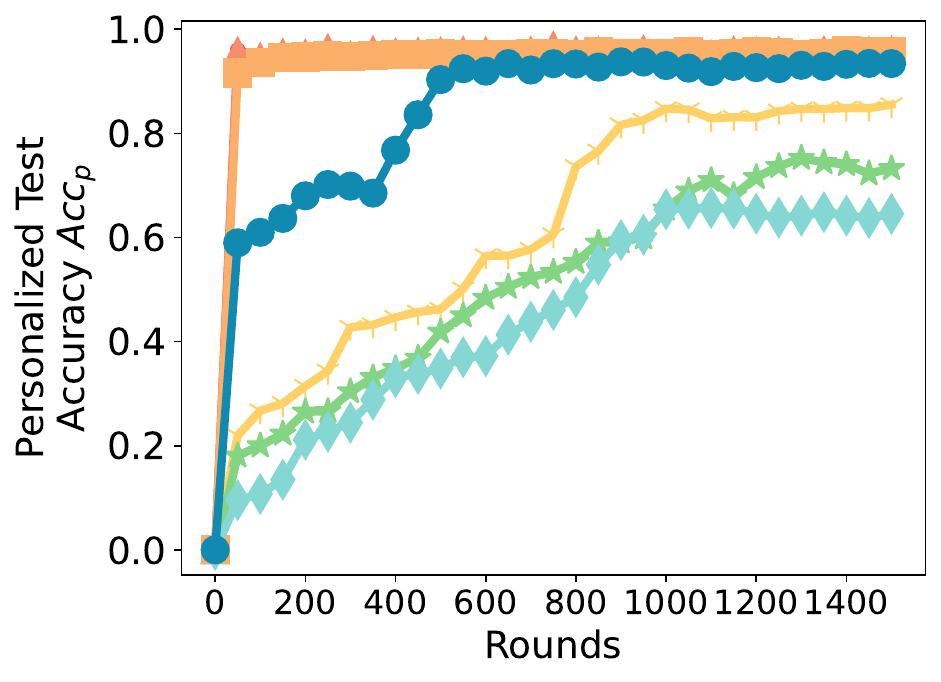}
         \caption{SST2 with RoBERTa Large}
         \label{fig:sst2-roberta-large-pers-hetero}
     \end{subfigure}
     \hfill
     \begin{subfigure}[b]{0.32\textwidth}
         \centering
         \includegraphics[width=\textwidth]{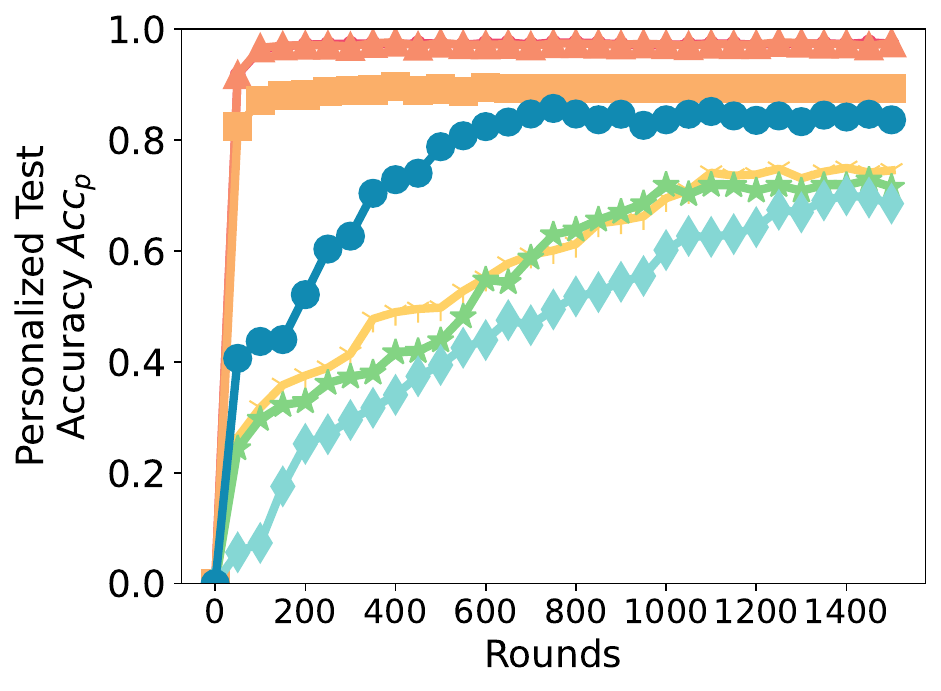}
         \caption{SNLI with RoBERTa Large}
         \label{fig:snli-roberta-large-pers-hetero}
     \end{subfigure}
     \hfill
     \begin{subfigure}[b]{0.32\textwidth}
         \centering
         \includegraphics[width=\textwidth]{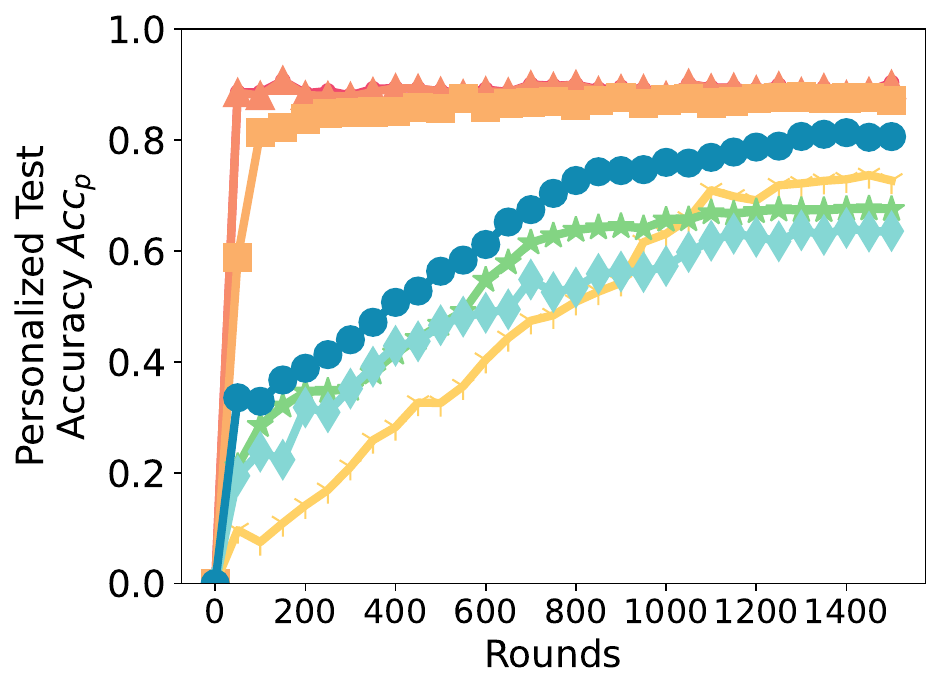}
         \caption{MNLI with RoBERTa Large}
         \label{fig:mnli-roberta-large-pers-hetero}
     \end{subfigure}
     \hfill
     \begin{subfigure}[b]{0.32\textwidth}
         \centering
         \includegraphics[width=\textwidth]{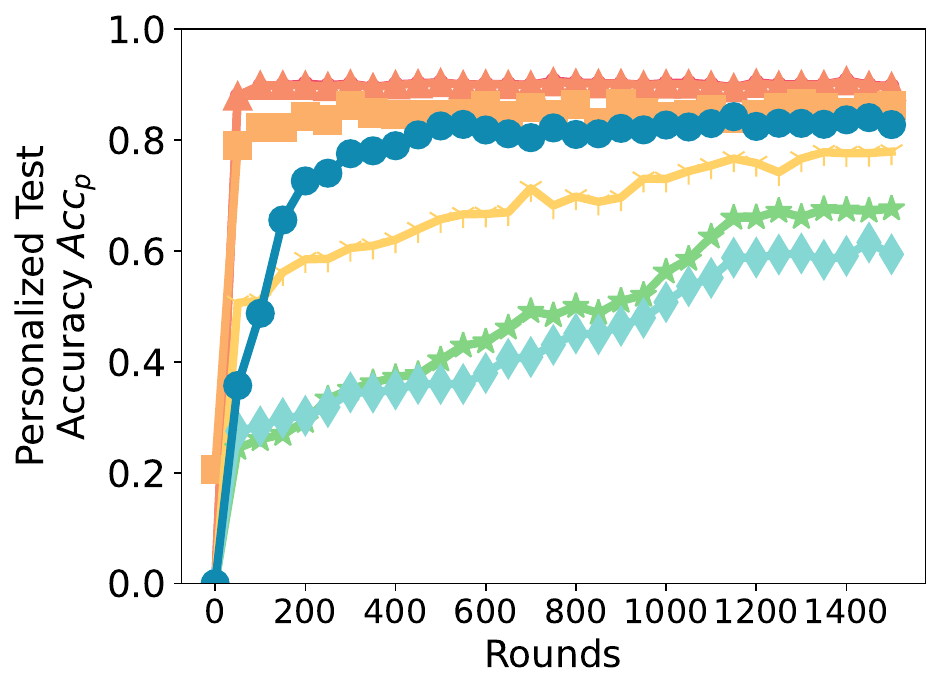}
         \caption{Yahoo with RoBERTa Large}
         \label{fig:yahoo-roberta-large-pers-hetero}
     \end{subfigure}
     \hfill
     \begin{subfigure}[b]{0.32\textwidth}
         \centering
         \includegraphics[width=\textwidth]{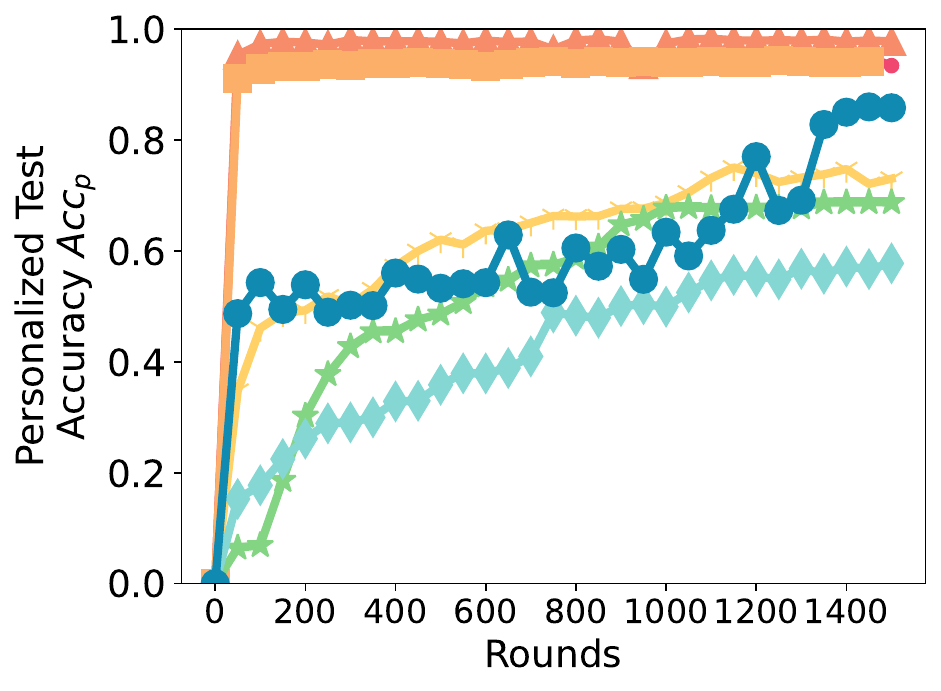}
         \caption{Yelp with RoBERTa Large}
         \label{fig:yelp-roberta-large-pers-hetero}
     \end{subfigure}
     \hfill
     \begin{subfigure}[b]{0.32\textwidth}
         \centering
         \includegraphics[width=\textwidth]{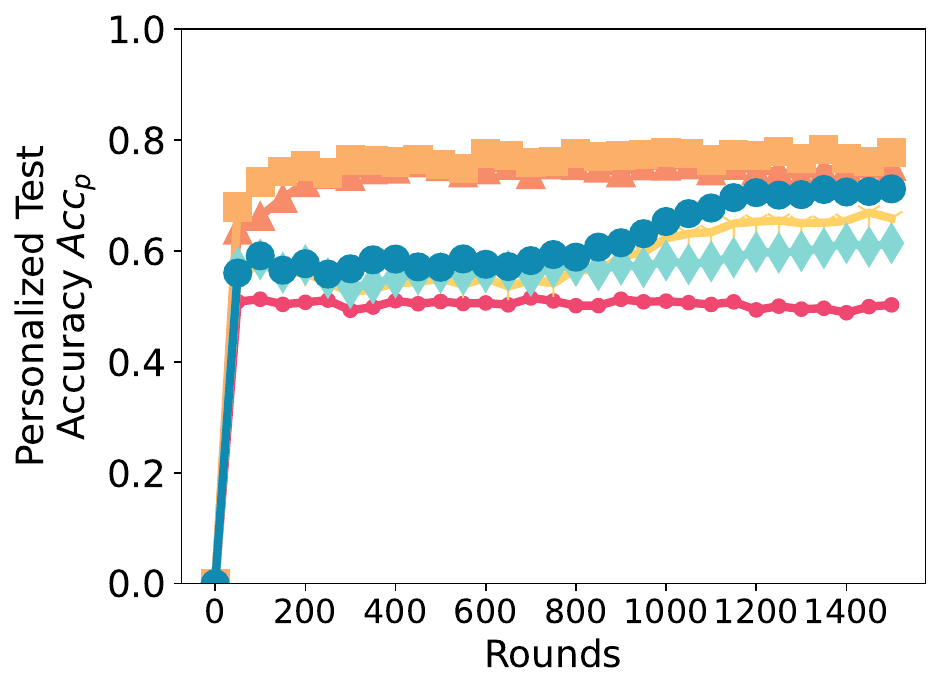}
         \caption{MultiRC with Llama2-7B}
         \label{fig:multirc-llama2-7B-pers-hetero}
     \end{subfigure}
     \hfill
     \begin{subfigure}[b]{0.32\textwidth}
         \centering
         \includegraphics[width=\textwidth]{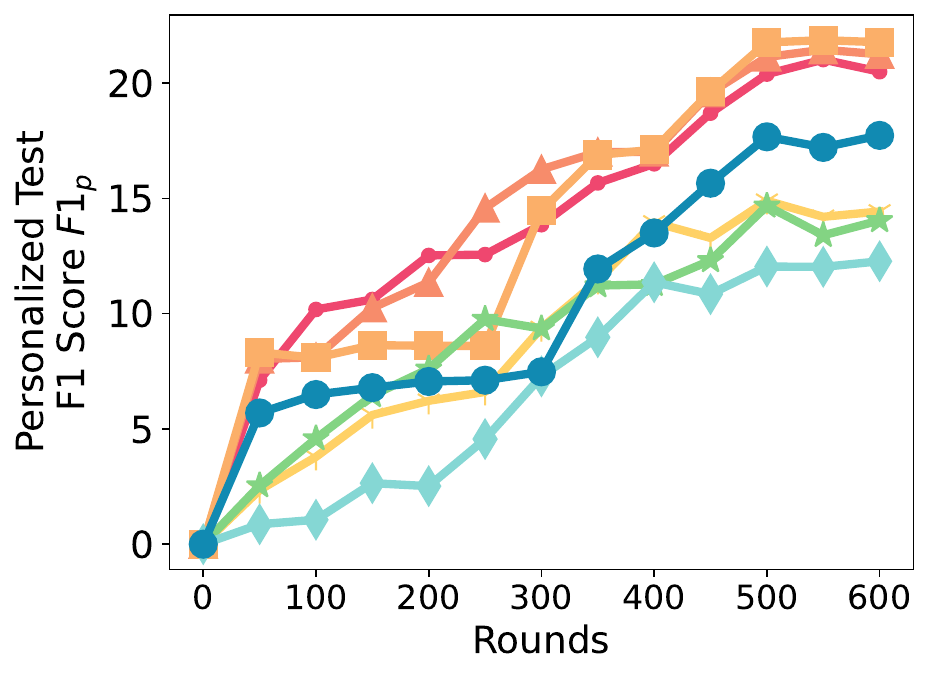}
         \caption{SQuADv2 with OPT6.7B}
         \label{fig:squadv2-opt-6.7B-pers-f1-hetero}
     \end{subfigure}
     \hfill
     \begin{subfigure}[b]{0.32\textwidth}
         \centering
         \includegraphics[width=\textwidth]{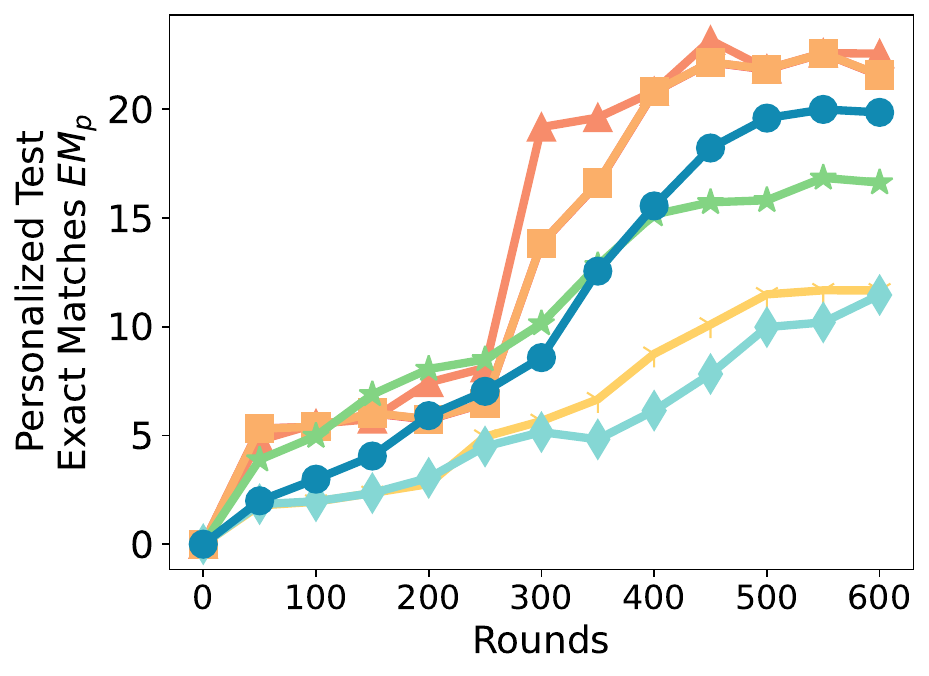}
         \caption{SQuADv2 with OPT6.7B}
         \label{fig:squadv2-opt-6.7B-pers-em-hetero}
     \end{subfigure}
     \hfill
     \begin{subfigure}[b]{0.32\textwidth}
         \centering
         \includegraphics[width=\textwidth]{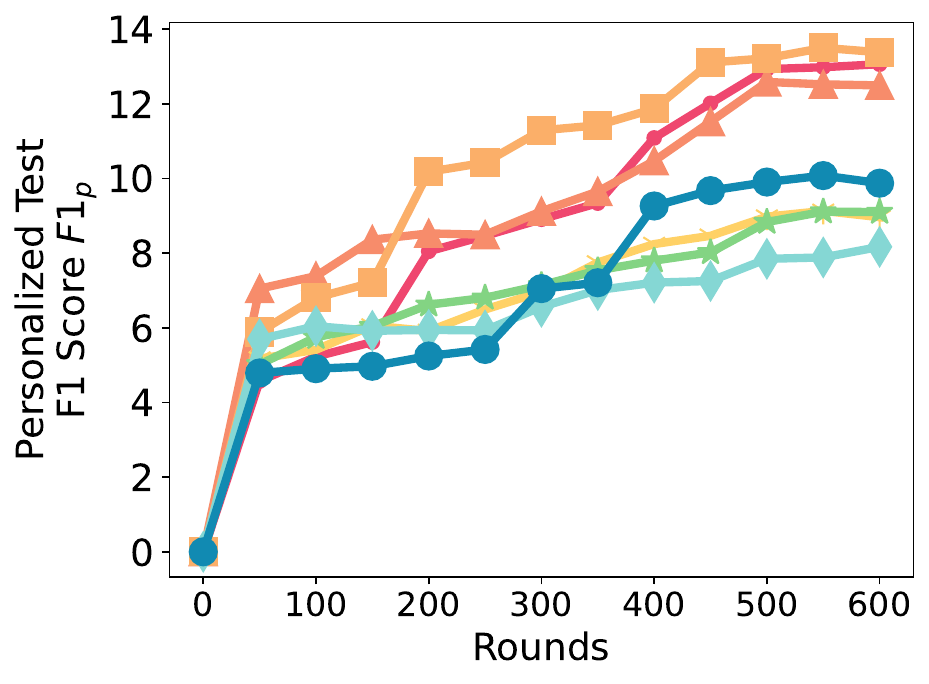}
         \caption{SQuADv2 with OPT13B}
         \label{fig:squadv2-opt-13B-pers-f1-hetero}
     \end{subfigure}
     \hfill
     \begin{subfigure}[b]{0.32\textwidth}
         \centering
         \includegraphics[width=\textwidth]{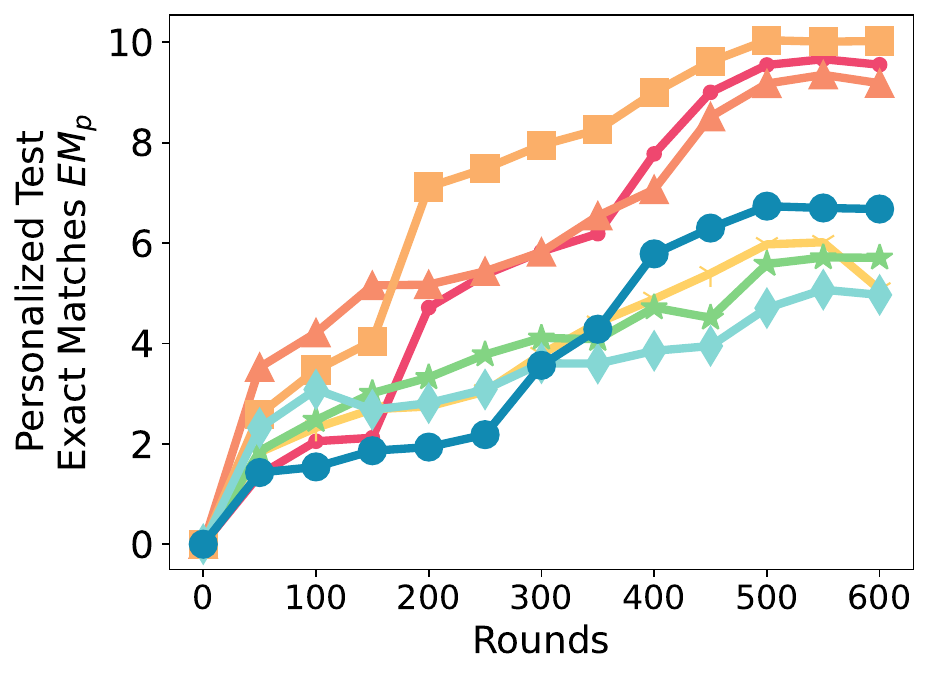}
         \caption{SQuADv2 with OPT13B}
         \label{fig:squadv2-opt-13B-pers-em-hetero}
     \end{subfigure}
        \caption{Personalized accuracy / F1 score / Exact matches for\\ heterogeneous clients (Dirichlet $\alpha=0.1$) setting}
        \label{fig:personalized-accu-hetero}
\end{figure}

Personalized results on homogeneous and heterogeneous clients with Dir $\alpha=1.0$ and $\alpha=0.1$ are shown in 
(a)~Figures~\ref{fig:personalized-accu-homo} and~\ref{fig:personalized-accu-hetero} for RoBERTa Large, Llama2-7B, OPT6.7B, OPT13B; and (b)~Figure~\ref{fig:personalized-accu-homo-various-models} for BERT Large, BERT Base, DistilBert Base, Albert Large v2.

Table~\ref{tbl:acc-pers-adx} shows accuracy and F1 scores of \projectname and its backpropagation and zero-order based counterparts.

\begin{table}
\begin{center}
\captionsetup{justification=centering}
\scriptsize
\tabcolsep=0.06cm
\caption{Personalized accuracy ($Acc_p$) for \projectname and its backpropagation- and zero-order-based counterparts on RoBERTa Large and LLMs. 
SQuADv2 uses F1 score.
$\uparrow$ shows that higher values are better.
The datasets are split with Dir $\alpha=0.1$.
$\diamond$ = Llama2-7B. 
$\star$ = OPT6.7B. 
$\square$ = OPT13B.\\ 
\projectname significantly outperforms the best-performing zero-order-based methods. 
}
\label{tbl:acc-pers-adx}
\begin{tabular}{lcccccccc}
\toprule 
 & \multicolumn{2}{c}{\begin{tabular}[c]{@{}c@{}}Backpropagation-based \\ Methods $\uparrow$ \end{tabular}} & \multicolumn{3}{c}{\begin{tabular}[c]{@{}c@{}}Zero-order-based \\ Methods $\uparrow$\end{tabular}} & \begin{tabular}[c]{@{}c@{}}First-order \\ Forward Mode AD $\uparrow$\end{tabular} & \multicolumn{2}{c}{\begin{tabular}[c]{@{}c@{}}Difference between \\ performances of \projectname  and  \\\end{tabular}} \\ \cmidrule{2-7} 
 & \textsc{FedAvg} & \multicolumn{1}{c}{\textsc{FedYogi}} & \textsc{FwdLLM+} & \textsc{FedMeZO} & \multicolumn{1}{c}{\textsc{Baffle+}} & \projectname & \begin{tabular}[c]{@{}c@{}}best-performing \\ backpropagation \\ method $\uparrow$\end{tabular} & \begin{tabular}[c]{@{}c@{}}best-performing\\ zero-order \\ method $\uparrow$\end{tabular} \\ \midrule 
AG News & 97.76\% & \multicolumn{1}{c}{97.71\%} & 79.94\% & 72.69\% & \multicolumn{1}{c}{60.89\%} & 89.91\% & \phantom{0}-7.85\% & \phantom{0}9.97\% \\
SST2 & 95.84\% & \multicolumn{1}{c}{95.90\%} & 85.51\% & 73.26\% & \multicolumn{1}{c}{64.55\%} & 93.40\% & \phantom{0}-2.50\%  & \phantom{0}7.89\%\\
SNLI & 97.41\% & \multicolumn{1}{c}{97.57\%} & 74.53\% & 71.54\% & \multicolumn{1}{c}{68.55\%} & 83.45\% & -14.12\%  & \phantom{0}8.92\%\\
MNLI & 90.38\% & \multicolumn{1}{c}{90.03\%} & 72.71\% & 67.53\% & \multicolumn{1}{c}{63.58\%} & 80.63\%  & \phantom{0}-9.75\%  & \phantom{0}7.92\% \\
Yahoo & 89.76\% & \multicolumn{1}{c}{89.64\%} & 77.93\% & 67.64\% & \multicolumn{1}{c}{59.40\%} & 82.80\% & \phantom{0}-6.96\% & \phantom{0}4.87\% \\
Yelp & 93.44\% & \multicolumn{1}{c}{97.81\%} & 73.04\% & 68.77\% & \multicolumn{1}{c}{57.78\%} & 85.83\% & -11.98\%  & 12.79\%\\
MultiRC $\diamond$ & 50.28\% & \multicolumn{1}{c}{75.21\%} & 65.91\% & N/A & \multicolumn{1}{c}{61.41\%} & 71.20\% & \phantom{0}-4.01\%  & \phantom{0}5.29\% \\
SQuADv2 $\star$ & 20.49\phantom{\%} & \multicolumn{1}{c}{21.25\phantom{\%}} & 14.43\phantom{\%} & 14.04\phantom{\%} & \multicolumn{1}{c}{12.27\phantom{\%}} & 17.73\phantom{\%} & \phantom{0}-3.52\phantom{\%} & \phantom{0}3.30\phantom{\%}\\
SQuADv2 $\square$ & 13.06\phantom{\%} & \multicolumn{1}{c}{12.49\phantom{\%}} & \phantom{0}8.96\phantom{\%} & \phantom{0}9.10\phantom{\%} & \multicolumn{1}{c}{\phantom{0}8.17\phantom{\%}} & \phantom{0}9.88\phantom{\%} & \phantom{0}-3.18\phantom{\%}& \phantom{0}0.78\phantom{\%} \\
\bottomrule
\end{tabular}
\vspace{-0.65cm}
\end{center}
\end{table}

\clearpage
\subsection{Experiment Variance}
\label{adx:experiment-variance}
Tables~\ref{tbl:experimental-var-gen-pers-medium} and \ref{tbl:experimental-var-gen-pers-large} show the variance of running the same experiments thrice with the random seeds 0, 1, and 2.
\begin{table}[h]
\begin{center}
\captionsetup{justification=centering}
\footnotesize
\tabcolsep=0.06cm
\caption{Experimental variance ($\pm$) for \projectname{} and its counterparts on RoBERTa Large.
}
\label{tbl:experimental-var-gen-pers-medium}
\begin{tabular}{lcccccccccccc}
\toprule
&
\multicolumn{4}{c}{\begin{tabular}[c]{@{}c@{}}Backpropgation-based \\ Methods\end{tabular}} & \multicolumn{6}{c}{\begin{tabular}[c]{@{}c@{}}Zero-order-based \\ Methods\end{tabular}} & \multicolumn{2}{c}{\begin{tabular}[c]{@{}c@{}}First-order \\ Forward-mode AD\end{tabular}} \\ 
\cmidrule{2-13}
 & \multicolumn{2}{c}{\textsc{FedAvg}} & \multicolumn{2}{c}{\textsc{FedYogi}} & \multicolumn{2}{c}{\textsc{FwdLLM+}} & \multicolumn{2}{c}{\textsc{FedMeZO}} & \multicolumn{2}{c}{\textsc{Baffle+}} & \multicolumn{2}{c}{\projectname} \\
\cmidrule{2-13}
 & $Acc_g$ & \multicolumn{1}{l}{$Acc_p$} & $Acc_g$ & \multicolumn{1}{l}{$Acc_p$} & $Acc_g$ & \multicolumn{1}{l}{$Acc_p$} & $Acc_g$ & \multicolumn{1}{l}{$Acc_p$} & $Acc_g$ & \multicolumn{1}{l}{$Acc_p$} & $Acc_g$ & $Acc_p$ \\ 
 \midrule 
AG News & 0.51\% & \multicolumn{1}{l}{0.44\%} & 0.26\% & 0.21\% & 0.73\% & \multicolumn{1}{l}{0.71\%} & 1.34\% & \multicolumn{1}{l}{1.27\%} & 0.68\% & 0.51\% & 1.16\% & 1.08\% \\
SST2 & 0.37\% & \multicolumn{1}{l}{0.47\%} & 0.43\% & 0.37\% & 1.26\% & \multicolumn{1}{l}{1.14\%} & 1.13\% & \multicolumn{1}{l}{1.06\%} & 0.41\% & 0.45\% & 0.77\% & 0.95\% \\
SNLI & 0.45\% & \multicolumn{1}{l}{0.39\%} & 0.17\% & 0.11\% & 0.82\% & \multicolumn{1}{l}{0.67\%} & 0.74\% & \multicolumn{1}{l}{0.65\%} & 0.84\% & 0.78\% & 0.51\% & 0.45\% \\
MNLI & 0.63\% & \multicolumn{1}{l}{0.58\%} & 0.29\% & 0.24\% & 1.39\% & \multicolumn{1}{l}{1.25\%} & 1.98\% & \multicolumn{1}{l}{1.85\%} & 1.15\% & 1.01\% & 1.45\% & 1.32\% \\
Yahoo & 0.24\% & \multicolumn{1}{l}{0.33\%} & 0.53\% & 0.47\% & 0.47\% & \multicolumn{1}{l}{0.44\%} & 1.06\% & \multicolumn{1}{l}{0.93\%} & 0.59\% & 0.48\% & 0.99\% & 0.85\% \\
Yelp & 0.22\% & \multicolumn{1}{l}{0.25\%} & 0.36\% & 0.27\% & 0.54\% & \multicolumn{1}{l}{0.49\%} & 0.82\% & \multicolumn{1}{l}{0.66\%} & 0.83\% & 0.71\% & 0.76\% & 0.61\% \\
\bottomrule
\end{tabular}
\end{center}
\end{table}

\begin{table}[h]
\begin{center}
\captionsetup{justification=centering}
\footnotesize
\tabcolsep=0.06cm
\caption{Experimental variance ($\pm$) for generalized ($Acc_g$ for MultiRC / $F1_g$ for SQuADv2) and personalized ($Acc_p$ for MultiRC / $F1_p$ for SQuADv2) accuracy or F1 score for \projectname{} and its counterparts. 
$\diamond$ = Llama2 7B. 
$\star$ = OPT 6.7B. 
$\square$ = OPT 13B.
}
\label{tbl:experimental-var-gen-pers-large}
\begin{tabular}{lcccccccccccc}
\toprule
&
\multicolumn{4}{c}{\begin{tabular}[c]{@{}c@{}}Backpropgation-based \\ Methods\end{tabular}} & \multicolumn{6}{c}{\begin{tabular}[c]{@{}c@{}}Zero-order-based \\ Methods\end{tabular}} & \multicolumn{2}{c}{\begin{tabular}[c]{@{}c@{}}First-order \\ Forward-mode AD\end{tabular}} \\ 
\cmidrule{2-13}
 & \multicolumn{2}{c}{\textsc{FedAvg}} & \multicolumn{2}{c}{\textsc{FedYogi}} & \multicolumn{2}{c}{\textsc{FwdLLM+}} & \multicolumn{2}{c}{\textsc{FedMeZO}} & \multicolumn{2}{c}{\textsc{Baffle+}} & \multicolumn{2}{c}{\projectname{}} \\
\cmidrule{2-13}
 & $Acc_g$ & \multicolumn{1}{l}{$Acc_p$} & $Acc_g$ & \multicolumn{1}{l}{$Acc_p$} & $Acc_g$ & \multicolumn{1}{l}{$Acc_p$} & $Acc_g$ & \multicolumn{1}{l}{$Acc_p$} & $Acc_g$ & \multicolumn{1}{l}{$Acc_p$} & $Acc_g$ & $Acc_p$ \\ 
 \midrule
MultiRC $\diamond$ & 0.58\% & \multicolumn{1}{l}{0.43\%} & 0.34\% & 0.29\% & 0.89\% & \multicolumn{1}{l}{0.74\%} & N/A & \multicolumn{1}{l}{N/A} & 1.34\% & 1.02\% & 0.97\% & 0.81\% \\
SQuADv2 $\star$ & 1.73\phantom{\%} & \multicolumn{1}{l}{1.07} & 1.42\phantom{\%} & 0.84\phantom{\%} & 2.17\phantom{\%} & \multicolumn{1}{l}{1.87} & 1.78 & \multicolumn{1}{l}{1.51} & 2.78\phantom{\%} & 2.01\phantom{\%} & 1.76\phantom{\%} & 0.99\phantom{\%} \\
SQuADv2 $\square$ & 0.86\phantom{\%} & \multicolumn{1}{l}{0.43} & 0.61\phantom{\%} & 0.45\phantom{\%} & 1.16\phantom{\%} & \multicolumn{1}{l}{0.97} & 1.03 & \multicolumn{1}{l}{0.87} & 1.34\phantom{\%} & 1.14\phantom{\%} & 0.97\phantom{\%} & 0.81\phantom{\%} \\
\bottomrule 
\end{tabular}
\end{center}
\end{table}

\clearpage
\section{Proofs}
\label{adx:proofs}
\subsection{Basics}
\paragraph{Server Update.}
The server update of \textsc{Spry} uses adaptive optimizer \textsc{FedYogi}.
However, to simplify the proofs without losing generality, we use the server update of \textsc{FedAdam}~\cite{reddi2021adaptive}.
\textsc{FedAdam} has the exact update rule as \textsc{FedYogi} but without a \texttt{sign} function in its calculation of the second moment of the gradients (See Algorithm 2 of AFO~\cite{reddi2021adaptive}).

Hence, the server update of \textsc{Spry} is 
\begin{align}
    \wround{r} \gets \wround{r-1} + \eta \frac{\deltar{r}}{\sqrt{\vr{r}} + \tau} \qquad \forall \text{ trainable weights } \wround{r} \in [d]. \label{eq:server-update}
\end{align}
$\deltar{r}$ is the square of accumulated gradients from all clients. 
$\vr{r}$ is the second moment of $\deltar{r}$.
$\tau$ is a small positive real number, to prevent division by zero errors. 
Note that we are assuming flattened weights $w \in \bR^d$ without the loss of generality.

With \projectname, the aim is to solve the following optimization problem:
\begin{align}
    \min_{w \in \bR^d} f(w) = \frac{1}{m} \sum_{m=1}^M F_m(\partialw), \label{eq:global-objective}
\end{align}
where $\mrange$, $F_m(\partialw) = \bE_{(x, y) \sim \cD_m} [f_m (\partialw, (x, y))] $ is the objective function, $\cD_m$ is the dataset, and $f_m$ is the loss function of a client $m \in [M]$.

\paragraph{Accumulated Gradients.}
With \projectname, each client trains a subset of weights $\partialw$.
In a low participation rate setting, each $\partialw$ is only trained by one of the participating clients from the set of available clients $\cM$. 
Although we make our analysis more generally applicable by showing multiple clients training the same $\partialw$.

The true global gradients can be written as,
\begin{align}
    \nabla f(w) = \left[ \frac{1}{\subM} \sum_{m \in \subcM} \nabla F(\partialw) \; \Big| \; \mrange , \subcM \subset \cM \right] \label{eq:global-gradient}
\end{align}
where $\subcM$ is a set of clients training the subset of the weights $\partialw$. $\subM$ is the size of $\subcM$.

\paragraph{Client Update.}
The directional derivative of Forward-mode AD is denoted as $\nabla \hat{f}_v(w; (x,y))$, where $v \in \bR^d$ is the random perturbation of weights $w$ and $(x,y)$ are sampled from a dataset $\cD$.
For each client $m$, through Forward-mode AD, we have $\nabla \hat{f}_m(\partialw, \partialv; (x,y)) = (\nabla \hat{f}_{m, v}(\partialw; (x,y))\cdot \partialv)$ to estimate the true gradient $\nabla F_m(\partialw)$:
\begin{align}
    \bE_{\partialv, \cD_m} \left[\nabla \hat{f}_m(\partialw, \partialv; \cD_m)\right] &= \frac{1}{DK} \sum_{(x,y)\sim \cD_m} \sum_{i=1}^K \bE_{v_{i,{\overline{m}}}, (x,y)} \left[ \nabla f_m(\partialw; (x,y)) v_{i, {\overline{m}}} v_{i,{\overline{m}}}^T \right] \label{eq:client-forward-gradient-estimation}\\
    \bE_{\partialv, \cD_m} \left[\nabla \hat{f}_m(\partialw, \partialv; \cD_m)\right] &= \nabla F_m(\partialw) \qquad\qquad\qquad\qquad \because\text{Theorem 1 of \textsc{Fgd}~\cite{baydin2022gradients}}
    \label{eq:client-expected-gradient}
\end{align}
Here, the expectation is under the randomness of sampled data $\cD$, and random perturbation $v$.
$K$ is the number of perturbations per batch. 
\projectname uses $K=1$ by default, but here we aim to make our analysis more general to see the impact of $K$ on various properties of \projectname.

\subsection{Assumptions}
\label{subsec:adx-assumptions}
We will be using the following assumptions to derive the first and second moment of the forward gradient and to calculate the rate of convergence of \projectname,
\begin{assumption}[Smoothness]
\label{asm:smoothness}
The gradient of function $F_m$ is $L$-Lipschitz,
    $$||\nabla F_m(w_1) - \nabla F_m(w_2)|| \leq L ||w_1 - w_2||; \; \forall m\in [M] \text{ and } w_1, w_2 \in \bR^d.$$
\end{assumption}

\begin{assumption} [Bounded Global Variance (Assumption 2 in AFO~\cite{reddi2021adaptive})]
\label{asm:bounded-global-variance}
    The variance of the global objective function $f$ is bounded by $\sigma_g$ as $$\frac{1}{M} \sum_{m=1}^M || \left[ \nabla F_m(\partialw)\right]_j - \left[ \nabla f(\partialw) \right]_j ||^2 \leq \sigma^2_{g, j}; \; \forall m\in [M], w \in \bR^d \text{ and } \forall j \in [d],$$
    where $\nabla F_m$ is the true gradient of client $m$. As defined earlier, $\mrange$.
\end{assumption}


\begin{assumption} [Bounded Gradients (Assumption 3 in AFO~\cite{reddi2021adaptive})]
\label{asm:bounded-gradients}
    The function $f_m(w; (x,y))$ has $G$-bounded gradients such that for any client $m\in [M]$, weights $w \in \bR^d$, and sampled data $(x,y) \sim \cD_m$; we have $$|\left[\nabla f_m(w; (x,y))\right]_j| < G, \; \forall j \in [d]$$
\end{assumption}

\subsection{First and Second Moments of Forward Gradients}
We first prove that in \projectname, estimation of $\nabla f$ by the accumulated forward mode gradients $\nabla \hat{f}$ across all clients of an arbitrary round $r$, depends on the heterogeneity across client datasets.
Statements on homogeneous data have been proven for \textsc{Fgd}~\cite{baydin2022gradients} and \textsc{MeZO}~\cite{malladi2023mezo} in single-client or centralized settings. 
Here we focus on the specific federated setting of \projectname, where gradients are accumulated differently than in traditional \textsc{FedAvg}~\cite{macmahan2017aFedLearning}. 
In \projectname, we have 
\begin{align}
    \nabla f (w) = \bE_v \left[ \nabla \hat{f} (w, v) \right] = \left[ \frac{1}{\subM} \sum_{m \in \subcM} \bE \left[ \nabla \hat{f}_m (\partialw,\partialv; \cD_m) \right] \Big| \; \forall \subcM \subset \cM; \mrange \right], \label{eq:true-global-gradient}
\end{align}
where $w$ represents weights, $v$ are their corresponding perturbations, and $\cD_m$ is the dataset of client $m$. 

We omit the round index of the model weights $w$ and their perturbations $v$ for this section since the same relationship will hold for any arbitrary round $r$.
    
\begin{theorem} [Estimation of the Global Gradient]
\label{thm:estimation-global-gradient}
    In \projectname, global forward gradient $\nabla \hat{f}$ of the trainable weights $w \in \bR^d$, with the corresponding weight perturbations $v\in \bR^d$, computed by $M$ participating clients is estimated in terms of true global gradient $\nabla f$ as,
    \begin{align}
    \bE_{v, \cD} &[ \nabla \hat{f}(w, v; \cD)] = \nabla f(w) \nonumber \\ &\qquad + \frac{1}{\subM}
        \begin{bmatrix}
            \sum_{m \in \subcM_1} \sum_{c=1}^{C} \alpha_{m,c} \bE_{(x,y_c) \in \cD} \left[ \nabla \hat{f}_m (w_{\left[1, \frac{d}{M} \right]}, v_{\left[1, \frac{d}{M} \right]}; (x,y_c)) \right]\\
            \sum_{m \in \subcM_2} \sum_{c=1}^{C} \alpha_{m,c} \bE_{(x,y_c) \in \cD} \left[ \nabla \hat{f}_m (w_{\left[\frac{d}{M}+1,\frac{2d}{M}\right]}, v_{\left[\frac{d}{M}+1,\frac{2d}{M}\right]}; (x,y_c)) \right]\\
        \vdots 
        \end{bmatrix}^T \nonumber
    \end{align}
    where $C$ is total number of classes and $\alpha_{m,c} = \left(\frac{n_c}{|\cD|} - \frac{n_{m, c} \alpha_c}{|\cD_m|} \right)$. For a class $c$, $n_c$ is its sample count, $\alpha_c$ is its Dirichlet concentration parameter. For a client $m$; $n_{m,c}$ is the sample count of the $c^{th}$ class, and $\cD_m$ is the size of the data of client $m$. The global data is $\cD = \sum_{m \in \cM} \cD_m$. $\subcM$ is the set of clients training an arbitrary subset of weights, $\subM = |\subcM_i|; \; \forall i \in [M/d]$.
\end{theorem}
\begin{proof}
Suppose the global dataset $\cD$ is defined as a combination of all $\cD_m$. 
Hence, $\cD = \cup_{m \in [M]} \cD_m$.

In \projectname, the global forward gradient  is defined as 
\begin{align}
    \bE_{v, \cD} [ \nabla \hat{f}(w, v; \cD)] &= \bE 
    \begin{bmatrix}
        \frac{1}{\subM} \sum_{m \in \subcM_1} \nabla \hat{f}_{m} (w_{\left[1,\frac{d}{M}\right]}, v_{\left[1,\frac{d}{M}\right]}; \cD) \\
        \frac{1}{\subM} \sum_{m \in \subcM_2} \nabla \hat{f}_{m} (w_{\left[\frac{d}{M}+1,\frac{2d}{M}\right]}, v_{\left[\frac{d}{M}+1,\frac{2d}{M}\right]}; \cD) \\
        \vdots \\
        \frac{1}{\subM} \sum_{m \in \subcM_{M/d}} \nabla \hat{f}_{m} (w_{\left[\frac{(M-1)d}{M}+1,d\right]}, v_{\left[\frac{(M-1)d}{M}+1,d \right]}; \cD) 
    \end{bmatrix}^T \label{eq:estimation-of-global-forward-gradient}
\end{align}
To combine the gradient estimates from the above equation for all clients $m\in \cM$, we first consider the \emph{bias} $b_m$ between the gradient estimate under \textbf{(a)} Globally combined data $\cD$ and \textbf{(b)} Data $\cD_m$ of an arbitrary client $m$. 

\textbf{Measuring the dataset bias.}
Note that we consider samples $(x,y)$ of $\cD_m$ to be sampled from $\cD$, since $\cD_m \subset \cD$.
\begin{align}
    \bE_{\partialv, \cD} \left[ \nabla \hat{f}_m (\partialw, \partialv; \cD) \right] &= \bE_{\partialv, \cD_m} \left[ \nabla \hat{f}_m (\partialw, \partialv; \cD_m) \right] + b_m \label{eq:biased-estimation}
\end{align}
Computing the \emph{bias} term $b$ requires information on dataset distribution.
In FL settings, Dirichlet distribution to spread the data into heterogeneous splits is a popular way to simulate heterogeneity~\cite{panchal2023flow}.
The biased dataset $\cD_m$ is sampled from the combined dataset $\cD$ using the Dirichlet distribution. 
This allows us to utilize the properties of the distribution to derive the relationship between forward gradients on global data and on individual local data:

For classification tasks, let's say $\cD$ has $C$ total classes: $y_1, \dots, y_C$.
We have $\alpha_1, \dots, \alpha_C$ as concentration parameters of the Dirichlet distribution.
The expected forward gradient for the combined dataset $\cD$ can be expressed as a weighted sum of the expected forward gradients for each class $C$:
\begin{align}
    \bE_{\partialv, \cD} \left[ \nabla \hat{f}_m (\partialw, \partialv; \cD)  \right] &= \sum_{c=1}^{C} \frac{n_c}{|\cD|} \bE_{\partialv, (x,y_c)\in\cD} \left[ \nabla \hat{f}_m (\partialw, \partialv; (x,y_c)) \right], \label{eq:forward-gradient-estimation-on-combined-data}
\end{align}
where $n_c$ is the sample count of class $c$.
And for the biased dataset $\cD_m$ sampled with Dirichlet concentration parameters $\alpha_1, \dots, \alpha_C$, the expected forward gradient can be expressed as,
\begin{align}
    \bE_{\partialv, \cD_m} \left[ \nabla \hat{f}_m (\partialw, \partialv; \cD_m) \right] &= \sum_{c=1}^{C} \frac{n_{m, c} \alpha_c}{|\cD_m|} \bE_{\partialv, (x,y_c)\in \cD} \left[ \nabla \hat{f}_m (\partialw, \partialv; (x,y_c)) \right], \label{eq:forward-gradient-estimation-on-client-data}
\end{align}
where $n_{m,c}$ is the sample count of class $c$ for client $m$.

Subtracting Equation~\ref{eq:forward-gradient-estimation-on-client-data} from Equation~\ref{eq:forward-gradient-estimation-on-combined-data},
\begin{align}
    b_m &= \bE_{\partialv, \cD} \left[ \nabla \hat{f}_m (\partialw, \partialv; \cD) \right] - \bE_{\partialv, \cD_m} \left[ \nabla \hat{f}_m (\partialw, \partialv; \cD_m) \right] \label{eq:bias-v1}\\
    &= \sum_{c=1}^{C} \left(\frac{n_c}{|\cD|} - \frac{n_{m, c} \alpha_c}{|\cD_m|} \right) \bE_{\partialv, (x,y_c) \in \cD} \left[ \nabla \hat{f}_m (\partialw, \partialv; (x,y_c)) \right] \label{eq:bias-v2}
\end{align}
For any $\mrange$, we have $\bE_{\partialv, \cD_m} \left[ \nabla \hat{f}_{m} (\partialw, \partialv; \cD_m) \right] = F_m(w)$ from Lemma 1 in FGD~\cite{baydin2022gradients}.

Plugging in the above result and Equation~\ref{eq:bias-v2} in Equation~\ref{eq:biased-estimation}, 
\begin{align}
    \bE_{\partialv, \cD} &\left[ \nabla \hat{f}_m (\partialw, \partialv; \cD) \right] = \bE_{\partialv, \cD_m} \left[ \nabla \hat{f}_m (\partialw, \partialv; \cD_m) \right] + b_m \nonumber\\
    &= \nabla F_m(\partialw) + \sum_{c=1}^{C} \left(\frac{n_c}{|\cD|} - \frac{n_{m, c} \alpha_c}{|\cD_m|} \right) \bE_{\partialv, (x,y_c) \in \cD} \left[ \nabla \hat{f}_m (\partialw, \partialv; (x,y_c)) \right] 
\end{align}
For $i^{th}$ row in Equation~\ref{eq:estimation-of-global-forward-gradient},
\begin{align}
    \bE_{\partialv, \cD} &\left[ \nabla \hat{f}_i (\partialw, \partialv; \cD) \right] = \frac{1}{\subM} \sum_{m \in \subcM_1} \bE_{\partialv, \cD} \left[ \nabla \hat{f}_{m} (\partialw, \partialv; \cD) \right] \\
    &= \frac{1}{\subM} \sum_{m \in \subcM_1} \left( \nabla F_m(\partialw) + \sum_{c=1}^{C} \left(\frac{n_c}{|\cD|} - \frac{n_{m, c} \alpha_c}{|\cD_m|} \right) \bE_{\partialv, (x,y_c) \in \cD} \left[ \nabla \hat{f}_m (\partialw, \partialv; (x,y_c)) \right]  \right) \\
    &= \nabla f (\partialw) + \frac{1}{\subM} \sum_{m \in \subcM_1} \sum_{c=1}^{C} \left(\frac{n_c}{|\cD|} - \frac{n_{m, c} \alpha_c}{|\cD_m|} \right) \bE_{\partialv, (x,y_c) \in \cD} \left[ \nabla \hat{f}_m (\partialw, \partialv; (x,y_c)) \right] \label{eq:biased-global-forward-gradient}
\end{align}

Putting Equation~\ref{eq:biased-global-forward-gradient} in Equation~\ref{eq:estimation-of-global-forward-gradient},
\begin{align}
    \bE_{v, \cD} &[ \nabla \hat{f}(w, v; \cD)] \\
    &=  
    \begin{bmatrix}
        \nabla f (w_{\left[1, \frac{d}{M} \right]}) + \frac{1}{\subM} \sum_{m \in \subcM_1} \sum_{c=1}^{C} \left(\frac{n_c}{|\cD|} - \frac{n_{m, c} \alpha_c}{|\cD_m|} \right) \bE_{(x,y_c) \in \cD} \left[ \nabla \hat{f}_m (w_{\left[1, \frac{d}{M} \right]}, v_{\left[1, \frac{d}{M} \right]}; (x,y_c)) \right]\\
        \nabla f (w_{\left[\frac{d}{M}+1,\frac{2d}{M}\right]}) + \frac{1}{\subM} \sum_{m \in \subcM_2} \sum_{c=1}^{C} \left(\frac{n_c}{|\cD|} - \frac{n_{m, c} \alpha_c}{|\cD_m|} \right) \bE_{(x,y_c) \in \cD} \left[ \nabla \hat{f}_m (w_{\left[\frac{d}{M}+1,\frac{2d}{M}\right]}, v_{\left[\frac{d}{M}+1,\frac{2d}{M}\right]}; (x,y_c)) \right]\\
        \vdots 
    \end{bmatrix}^T \\
    &= \begin{bmatrix}
        \nabla f (w_{\left[1, \frac{d}{M} \right]}) \\
        \nabla f (w_{\left[\frac{d}{M}+1,\frac{2d}{M}\right]})\\
        \vdots
    \end{bmatrix}^T + \frac{1}{\subM}
    \begin{bmatrix}
        \sum_{m \in \subcM_1} \sum_{c=1}^{C} \left(\frac{n_c}{|\cD|} - \frac{n_{m, c} \alpha_c}{|\cD_m|} \right) \bE_{(x,y_c) \in \cD} \left[ \nabla \hat{f}_m (w_{\left[1, \frac{d}{M} \right]}, v_{\left[1, \frac{d}{M} \right]}; (x,y_c)) \right]\\
        \sum_{m \in \subcM_2} \sum_{c=1}^{C} \left(\frac{n_c}{|\cD|} - \frac{n_{m, c} \alpha_c}{|\cD_m|} \right) \bE_{(x,y_c) \in \cD} \left[ \nabla \hat{f}_m (w_{\left[\frac{d}{M}+1,\frac{2d}{M}\right]}, v_{\left[\frac{d}{M}+1,\frac{2d}{M}\right]}; (x,y_c)) \right]\\
        \vdots 
    \end{bmatrix}^T \\
    &= \nabla f(w) + \frac{1}{\subM}
    \begin{bmatrix}
        \sum_{m \in \subcM_1} \sum_{c=1}^{C} \left(\frac{n_c}{|\cD|} - \frac{n_{m, c} \alpha_c}{|\cD_m|} \right) \bE_{(x,y_c) \in \cD} \left[ \nabla \hat{f}_m (w_{\left[1, \frac{d}{M} \right]}, v_{\left[1, \frac{d}{M} \right]}; (x,y_c)) \right]\\
        \sum_{m \in \subcM_2} \sum_{c=1}^{C} \left(\frac{n_c}{|\cD|} - \frac{n_{m, c} \alpha_c}{|\cD_m|} \right) \bE_{(x,y_c) \in \cD} \left[ \nabla \hat{f}_m (w_{\left[\frac{d}{M}+1,\frac{2d}{M}\right]}, v_{\left[\frac{d}{M}+1,\frac{2d}{M}\right]}; (x,y_c)) \right]\\
        \vdots 
    \end{bmatrix}^T
\end{align}

\end{proof}  

Next, we formulate the norm of the forward gradient $\nabla \hat{f}$.
\begin{lemma} [Norm of the Forward Gradient]
\label{lma:forward-gradient-second-moment}
    Under Assumption~\ref{asm:bounded-global-variance}, and at the participation rate of $s$, $M$ participating clients training weights $w \in \bR^d$ through random perturbations $v \in \bR^d$ in \projectname derives the accumulated forward gradient $\nabla \hat{f}(w, v; \cD)$ such that,
    \begin{align}
        \bE_{v, \cD} || \nabla \hat{f} (w, v; \cD) ||^2 &=  \bE || \nabla f(w) ||^2 \left( 1 + \left(\frac{2(3d + K - 1)}{\subM K}\right) \sum_{m \in \cM} \sum_{c \in [C]} \left(\frac{n_c}{|\cD|} - \frac{n_{m, c} \alpha_c}{|\cD_m|} \right)^2 \right) 
        \label{eq:forward-gradient-norm} \nonumber \\
        &\qquad + \left(\frac{2 \sigma_g^2 (1 - s) (3d + K - 1)}{\subM K}\right)  \sum_{m \in \cM}  \sum_{c \in [C]} \left(\frac{n_c}{|\cD|} - \frac{n_{m, c} \alpha_c}{|\cD_m|} \right)^2,
    \end{align}
    where $\cD$ is the combination of datasets of all $M$ clients, $K$ is the number of perturbations per batch, $\subM$ is the count of clients training a particular weight subset, $\sigma_g^2$ is the upper bound of the global gradient variance. 
    For a total of $C$ classes in $\cD$, we define $n_c$ as the sample count of the $c^{th}$ class, $\alpha_c$ is Dirichlet concentration parameter for the $c^{th}$ class, $\cD$ is the size of the global data. 
    For a client $m$; $n_{m,c}$ is the sample count of the $c^{th}$ class for client $m$, and $\cD_m$ is the size of the data of client $m$.
\end{lemma}
\begin{proof}
    The proof follows a similar style of Lemma 2 in \textsc{MeZO}~\cite{malladi2023mezo}. 
    The difference in our setting is that we have $ \nabla \hat{f}(w, v; \cD)$ which is an aggregate of $ \nabla \hat{f}_m(w, v; \cD)$ derived from the federated clients, while \textsc{MeZO} has results under the setting of a single client.
    
    
    From Theorem~\ref{thm:estimation-global-gradient} we have, 
    \begin{align}
        \bE_{v, \cD} &\left[ \nabla \hat{f}(w, v; \cD) \right] \nonumber \\ 
        &= \begin{bmatrix}
        \nabla f (w_{\left[1, \frac{d}{M} \right]}) + \frac{1}{\subM} \sum_{m \in \subcM_1} \sum_{c=1}^{C} \left(\frac{n_c}{|\cD|} - \frac{n_{m, c} \alpha_c}{|\cD_m|} \right) \bE_{(x,y_c) \in \cD} \left[ \nabla \hat{f}_m (w_{\left[1, \frac{d}{M} \right]}, v_{\left[1, \frac{d}{M} \right]}; (x,y_c)) \right]\\
        \nabla f (w_{\left[\frac{d}{M}+1,\frac{2d}{M}\right]}) + \frac{1}{\subM} \sum_{m \in \subcM_2} \sum_{c=1}^{C} \left(\frac{n_c}{|\cD|} - \frac{n_{m, c} \alpha_c}{|\cD_m|} \right) \bE_{(x,y_c) \in \cD} \left[ \nabla \hat{f}_m (w_{\left[\frac{d}{M}+1,\frac{2d}{M}\right]}, v_{\left[\frac{d}{M}+1,\frac{2d}{M}\right]}; (x,y_c)) \right]\\
        \vdots 
    \end{bmatrix}^T \label{eq:repeat-expected-global-forward-gradient}
    \end{align}
    where $$\bE_{\partialv, (x, y_c) \sim \cD } \left[ \nabla \hat{f}_m(\partialw, \partialv; (x, y_c)) \right] =\frac{1}{|\cD|K}\sum_{(x,y_c)\sim \cD}\sum_{i \in [K]} \bE \left[ \left( \nabla f_m\left(\partialw; (x,y_c) \right) \cdot v_{i, {\overline{m}}}\right){v}_{i, {\overline{m}}}^T \right],$$ using Equation~\ref{eq:client-forward-gradient-estimation}.
    The right hand side expectation is also under the randomness of partial perturbations. 
    Here, $K$ is perturbation count, and $|\cD|$ is the size of the combined dataset $\cD = \cup_{m \in [M]} \cD_m$.\\

    To compute the second moment for a client $m$,
    \begin{align}
        \bE_{\partialv, (x, y_c)\sim \cD} &\left[ \nabla \hat{f}_m(\partialw, \partialv; (x, y_c)) \cdot \nabla \hat{f}_m(\partialw, \partialv; (x, y_c))^T \right] \nonumber\\
        &= \frac{1}{|\cD|^2K^2} \sum_{\substack{(x, y_c \sim \cD)\\(\overline{x}, \overline{y}_c \sim \cD)}} \sum_{\substack{i\in[K]\\j\in [K]}} \bE \Bigg[ \left(\nabla f_m(\partialw; (x,y_c))v_{i,{\overline{m}}}v_{i, {\overline{m}}}^T \right) \cdot  \left(\nabla f_m(\partialw; (x,y_c))^Tv_{j,{\overline{m}}}v^T_{j, {\overline{m}}} \right) \Bigg] \label{eq:expanded-intermediate-norm-equation-1}
    \end{align}
    For simplicity, let $a$ and $b$ be two arbitrary vectors representing ${\bE \left[ \nabla f_m(\partialw; (x, y_c)) \right]}$ and ${\bE \left[ \nabla f_m(\partialw; (\overline{x}, \overline{y}_c))^T \right]}$, respectively.

    For the sum over all perturbations, we have two cases (all the expectations are under the randomness of partial $v$), 
    \begin{enumerate}
        \item If $i \neq j$. (Occurs $K(K-1)$ times)
        \begin{align}
            \bE &\left[ v_{i, \overline{m}}v_{i, \overline{m}}^T \; ab^T\; v_{j, \overline{m}}v_{j, \overline{m}}^T \right] = \bE \left[ v_{i, \overline{m}}v_{i, \overline{m}}^T \right] \cdot ab^T \cdot \bE \left[ v_{j, \overline{m}}v_{j, \overline{m}}^T \right]\\
            &\text{(since $v_i$ and $v_j$ are independent of each other and $a, b$ don't depend on $v$)} \nonumber \\
            &= I \cdot ab^T \cdot I = ab^T
        \end{align}
        \item If $i = j$. (Occurs $K$ times)
        \begin{align}
            \bE \left[ v_{i, \overline{m}}v_{i, \overline{m}}^T \; ab^T\; v_{j, \overline{m}}v_{j, \overline{m}}^T \right] &= \bE \left[ v_{i, \overline{m}}v_{i, \overline{m}}^T \; ab^T\; v_{i, \overline{m}}v_{i, \overline{m}}^T \right]\\
            &= \bE_{v_i} \left[ v_i^4 \right] \langle a, b \rangle
        \end{align}
        For all such $v_i$ with $i \in [K]$,
        \begin{align}
            \bE_v [v^{\otimes 4}] \langle a, b \rangle &= 3d \;\text{Sym}(I^{\otimes 4}) \langle a, b \rangle \\ 
            &= 2d \cdot ab^T + d\cdot I\cdot a^Tb
        \end{align}
    \end{enumerate}
    Plugging in the results of the above two cases in Equation~\ref{eq:expanded-intermediate-norm-equation-1},
    \begin{align}
        \therefore \bE &\left[ \nabla \hat{f}_m (\partialw, \partialv; (x, y_c)) \cdot \nabla \hat{f}_m(\partialw, \partialv; (x, y_c))^T \right] \nonumber\\
        &= \frac{1}{|\cD|^2 K^2} \sum_{\substack{(x, y_c \sim \cD)\\(\overline{x}, \overline{y}_c \sim \cD)}} \left[ K(K-1) ab^T + 2dK ab^T + dK \cdot I \cdot a^Tb\right] \\
        &= \frac{1}{|\cD|^2K} \sum_{\substack{(x, y_c \sim \cD)\\(\overline{x}, \overline{y}_c \sim \cD)}} \Bigg[ (2d + K - 1)  \bE\left[ \nabla f_m(\partialw; (x, y_c)) \nabla f_m(\partialw; (\overline{x}, \overline{y}_c))^T \right]  \nonumber\\
        &\qquad\qquad\qquad\qquad \qquad + d\cdot I \cdot  \bE\Big[ \nabla f_m (\partialw; (x, y_c)) \nabla f_m (\partialw; (\overline{x}, \overline{y}_c))^T \Big] \Bigg] \label{eq:expanded-intermediate-norm-equation-2}
    \end{align}
    Here, the randomness is under the sampled data. 
    
    For the sum over samples of $\cD$, we have two cases,
    \begin{enumerate}
        \item If $(x, y_c) \neq (\overline{x}, \overline{y}_c)$. (Occurs $|\cD|(|\cD|-1)$ times)
        \begin{align}
            \bE[\nabla f_m(\partialw; (x, y_c)) \nabla f_m(\partialw; (\overline{x}, \overline{y}_c))^T] &= \nabla F_m(\partialw) \nabla F_m(\partialw)^T
        \end{align}
        \item If $(x, y_c) = (\overline{x}, \overline{y}_c)$. (Occurs $|\cD|$ times)
        \begin{align}
            \bE[\nabla f_m(\partialw; (x, y_c)) \nabla f_m(\partialw; (\overline{x}, \overline{y}_c))^T] &= \nabla F_m(\partialw) \nabla F_m(\partialw)^T + \Sigma(\partialw)
        \end{align}
    \end{enumerate}
    Combining both the cases, we get 
    \begin{align}
        \bE_{(x, y_c)\sim\cD} [\nabla f_m(\partialw; (x, y_c)) \nabla f_m(\partialw; (\overline{x}, \overline{y}_c))^T] &= |\cD|^2 \nabla F_m(\partialw) \nabla F_m(\partialw)^T + |\cD| \cdot \Sigma(\partialw) \label{eq:expanded-intermediate-norm-equation-3}
    \end{align}
    Plugging in Equation~\ref{eq:expanded-intermediate-norm-equation-3} in Equation~\ref{eq:expanded-intermediate-norm-equation-2},
    \begin{align}
        \bE_{\partialv, (x, y_c)\sim\cD} &[\nabla \hat{f}_m(\partialw, \partialv; (x, y_c)) \nabla \hat{f}_m(\partialw, \partialv; (x, y_c))^T] \nonumber\\
        &= \frac{1}{|\cD|^2K} \Bigg[ (2d+K-1)|\cD|(|\cD| \nabla F_m(\partialw)\nabla F_m(\partialw)^T + \Sigma(\partialw) )  \nonumber\\
        &\qquad\qquad + d \cdot |\cD| \left(|\cD| \cdot ||\nabla F_m(\partialw)||^2 + \text{tr}\left(\Sigma(\partialw)\right) \right) \Bigg] \\
        &= \frac{(2d+K-1)}{K} \left( \nabla F_m(\partialw)\nabla F_m(\partialw)^T  
        +  \frac{1}{|\cD|} \Sigma(\partialw) \right)  \nonumber\\
        &\qquad + \frac{d}{K} \left(||\nabla F_m(\partialw)||^2 + \frac{1}{|\cD|} \text{tr}\left(\Sigma(\partialw)\right) \right) \\
        & = \frac{3d + K - 1}{K} \bE_{(x, y_c)\sim\cD} ||\nabla f_m(\partialw; (x, y_c))||^2 
    \end{align}
    Hence for client $m$, the expected norm of forward gradients under randomness of perturbations $v$ is,
    \begin{align}
        \bE_{\partialv, (x, y_c)\sim\cD} || \nabla \hat{f}_m(\partialw, \partialv; (x, y_c)) ||^2 &= \frac{3d + K - 1}{K} \bE_{(x, y_c)\sim\cD} ||\nabla f_m(\partialw; (x, y_c))||^2 \\
        &= \frac{3d + K - 1}{K} ||\nabla F_m(\partialw)||^2  \label{eq:one-client-forward-gradient-norm}  
    \end{align}
    For $i^{th}$ row of $\bE \left[\nabla \hat{f}(w, v; \cD)\right]$ of Equation~\ref{eq:repeat-expected-global-forward-gradient},
    \begin{align}
        \bE &\left|\left| \nabla \hat{f}_i(w_{\left[ \frac{(i-1)d}{M} + 1, \frac{id}{M}\right]}, v_{\left[ \frac{(i-1)d}{M} + 1, \frac{id}{M}\right]}; \cD)\right| \right|^2 = \bE \left| \left| \nabla f_i (w_{\left[ \frac{(i-1)d}{M} + 1, \frac{id}{M}\right]}) \right| \right|^2 \nonumber \\
        &\; + \frac{1}{(\subM)^2} \sum_{m \in \subcM_i} \sum_{c \in [C]} \left(\frac{n_c}{|\cD|} - \frac{n_{m, c} \alpha_c}{|\cD_m|} \right)^2 \bE_{(x,y_c) \sim \cD} \left| \left| \nabla \hat{f}_m (w_{\left[ \frac{(i-1)d}{M} + 1, \frac{id}{M}\right]}, v_{\left[ \frac{(i-1)d}{M} + 1, \frac{id}{M}\right]}; (x, y_c))  \right| \right|^2 
    \end{align}
    The left-hand side expectation under the randomness of sampled data $\cD$ and subset of random perturbations $v$.
    
    Plugging in Equation~\ref{eq:one-client-forward-gradient-norm} in the above equation and using $\overline{i} = \left[ \frac{(i-1)d}{M} + 1, \frac{id}{M}\right]$,
    \begin{align}
        \bE_{v_{\overline{i}}; \cD} &\left|\left| \nabla \hat{f}_i(w_{\overline{i}}, v_{\overline{i}}; \cD)\right| \right|^2 = \bE \left| \left| \nabla f_i (w_{\overline{i}}) \right| \right|^2 + \frac{(3d + K - 1)}{(\subM)^2 K} \sum_{m \in \subcM_i} \left| \left| \nabla F_m (w_{\overline{i}})\right| \right|^2\sum_{c \in [C]} \left(\frac{n_c}{|\cD|} - \frac{n_{m, c} \alpha_c}{|\cD_m|} \right)^2
    \end{align}
    
    Since $\nabla \hat{f}_i, \; \forall \subcM \subset \cM$ are independent of each other, we can compute the norm of $\nabla \hat{f}$ as follows,
    \begin{align}
        \bE_{v,\cD} &|| \nabla \hat{f} (w, v; \cD) ||^2  = \bE || \nabla f(w) ||^2 +\left( \frac{3d + K - 1}{(\subM)^2 K}\right) \sum_{m \in \cM} \bigg|\bigg| \nabla F_m(\partialw) \bigg|\bigg|^2 \sum_{c \in [C]} \left(\frac{n_c}{|\cD|} - \frac{n_{m, c} \alpha_c}{|\cD_m|} \right)^2\\
        &= \bE || \nabla f(w) ||^2 + \left( \frac{3d + K - 1}{(\subM)^2 K}\right) \sum_{m \in \cM} \bigg|\bigg|  \nabla F_m(\partialw) - \nabla f(\partialw) + \nabla f(\partialw) \bigg|\bigg|^2 \sum_{c \in [C]} \left(\frac{n_c}{|\cD|} - \frac{n_{m, c} \alpha_c}{|\cD_m|} \right)^2 \\
        &= \bE || \nabla f(w) ||^2 + \left(\frac{2(3d + K - 1)}{(\subM)^2 K}\right) \sum_{m \in \cM} \bigg|\bigg| \nabla F_m(\partialw) - \nabla f(\partialw) \bigg|\bigg|^2 \sum_{c \in [C]} \left(\frac{n_c}{|\cD|} - \frac{n_{m, c} \alpha_c}{|\cD_m|} \right)^2 \nonumber\\
        &\qquad + \left(\frac{2(3d + K - 1)}{(\subM)^2 K}\right) \sum_{m \in \cM} \bigg|\bigg| \nabla f(\partialw) \bigg|\bigg|^2 \sum_{c \in [C]} \left(\frac{n_c}{|\cD|} - \frac{n_{m, c} \alpha_c}{|\cD_m|} \right)^2
    \end{align}
    Using Assumption~\ref{asm:bounded-global-variance} and limited participation rate of $s$,
    \begin{align}
        \bE_{v, \cD} &|| \nabla \hat{f} (w, v; \cD) ||^2 = \bE || \nabla f(w) ||^2 + \left(\frac{2 \sigma_g^2 (1 - s) (3d + K - 1)}{\subM K}\right)  \sum_{m \in \cM}  \sum_{c \in [C]} \left(\frac{n_c}{|\cD|} - \frac{n_{m, c} \alpha_c}{|\cD_m|} \right)^2 \nonumber\\
        &\qquad + \left(\frac{2(3d + K - 1)}{\subM K}\right) \bE \left|\left| \nabla f(w) \right|\right|^2  \sum_{m \in \cM} \sum_{c \in [C]} \left(\frac{n_c}{|\cD|} - \frac{n_{m, c} \alpha_c}{|\cD_m|} \right)^2
    \end{align}
    Rearranging the terms, we get,
    \begin{align}
        \bE_{v, \cD} || \nabla \hat{f} (w, v; \cD) ||^2 &=  \bE || \nabla f(w) ||^2 \left( 1 + \left(\frac{2(3d + K - 1)}{\subM K}\right) \sum_{m \in \cM} \sum_{c \in [C]} \left(\frac{n_c}{|\cD|} - \frac{n_{m, c} \alpha_c}{|\cD_m|} \right)^2 \right)  \nonumber \\
        &\qquad + \left(\frac{2 \sigma_g^2 (1 - s) (3d + K - 1)}{\subM K}\right)  \sum_{m \in \cM}  \sum_{c \in [C]} \left(\frac{n_c}{|\cD|} - \frac{n_{m, c} \alpha_c}{|\cD_m|} \right)^2
    \end{align}
\end{proof}

\subsection{Convergence Rate of \projectname}
The general template for the convergence analysis of \projectname is similar to \textsc{FedAdam}, hence we will follow Theorem 2 of \textsc{AFO}~\cite{reddi2021adaptive}.
Our aim here is to highlight the differences in treatment of our gradient estimator $\nabla \hat{f}_m\; \forall m \in [M]$, and the aggregate global gradient $\nabla f$ as shown in Equation~\ref{eq:global-gradient}.

\begin{theorem}
Under the assumptions on $L$-smoothness (Asmp~\ref{asm:smoothness}), bounded global variance $\sigma_g^2$ of accumulated gradients (Asmp~\ref{asm:bounded-global-variance}), and bound on gradient magnitude $G$ (Asmp~\ref{asm:bounded-gradients}) and the following conditions on the local learning rate $\eta_\ell$,
    {
    \footnotesize
    \begin{align}
        \eta_\ell &= \min \Bigg \{ \cO \left( \frac{\tau^2}{\sqrt{\beta_2} \eta G L} \right)^{\frac{1}{2}}, \cO \left( \frac{1}{\sqrt{\beta_2} G} \right), \cO \left( \frac{\tau^3}{\sqrt{\beta_2} \sqrt{1 - \beta_2} G^2}\right)^{\frac{1}{2}}, \nonumber\\
        &\qquad \qquad \qquad \cO \left( \frac{\subM K}{\beta_2 G (3d + K - 1) \sum_{m \in [M]} \sum_{c \in [C]} \alpha_{m, c}^2} \right) \Bigg \};
    \end{align}
    }%
    \projectname satisfies the following bound,
    \begin{align}
        & \min_{0 \leq r \leq R} \bE_r || \nabla f(\wround{r})||^2 \leq \frac{f(\wround{0}) - \bE_R [f(\wround{R})]}{\eta R} \nonumber \\
        &\qquad + \left(2 + \frac{\eta \eta_\ell L}{2 \tau^2} + \frac{ \sqrt{1 - \beta_2}G \eta_\ell}{\tau^3} \right) \left(\frac{\sigma_g^2 (1 - s) (3d + K - 1)}{\subM K}\right)  \sum_{m \in \cM}  \sum_{c \in [C]} \alpha_{m, c}^2, 
    \end{align}
    where $R$ is the total round count, $w \in \bR^d$ are the trainable weights, $v \in \bR^d$ are the random perturbations, $K$ is the count of random perturbations per batch, $\eta$ is the global learning rate, $\tau$ is adaptability hyperparameter, $s$ is client sampling rate.
    The rest of the symbols are defined in Theorem~\ref{thm:estimation-global-gradient}.  
\end{theorem}
\begin{proof}
    As shown in Equation~\ref{eq:server-update}, an update of the model weights $w$ at server-side in \projectname looks like,
    \begin{align}
        \wround{r+1} = \wround{r} + \eta \frac{\deltar{r}}{\sqrt{\vr{r}} + \tau}
    \end{align}
    Using Assumption~\ref{asm:smoothness} and then the server-side update rule, we have
    \begin{align}
        f(\wround{r+1}) &\leq f(\wround{r}) + \left\langle \nabla f(\wround{r}), \wround{r+1} - \wround{r} \right\rangle + \frac{L}{2} || \wround{r+1} - \wround{r}||^2 \\
        &= f(\wround{r}) + \eta \left\langle \nabla f(\wround{r}), \frac{\deltar{r}}{\sqrt{\vr{r}} + \tau} \right\rangle + \frac{\eta^2L}{2} \sum_{i \in [d]} \frac{(\deltar{r}_i)^2}{\left(\sqrt{\vr{r}_i} + \tau\right)^2}
    \end{align}
    Taking expectation over randomness of round $r$ and simplifying the terms,
    \begin{align}
        \bE_r [f(\wround{r+1})] &\leq f(\wround{r}) + \eta \underbrace{\left\langle \nabla f(\wround{r}), \bE_r \left[ \frac{\deltar{r}}{\sqrt{\beta \vr{r-1}} + \tau}\right]  \right\rangle}_{R_1} + \frac{\eta^2 L}{2} \sum_{i\in[d]} \bE_r \left[ \frac{(\deltar{r}_i)^2}{(\sqrt{\vr{r}_i} + \tau)^2} \right] \nonumber\\
        &\qquad + \eta \underbrace{\left \langle \nabla f(\wround{r}), \bE_r \left[ \frac{\deltar{r}}{\sqrt{\vr{r}} + \tau} - \frac{\deltar{r}}{\sqrt{\beta \vr{r-1}} + \tau}\right] \right \rangle}_{R_2} \label{eq:one-round-progress}
    \end{align}
    Bounds for $R_2$ are derived in the exactly same manner as ``Bounding $R_2$'' in Theorem 2 of \textsc{FedAdam}~\cite{reddi2021adaptive},\begin{align}
        R_2 \leq \sqrt{1 - \beta_2} \bE_r \sum_{j=1}^d \frac{G}{\tau} \times \left[ \frac{(\deltar{r}_j)^2}{\sqrt{\vr{r}_j} + \tau} \right]
    \end{align} 
    \textbf{Bounding $R_1$} has a  different treatment due to the distinct aggregation strategy of \projectname:
    \begin{align}
        R_1 &= \left\langle \nabla f(\wround{r}), \bE_r \left[ \frac{\deltar{r}}{\sqrt{\beta \vr{r-1}} + \tau}\right]  \right\rangle
    \end{align}
    Since $\deltar{r}$ is piece-wise made of aggregations of forward gradients for several parts of the model weights, as shown in Equation~\ref{eq:true-global-gradient}, we first center each gradient piece for the computation of the squared norm,
    \begin{align}
        \therefore R_1 &= \left\langle \nabla f(\wround{r}), \bE_r \left[ \frac{- \eta_\ell \nabla \hat{f} (\wround{r}, v^{(r)}, \cD)}{\sqrt{\beta \vr{r-1}} + \tau}\right]  \right\rangle
    \end{align}
    Using $ab \leq (a^2 + b^2)/2$,
    \begin{align}
        R_1 \leq -\frac{\eta_\ell}{2} \sum_{j \in [d]} \frac{[\nabla f (\wround{r})]^2}{\sqrt{\beta \vr{r-1}_j} + \tau} + \frac{\eta_\ell}{2} \bE_r || \nabla \hat{f} (\wround{r}, v^{r}, \cD) ||^2
    \end{align}
    Using the result of Lemma~\ref{lma:forward-gradient-second-moment},
    \begin{align}
        \therefore R_1 &\leq -\frac{\eta_\ell}{2} \sum_{j \in [d]} \frac{[\nabla f (\wround{r})]^2}{\sqrt{\beta \vr{r-1}_j} + \tau} + \frac{\eta_\ell}{2} \left(\frac{2 \sigma_g^2 (1 - s) (3d + K - 1)}{\subM K}\right)  \sum_{m \in \cM}  \sum_{c \in [C]} \left(\frac{n_c}{|\cD|} - \frac{n_{m, c} \alpha_c}{|\cD_m|} \right)^2 \nonumber\\
        &\qquad + \frac{\eta_\ell}{2} \left( 1 + \left(\frac{2(3d + K - 1)}{\subM K}\right) \sum_{m \in \cM} \sum_{c \in [C]} \left(\frac{n_c}{|\cD|} - \frac{n_{m, c} \alpha_c}{|\cD_m|} \right)^2 \right) \bE_r || \nabla f(\wround{r}) ||^2
    \end{align}

    Putting $R_1$ and $R_2$ bounds in Equation~\ref{eq:one-round-progress},
    \begin{align}
        \bE_r [f(\wround{r+1})] &\leq f(\wround{r}) - \frac{\eta \eta_\ell}{2} \sum_{j \in [d]} \frac{[\nabla f (\wround{r})]^2}{\sqrt{\beta \vr{r-1}_j} + \tau} \nonumber\\
        &\qquad + \frac{\eta \eta_\ell}{2} \left(\frac{2 \sigma_g^2 (1 - s) (3d + K - 1)}{\subM K}\right)  \sum_{m \in \cM}  \sum_{c \in [C]} \left(\frac{n_c}{|\cD|} - \frac{n_{m, c} \alpha_c}{|\cD_m|} \right)^2 \nonumber\\
        &\qquad + \frac{\eta \eta_\ell}{2} \left( 1 + \left(\frac{2(3d + K - 1)}{\subM K}\right) \sum_{m \in \cM} \sum_{c \in [C]} \left(\frac{n_c}{|\cD|} - \frac{n_{m, c} \alpha_c}{|\cD_m|} \right)^2 \right) \bE_r || \nabla f(\wround{r}) ||^2 \nonumber \\
        &\qquad + \frac{\eta^2 L}{2} \sum_{i\in[d]} \bE_r \left[ \frac{(\deltar{r}_i)^2}{\vr{r}_i + \tau^2} \right] +  \frac{\eta \sqrt{1 - \beta_2}G}{\tau} \sum_{i\in[d]} \bE_r \left[ \frac{(\deltar{r}_i)^2}{\sqrt{\vr{r}_i} + \tau} \right]
    \end{align}
    Summing over $r=0$ to $R-1$ and using telescoping sum, we get
    \begin{align}
        \bE_R [f(\wround{R})] &\leq f(\wround{0}) 
        - \frac{\eta \eta_\ell}{2} \sum_{r=0}^{R-1} \sum_{j \in [d]} \frac{[\nabla f (\wround{r})]^2}{\sqrt{\beta \vr{r-1}_j} + \tau} \nonumber\\
        &\qquad + \frac{\eta \eta_\ell R}{2} \left(\frac{2 \sigma_g^2 (1 - s) (3d + K - 1)}{\subM K}\right)  \sum_{m \in \cM}  \sum_{c \in [C]} \left(\frac{n_c}{|\cD|} - \frac{n_{m, c} \alpha_c}{|\cD_m|} \right)^2 \nonumber\\
        &\qquad + \frac{\eta \eta_\ell}{2} \left( 1 + \left(\frac{2(3d + K - 1)}{\subM K}\right) \sum_{m \in \cM} \sum_{c \in [C]} \left(\frac{n_c}{|\cD|} - \frac{n_{m, c} \alpha_c}{|\cD_m|} \right)^2 \right) 
        \sum_{r=0}^{R-1} \bE_r || \nabla f(\wround{r}) ||^2 \nonumber \\
        &\qquad + \left( \frac{\eta^2 L}{2} + \frac{\eta \sqrt{1 - \beta_2}G}{\tau} \right) \underbrace{\sum_{r=0}^{R-1} \sum_{i\in[d]} \bE_r \left[ \frac{(\deltar{r}_i)^2}{\vr{r}_i + \tau^2} \right]}_{R_4} \label{eq:telescoping-sum}
    \end{align}
    \textbf{Bounding} $R_4$ follows a similar derivation as ``Bounding $R_1$'',
    \begin{align}
        R_4 &= \bE \sum_{r=0}^{R-1} \sum_{i\in[d]} \frac{(\deltar{r}_i)^2}{\vr{r}_i + \tau^2} = \bE \sum_{r=0}^{R-1} \sum_{i\in[d]} \frac{[-\eta_\ell \nabla \hat{f}(\wround{r}, v^{(r)}, \cD)]_i^2}{\vr{r}_i + \tau^2} \\
        &\leq \eta_{\ell}^2 \bE \sum_{r=0}^{R-1} \left| \left| \frac{\nabla \hat{f}(\wround{r}, v^{(r)}, \cD)}{\tau^2} \right| \right|^2
    \end{align}
    Using Lemma~\ref{lma:forward-gradient-second-moment} once again,
    \begin{align}
        \therefore R_4 &\leq \frac{\eta_{\ell}^2}{\tau^2} \left( 1 + \left(\frac{2(3d + K - 1)}{\subM K}\right) \sum_{m \in \cM} \sum_{c \in [C]} \left(\frac{n_c}{|\cD|} - \frac{n_{m, c} \alpha_c}{|\cD_m|} \right)^2 \right)  \bE \sum_{r=0}^{R-1} || \nabla f(\wround{r}) ||^2
        \nonumber \\
        &\qquad + \frac{\eta_{\ell}^2}{\tau^2} \left(\frac{2 \sigma_g^2 (1 - s) R (3d + K - 1)}{\subM K}\right)  \sum_{m \in \cM}  \sum_{c \in [C]} \left(\frac{n_c}{|\cD|} - \frac{n_{m, c} \alpha_c}{|\cD_m|} \right)^2
    \end{align}
    Updating Equation~\ref{eq:telescoping-sum} with the bounds of $R_4$,
    \begin{align}
        \bE_R &[f(\wround{R})] \leq f(\wround{0}) 
        - \frac{\eta \eta_\ell}{2} \sum_{r=0}^{R-1} \sum_{j \in [d]} \frac{[\nabla f (\wround{r})]^2}{\sqrt{\beta \vr{r-1}_j} + \tau} \nonumber\\
        &\qquad + \frac{\eta \eta_\ell R}{2} \left(\frac{2 \sigma_g^2 (1 - s) (3d + K - 1)}{\subM K}\right)  \sum_{m \in \cM}  \sum_{c \in [C]} \left(\frac{n_c}{|\cD|} - \frac{n_{m, c} \alpha_c}{|\cD_m|} \right)^2 \nonumber\\
        &\qquad + \frac{\eta \eta_\ell}{2} \left( 1 + \left(\frac{2(3d + K - 1)}{\subM K}\right) \sum_{m \in \cM} \sum_{c \in [C]} \left(\frac{n_c}{|\cD|} - \frac{n_{m, c} \alpha_c}{|\cD_m|} \right)^2 \right) 
        \sum_{r=0}^{R-1} \bE_r || \nabla f(\wround{r}) ||^2 \nonumber \\
        &\qquad + \left( \frac{\eta^2 L}{2} + \frac{\eta \sqrt{1 - \beta_2}G}{\tau} \right) \frac{\eta_{\ell}^2}{\tau^2} \left( 1 + \left(\frac{2(3d + K - 1)}{\subM K}\right) \sum_{m \in \cM} \sum_{c \in [C]} \left(\frac{n_c}{|\cD|} - \frac{n_{m, c} \alpha_c}{|\cD_m|} \right)^2 \right) \sum_{r=0}^{R-1} \bE_r || \nabla f(\wround{r}) ||^2 \nonumber \\
        &\qquad + \left( \frac{\eta^2 L}{2} + \frac{\eta \sqrt{1 - \beta_2}G}{\tau} \right) \frac{\eta_{\ell}^2}{\tau^2} \left(\frac{2 \sigma_g^2 (1 - s) R (3d + K - 1)}{\subM K}\right)  \sum_{m \in \cM}  \sum_{c \in [C]} \left(\frac{n_c}{|\cD|} - \frac{n_{m, c} \alpha_c}{|\cD_m|} \right)^2
    \end{align}
    Rearranging the terms,
    \begin{align}
        &\sum_{r=0}^{R-1} \sum_{j \in [d]} \frac{[\nabla f (\wround{r})]^2}{\sqrt{\beta \vr{r-1}_j} + \tau} \leq \frac{f(\wround{0}) - \bE_R [f(\wround{R})]}{\eta} \nonumber \\
        &\qquad + \left(\frac{2 R \sigma_g^2 (1 - s) (3d + K - 1)}{\subM K}\right)  \sum_{m \in \cM}  \sum_{c \in [C]} \left(\frac{n_c}{|\cD|} - \frac{n_{m, c} \alpha_c}{|\cD_m|} \right)^2 \nonumber\\
        &\qquad + \left( 1 + \left(\frac{2(3d + K - 1)}{\subM K}\right) \sum_{m \in \cM} \sum_{c \in [C]} \left(\frac{n_c}{|\cD|} - \frac{n_{m, c} \alpha_c}{|\cD_m|} \right)^2 \right) 
        \sum_{r=0}^{R-1} \bE_r || \nabla f(\wround{r}) ||^2 \nonumber \\
        &\qquad + \left({\eta L} + \frac{ 2\sqrt{1 - \beta_2}G}{\tau} \right) \frac{\eta_{\ell}}{\tau^2} \left( 1 + \left(\frac{2(3d + K - 1)}{\subM K}\right) \sum_{m \in \cM} \sum_{c \in [C]} \left(\frac{n_c}{|\cD|} - \frac{n_{m, c} \alpha_c}{|\cD_m|} \right)^2 \right) \sum_{r=0}^{R-1} \bE_r || \nabla f(\wround{r}) ||^2 \nonumber \\
        &\qquad + \left( \frac{\eta L}{2} + \frac{ \sqrt{1 - \beta_2}G}{\tau} \right) \frac{\eta_{\ell}}{\tau^2} \left(\frac{R \sigma_g^2 (1 - s) (3d + K - 1)}{\subM K}\right)  \sum_{m \in \cM}  \sum_{c \in [C]} \left(\frac{n_c}{|\cD|} - \frac{n_{m, c} \alpha_c}{|\cD_m|} \right)^2
    \end{align}
    Getting a lower bound for the left hand side term through using the fact $\sqrt{\vr{r-1}} \leq \eta_\ell K G$ from Theorem 2 of AFO~\cite{reddi2021adaptive},
    \begin{align}
        \sum_{r=0}^{R-1} \sum_{j=1}^d \frac{\bE_r [\nabla f(\wround{r})]^2_j}{\sqrt{\beta_2 \vr{r-1}_j} + \tau} &\geq  \sum_{r=0}^{R-1} \sum_{j=1}^d \frac{\bE_r [\nabla f(\wround{r})]^2_j}{\sqrt{\beta_2}\eta_\ell K G + \tau} \geq \frac{R}{\sqrt{\beta_2}\eta_\ell K G + \tau} \min_{0 \leq r \leq R} \bE_r || \nabla f(\wround{r})||^2
    \end{align}
    \begin{align}
        \therefore & \frac{R}{\sqrt{\beta_2}\eta_\ell K G + \tau} \min_{0 \leq r \leq R} \bE_r || \nabla f(\wround{r})||^2 \leq \frac{f(\wround{0}) - \bE_R [f(\wround{R})]}{\eta} \nonumber \\
        &\qquad + \left(2 + \frac{\eta \eta_\ell L}{2 \tau^2} + \frac{ \sqrt{1 - \beta_2}G \eta_\ell}{\tau^3} \right) \left(\frac{R \sigma_g^2 (1 - s) (3d + K - 1)}{\subM K}\right)  \sum_{m \in \cM}  \sum_{c \in [C]} \left(\frac{n_c}{|\cD|} - \frac{n_{m, c} \alpha_c}{|\cD_m|} \right)^2 \nonumber\\
        &\qquad + \left( 1 + \left(\frac{2(3d + K - 1)}{\subM K}\right) \sum_{m \in \cM} \sum_{c \in [C]} \left(\frac{n_c}{|\cD|} - \frac{n_{m, c} \alpha_c}{|\cD_m|} \right)^2 \right) 
         \bE \sum_{r=0}^{R-1} || \nabla f(\wround{r}) ||^2 \nonumber \\
        &\qquad + \left({\eta L} + \frac{ 2\sqrt{1 - \beta_2}G}{\tau} \right) \frac{\eta_{\ell}}{\tau^2} \left( 1 + \left(\frac{2(3d + K - 1)}{\subM K}\right) \sum_{m \in \cM} \sum_{c \in [C]} \left(\frac{n_c}{|\cD|} - \frac{n_{m, c} \alpha_c}{|\cD_m|} \right)^2 \right)  \bE \sum_{r=0}^{R-1} || \nabla f(\wround{r}) ||^2 
    \end{align}
    Considering the coefficients of $\bE ||\nabla f(\wround{r}) ||^2$ terms, conditioning on the following inequality,
    \begin{align}
        \frac{R}{\sqrt{\beta_2}\eta_\ell K G + \tau} \geq \left( \frac{\eta \eta_\ell L}{\tau^2} + \frac{2 \eta_\ell \sqrt{1 - \beta_2} G}{\tau^3} + 1\right) \left( 1 + \frac{2(3d + K - 1)}{\subM K} \sum_{m \in [M]} \sum_{c \in [C]} \left( \frac{n_c}{|\cD|} - \frac{n_{m,c}\alpha_c}{|\cD_m|} \right)^2 \right)
    \end{align}
    We get the following condition on the local learning rate, 
    \begin{align}
        \eta_\ell &= \min \Bigg \{ \cO \left( \frac{\tau^2}{\sqrt{\beta_2} \eta G L} \right)^{\frac{1}{2}}, \cO \left( \frac{1}{\sqrt{\beta_2} G} \right), \cO \left( \frac{\tau^3}{\sqrt{\beta_2} \sqrt{1 - \beta_2} G^2}\right)^{\frac{1}{2}}, \nonumber\\
        &\qquad \qquad \qquad \cO \left( \frac{\subM K}{\beta_2 G (3d + K - 1) \sum_{m \in [M]} \sum_{c \in [C]} \left( \frac{n_c}{|\cD|} - \frac{n_{m,c}\alpha_c}{|\cD_m|} \right)^2} \right) \Bigg \}
    \end{align}
    for the following bound on the gradient norm,
    \begin{align}
        & \min_{0 \leq r \leq R} \bE_r || \nabla f(\wround{r})||^2 \leq \frac{f(\wround{0}) - \bE_R [f(\wround{R})]}{\eta R} \nonumber \\
        &\qquad + \left(2 + \frac{\eta \eta_\ell L}{2 \tau^2} + \frac{ \sqrt{1 - \beta_2}G \eta_\ell}{\tau^3} \right) \left(\frac{\sigma_g^2 (1 - s) (3d + K - 1)}{\subM K}\right)  \sum_{m \in \cM}  \sum_{c \in [C]} \left(\frac{n_c}{|\cD|} - \frac{n_{m, c} \alpha_c}{|\cD_m|} \right)^2 
    \end{align}

\end{proof}



\end{document}